%% file: main.tex
\documentclass{article}

\PassOptionsToPackage{numbers,sort&compress}{natbib}

\usepackage[main,final]{neurips_2026}

\usepackage[utf8]{inputenc}
\usepackage[T1]{fontenc}
\usepackage{hyperref}
\usepackage{url}
\usepackage{booktabs}
\usepackage{amsfonts}
\usepackage{amsmath}
\usepackage{amssymb}
\usepackage{array}
\usepackage{multirow}
\usepackage{bm}
\usepackage{pifont}
\usepackage{graphicx}
\usepackage{float}
\usepackage{adjustbox}
\usepackage[most]{tcolorbox}
\usepackage[font=small,labelfont=bf,tableposition=bottom]{caption}
\usepackage[skip=3pt]{subcaption}
\usepackage{capt-of}
\usepackage{xcolor}
\usepackage{xspace}
\usepackage{microtype}
\usepackage{cleveref}

\newcommand{\myparagraph}[1]{\noindent\textbf{#1.}\xspace}

\DeclareFontFamily{U}{stix2bb}{}
\DeclareFontShape{U}{stix2bb}{m}{n}{<-> stix2-mathbb}{}
\NewDocumentCommand{\indicator}{}{\text{\usefont{U}{stix2bb}{m}{n}1}}

\input{preamble}

\title{SIEVES: Selective Prediction Generalizes through Visual Evidence Scoring}

\author{%
  Hector G. Rodriguez \\
  TU Darmstadt \& hessian.AI, Germany
  \And
  Marcus Rohrbach \\
  TU Darmstadt \& hessian.AI, Germany
}

\begin{document}

\maketitle
\input{sec/0_abstract}
\input{figs/ood_avg_cov_panels.tex}
\input{sec/1_intro}
\input{sec/2_related}

\input{sec/3_method}
\input{sec/4_experiments}
\input{sec/5_results}
\input{sec/6_conclusion}

\input{sec/6_limitations}

\input{sec/6_future_work}

\begin{ack}
This research was partially funded by an Alexander von Humboldt Professorship in
Multimodal Reliable AI, sponsored by the Federal Ministry of Research,
Technology, and Space (BMFTR), by a
LOEWE-Spitzen-Professur (LOEWE/4a//519/\allowbreak 05.00.002(0010)/93), and has
benefited from the Excellence Cluster ``Reasonable AI'' by the German Research
Foundation (Deutsche Forschungsgemeinschaft -- DFG) under Germany's Excellence
Strategy -- EXC-3057.
We gratefully acknowledge support from the hessian.AI Service Center (funded by
the Federal Ministry of Research, Technology and Space, BMFTR, grant no.
16IS22091) and the hessian.AI Innovation Lab (funded by the Hessian Ministry
for Digital Strategy and Innovation, grant no. S-DIW04/\allowbreak 0013/003).
We thank Tobias Wieczorek and Mohamed Abdelsalam for their feedback on a
previous version of this manuscript.
\end{ack}

{\small
\bibliographystyle{plainnat}
\bibliography{main}
}

\input{sec/X_suppl}

\newpage
\input{checklist.tex}

\end{document}

%% file: preamble.tex
\usepackage{colortbl}
\usepackage{placeins}
\usepackage{adjustbox}
\usepackage[most]{tcolorbox}

\newcommand{\zoomInTool}{\includegraphics[height=1em]{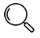}}

\newcommand{\strutless}{\vrule height 0.8em depth 0.15em width 0pt\relax} %
\newtcblisting{promptfloatlisting}[1][]{%
  enhanced,
  colback=white,
  colframe=black!80,
  boxrule=1pt,
  arc=2.5mm,
  left=2mm,
  right=2mm,
  top=0.5mm,
  bottom=0.5mm,
  title filled=false,
  colbacktitle=white,
  coltitle=black,
  fonttitle=\bfseries,
  listing only,
  listing options={
    basicstyle=\ttfamily\footnotesize,
    breaklines=true,
    breakatwhitespace=true,
    columns=fullflexible,
    keepspaces=true,
    showstringspaces=false
  },
  #1
}
\newcommand{\promptfloatmaxheight}{0.74\textheight}
\newenvironment{fitpromptbox}{%
  \begin{adjustbox}{max totalsize={\linewidth}{\promptfloatmaxheight},center}
  \begin{minipage}{\linewidth}
}{%
  \end{minipage}
  \end{adjustbox}
}

\usepackage{pifont}
\usepackage{bm}
\newcommand{\cmark}{\ding{51}}
\newcommand{\xmark}{\ding{55}}
\usepackage{multirow}
\newcommand{\accwithbars}[2][0.25ex]{%
  \begingroup
    \setbox0=\hbox{#2}\dimen0=\wd0
  \endgroup
  \vbox{\hbox to \dimen0{\hfil\rule{0.5pt}{0.9em}\hfil}%
        \vskip -#1
        \hbox to \dimen0{\hfil #2\hfil}%
        \vskip -#1
        \hbox to \dimen0{\hfil\rule{0.5pt}{0.9em}\hfil}}
}
\newcommand{\accbars}[4][0.25ex]{%
  \begingroup
    \setbox0=\hbox{#3}\dimen0=\wd0
  \endgroup
  \vbox{\hbox to \dimen0{\hfil\rule{0.5pt}{#2}\hfil}%
        \vskip -#1
        \hbox to \dimen0{\hfil #3\hfil}%
        \vskip -#1
        \hbox to \dimen0{\hfil\rule{0.5pt}{#4}\hfil}}
}

\usepackage{array}
\makeatletter
\newcommand{\SetUniformRowHeight}[1]{%
  \setlength{\extrarowheight}{\dimexpr #1 - \ht\@arstrutbox - \dp\@arstrutbox\relax}%
}
\newcommand{\accAutoInner}[3][0.25ex]{%
  \begingroup
    \dimen0=\ht\@arstrutbox
    \advance\dimen0 by \dp\@arstrutbox
    \advance\dimen0 by \extrarowheight
    \count0=#2\relax
    \dimen2=\dimen0 \multiply\dimen2 by \count0
    \setbox0=\hbox{#3}
    \dimen4=\ht0 \advance\dimen4 by \dp0
    \dimen6=#1 %
    \dimen8=\dimen2 \advance\dimen8 by -\dimen4 \advance\dimen8 by -2\dimen6
    \ifdim \dimen8<0pt \dimen8=0pt \fi
  \dimen10=\dimen8 \divide\dimen10 by 2 %
  \ifdim \dimen10<0.3ex \dimen10=0.3ex \fi
    \xdef\accHalfDim{\the\dimen10}
    \xdef\accPadDim{\the\dimen6}
    \xdef\accBoxWidth{\the\wd0}
  \endgroup
  \leavevmode
  \rlap{\raisebox{\dimexpr \ht0+\accPadDim\relax}{\makebox[\accBoxWidth]{\rule{0.5pt}{\accHalfDim}}}}%
  \rlap{\raisebox{\dimexpr -\accHalfDim-\accPadDim\relax}{\makebox[\accBoxWidth]{\rule{0.5pt}{\accHalfDim}}}}%
  #3%
}

\makeatother

%% file: sec/0_abstract.tex
\begin{abstract}
As the accuracy of Multimodal large language models (MLLMs) on traditional visual question answering benchmarks approaches saturation, reliable deployment requires satisfying low error tolerances in real-world out-of-distribution (OOD) scenarios.
Selective prediction aims to improve coverage, i.e.\ the share of inputs the system answers, while adhering to a user-defined risk level.
This is typically achieved by assigning a confidence score to each answer and abstaining on those that fall below a certain threshold.
Existing selective prediction methods rely on text-only reasoning and learn to predict confidences simply based on whether the answer was correct or not.
In contrast, we introduce SIEVES (Selective Prediction through Visual Evidence Scoring), which first requires the answering model to provide visual evidence supporting its answer, by returning crops of the image alongside their textual reasoning and answer.
Second, a lightweight selector is trained to predict the quality of this evidence, assessing how well the provided crops localize the relevant visual elements and if the final answer is coherent with this evidence.
We show that SIEVES improves coverage by up to six times on challenging OOD benchmarks (V* Bench, HR-Bench-8k, MME-RealWorld-Lite, VizWiz, and AdVQA), compared to non-grounding baselines.
Beyond better generalization to OOD tasks, the design of the SIEVES selector enables transfer to proprietary reasoners without access to their weights or logits, such as o3 and Gemini-3-Pro, providing coverage boosts beyond those attributable to accuracy alone.
We highlight that SIEVES generalizes across all tested OOD datasets and reasoner models, without benchmark- or reasoner-specific training or adaptation.

\end{abstract}

%% file: figs/ood_avg_cov_panels.tex
\begin{figure}[H]
\centering
    \includegraphics[width=\linewidth]{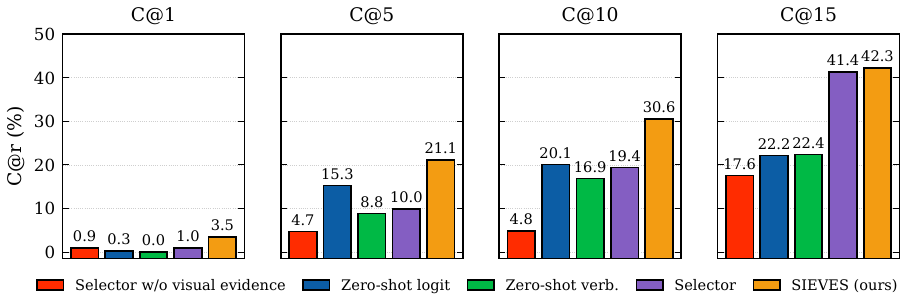}\vspace{-0.3cm}
    \caption{\textbf{OOD coverage at varying risk levels for frontier proprietary reasoners not used for training the (SIEVES) selector.}
    C@$r$ averaged across OOD benchmarks and o3 and Gemini-3-Pro reasoners, excluding trivial cases with reasoner error $<$ risk.
    SIEVES achieves the highest coverage, especially at low risk.
    }
    \label{fig:ood_avg_cov_panels}
\vspace{-6pt}
\end{figure}

%% file: sec/1_intro.tex
\section{Introduction}
\label{sec:intro}
Recent large language models (LLMs) \citep{anthropic-claude-3-5,deepseek-r1} and vision-language models (VLMs) \citep{openai-o3,qwen2.5-vl} can perform an increasing number of economically valuable tasks autonomously, from document understanding \citep{Venter2025,LucariniPelzSharpe2025} to assistive visual answering \citep{be-my-ai,lookout-google}.
Often in these tasks, mistakes can be costly, both in human and financial terms.
Selective prediction enables AI systems to decide which tasks are beyond the reach of these models, and for which it should abstain or defer to an expert \citep{wen2025knowyourlimits}.

Visual Question Answering (VQA) has been a good testbed for VLMs.
Nonetheless, even as traditional VQA benchmarks \citep{antol2015vqa,goyal2017making} become saturated from an accuracy perspective \citep{VQAChallenge2021}, selective prediction remains a challenge \citep{dancette2024selective}.
There are several reasons why selective prediction continues to be difficult.
Most importantly, machine learning (ML) systems are predominantly designed to answer as many questions correctly as possible, with little or no training to abstain when in doubt \citep{kalai2025language}.
Since the user's risk tolerance is unknown in advance, a truly binary abstain/answer decision is too rigid.
Selective prediction, therefore, assigns each answer a numerical confidence score, which makes the task harder: the selector must now rank the full distribution of answer-question pairs.

Existing selective prediction methods estimate confidence in a response given simply the answer \citep{bhatt2024taughttoknow}, standard textual reasoning \citep{verbalized_confidence_triggers_self_verification_2025}, or model internals \citep{chen-etal-2023-adaptation,mushtaq2025harmony,whitehead2022reliable}.
Estimating confidence using these signals is hard, as confidence should depend not only on the final answer, but also on the evidence used to reach it.

Methods that rely on model internals, such as log-probabilities or hidden activations, are also model-specific \citep{dancette2024selective,mushtaq2025harmony,whitehead2022reliable} and cannot be applied to frontier models when those signals are unavailable through proprietary APIs\footnote{OpenAI: \url{https://community.openai.com/t/logprobs-deprecated-for-gpt-5-models/1355427}; Anthropic: \url{https://docs.anthropic.com/en/api/openai-sdk\#detailed-openai-compatible-api-support}.}. %
Other approaches \citep{khan2024uncertainty,srinivasan2024recovrr} make selective prediction much more costly than the original answering task.
Here, we aim to solve    such  limitations.

We propose Selective Prediction through Visual Evidence Scoring (SIEVES) for generalizable selective prediction.
A strong reasoner that can zoom-in on the high-resolution image produces multimodal chain-of-thought reasoning with zoomed-in crops before providing a final answer.
This exposes rich evidence that makes confidence prediction easier.
We train the selector to take as input the full conversation, i.e.\ question, image, the multimodal chain-of-thought (MM-CoT) with crops, and the final answer, and output a scalar confidence that the answer is correct.
Crucially, the SIEVES selector does not only predict a single confidence score. %
Since the MM-CoT exposes where the model looked at before answering, the selector also explicitly assesses the quality of the localization shown in the MM-CoT.
Concretely, the selector predicts three signals: \emph{correctness} (is the final answer accurate?), \emph{localization} (did the model look at the right part of the image?), and \emph{coherence} (does the visual evidence support the answer?).
Training the selector to explicitly score this evidence at a fine-grained level helps distinguish well-supported answers from answers based on weak or irrelevant evidence, leading to more precise confidence estimates.

We evaluate on a wide range of VQA benchmarks, which are out-of-distribution for the reasoner and selector, requiring varying degrees of generalization.
These benchmarks range from high-resolution natural image benchmarks (V*-bench, HR-Bench-8k), diverse high resolution tasks including diagrams, tables, and remote sensing (MME-RealWorld-Lite), to real-world questions from blind users (VizWiz) and Human-Adversarial problems (AdVQA).
SIEVES improves the coverage at relevant risk tolerances by up to six times compared to non-grounding baselines, as shown in \Cref{tab:main_results}.

Our main contributions can be summarized as follows:

\begin{enumerate}
    \item We introduce visual evidence for selective prediction.
    A reasoner uses a zoom-in tool to provide a crop which serves as visual evidence to support the final answer.
    The evidence is used by the selector to better assess how trustworthy the answer is.
    The selector uses precise observable outputs instead of model internals, so it can be applied to different reasoners.
    \item We train the selector to score two fine-grained properties of the evidence in addition to answer correctness:
    whether the reasoner localized the relevant image region and whether the evidence supports the answer.
    This helps differentiate well-documented response from answers based on weak proof, hence allowing for improved confidence prediction.
\end{enumerate}

%% file: sec/2_related.tex
\section{Related work}
\label{sec:related}
\myparagraph{Reasoning with tools} 
Tool usage has become central to the recent explosion in reasoning capabilities of VLMs.
For VQA tasks, providing VLMs with the ability to use zoom-in tools has significantly improved the ability to answer questions about high-resolution images, through a form of multimodal chain-of-thought reasoning \citep{suPixelReasonerIncentivizing2025,zhengDeepEyesIncentivizingThinking2025,VisualSketchpad}.
Frontier VLMs can benefit from access to these tools without explicit training \citep{VisualSketchpad,openai_o3_visual_search_2025}, whereas smaller models are trained specifically to elicit such zooming in capability \citep{suPixelReasonerIncentivizing2025,zhengDeepEyesIncentivizingThinking2025}.
However, this zooming-in ability has only been, to the best of our knowledge, studied as a means of improving accuracy.
In this work, we propose leveraging the multimodal chain-of-thought that results from zooming in to assess how well grounded the final answer is.
Therefore, we do not focus on improving the abilities of these models to zoom-in, but rather use these existing abilities to make answer correctness prediction easier.
Unlike existing methods, which only show the benefits of zooming-in to improve accuracy in high-resolution VQA tasks, we employ the MM-CoT as a source of answer grounding evidence.
Using this MM-CoT, the selector model has a richer context to output a confidence score for each answer.
Intuitively, the ability to analyze where the model looked before answering eases confidence score prediction. %

\myparagraph{Test-time scaling via verification}
Our work has connections to the use of external models to rank different model generations to the same question \citep{cobbe2021trainingverifiers,lightman2023verifystep}, namely, using verifiers for test-time scaling.
However, we focus on the selective prediction task, which requires ranking over the joint space of questions and answers, whereas verification effectively factors the problem by conditioning on the question.
This makes the problem strictly harder, as scores must be comparable across different questions.
Furthermore, verifiers are usually models of the same size and inference budget (i.e., chain-of-thought length) as the reasoners \citep{singhi2025whentosolve}.
This implies that verification is as hard as answering the question itself.
Here, we demonstrate that by having a reasoner provide visual evidence for the answer, and explicitly learning to estimate the quality of such evidence, confidence score prediction is made easier, and a smaller discriminative selector can be used.

\myparagraph{Selective prediction}
Selective prediction formalizes reliable deployment by maximizing the portion of answered queries (coverage) under a user-specified error rate (risk), typically summarized with risk-coverage curves or AURC.
Early approaches rely on intrinsic model uncertainty, e.g., maximum softmax probability or logit margins, to decide when to abstain \citep{chow1970survey,cortes2016boosting,geifmanSelectiveClassificationDeep2017,xin2021artofabstention}. %
Such approaches are not directly applicable to generative models and suffer from the gap between the accuracy-maximizing training objective and the selective prediction task.
\citet{whitehead2022reliable} demonstrate that it is effective to learn a dedicated lightweight selector to predict confidence in the final answer given both model's internal representations and question-image input.
\citet{dancette2024selective} refine this for out-of-distribution settings, and \citet{mushtaq2025harmony} abstain on answers from VLMs.
However, hidden activations are entirely model dependent, tying the selector to a specific model architecture.
Furthermore, such model internals are not available for frontier models queried through proprietary APIs, which limits the applicability of such selectors.
In our work, we design a selector that relies only on observable outputs, and hence is model-agnostic.
Concurrently to this work, other methods train the answering model itself to verbalize its confidence \citep{damani2026beyond,rodriguez2026care}.
This requires training the reasoner, whereas SIEVES trains a small selector that can be combined with the strongest proprietary reasoners.
Moreover, these methods have only been applied to text-only tasks.

Given the limitations of trained selectors that require model internals, recent works have developed alternative methods for abstaining with generative LLMs or VLMs.
\citet{khan2024uncertainty} propose consistency as a proxy for certainty and use it to abstain. 
\citet{srinivasan2024recovrr} utilize an elaborate prompting scheme to use zero-shot selectors. 
Both of these methods make selective prediction much more expensive than answering, can only provide coarse confidence scores, and do not demonstrate significant coverage at demanding risk levels. 
In contrast, by using the multimodal reasoning process as a source of evidence for the selector, we avoid duplicating the compute devoted to solving the task, and can provide better confidence estimates with a smaller selector.

%% file: sec/3_method.tex
\section{\textbf{S}elective Prediction through \textbf{V}isual \textbf{E}vidence \textbf{S}coring}
\label{sec:method}
\input{figs/comparison_pipeline.tex}

We approach selective prediction for visual question answering by combining a reasoner that can zoom-in on the image and a selector that outputs a confidence score for the answer.

\subsection{Providing visual evidence via zooming-in tool}
The reasoner model $f$ is tasked with providing an answer $A$ to the question $Q$ about the image $I$.
In the process, since it benefits answer accuracy, it produces multimodal chain-of-thought reasoning, denoted as $R$.
Therefore, after the reasoner answers
\begin{align}
    \{R, A\} = f(Q, I),
    \label{eq:reasoner_output}
\end{align}
we have a conversation tuple $\{Q, I, R, A\}$.
We use reasoners that can use a zoom-in tool. 
Recipes for teaching open-source weaker models to learn this ability have become commonplace \citep{suPixelReasonerIncentivizing2025}, and several generations of frontier VLMs have been able to do this \citep{openai_o3_visual_search_2025}, so obtaining an answering system that is able to reason by pointing at the relevant sections of an image should not be a concern when tackling highest-stakes scenarios that selective prediction aims to solve.

 \subsection{Visual Evidence Scoring}

\myparagraph{Fine-grained confidence scoring using visual evidence}
Standard selective prediction learns to predict confidence scores that infer whether a final answer matches the ground-truth label \citep{dancette2024selective,mushtaq2025harmony,whitehead2022reliable}.
However, this signal alone does not capture \emph{why} an answer is correct or incorrect, making it prone to overfitting to in-distribution patterns like question domain or reasoner patterns, and degrading under distribution shift.
Intuitively, if the reasoner provides visual evidence for its answer, and the quality of that evidence can be evaluated, the confidence prediction task becomes easier and more robust.
For VQA, in addition to the standard answer correctness confidence ($c_{corr}$), two natural ways of scoring visual evidence signals arise: whether the model localized the relevant region ($c_{loc}$), and whether its final answer is coherent with the visual evidence it provides ($c_{coh}$).

As shown in \Cref{fig:comparison_pipeline}, the selector receives question, image, MM-CoT and final answer tuple, and outputs three confidence values: $c_{corr}$ that the answer is correct, $c_{loc}$ that the relevant visual information was localized, and $c_{coh}$ that the final answer is coherent with the evidence. 
Formally,  
\begin{align}
    \{c_{corr}, c_{loc}, c_{coh}\} &= s(Q, I, R, A)\in [0,1]^3.
    \label{eq:selector_output}
\end{align}
Finally, these individual confidence scores are aggregated to obtain a single confidence score $c_{sel}$.
This confidence score, paired with a threshold, can enable abstention to maintain a desired error rate on the answered questions.

\myparagraph{Learning localization confidence}
The predicted localization confidence score $c_{loc}$ aims to convey whether the model gathers the relevant visual information required to answer correctly.
To learn this confidence score, the model's predicted crops are matched to the ground-truth bounding boxes.
Concretely, the ground-truth localization signal $g_{loc}$ is computed as the intersection-over-ground-truth (IoGT) of the predicted crop(s) and the ground truth bounding box(es).
For the i-th ground-truth bounding box $B^i_{gt}$ and the j-th predicted crop $B^j_{crop}$, 
\begin{align}
    \text{IoGT}^{i,j} = \frac{B^i_{gt} \cap B^j_{crop}}{B^i_{gt}}.
\end{align}
If more than one ground truth bounding box is provided, we average the IoGT over all $N$ ground truth boxes, allowing for one predicted crop to match multiple ground truth boxes.
Explicitly, we define the mean IoGT, $mIoGT$, as
\begin{align}
    \text{mIoGT} &= \frac{1}{N} \sum_{i=1}^N \max_j \text{IoGT}^{i,j}. %
\end{align}
Empirically, we find that discretizing this signal, namely using a binary spatial recall \citep{sr24muller}, performs better than regressing the continuous value. More concretely, 
\begin{align}
    g_{loc}=\indicator[\text{mIoGT} \geq 0.75] \in \{0,1\}.
\end{align}
Although IoGT is less common than IoU in object detection, it is better aligned with our setting: the crop need not tightly segment the object, but must contain enough of it for the reasoner to answer.
\Cref{app:localization_target} explicitly studies IoGT against IoU-based alternatives and threshold variants.

\myparagraph{Learning crop-answer coherence}
The coherence confidence score tries to model whether the answering model, after correctly localizing the relevant region, has an answer that is coherent with the relevant crop.
Let $R^j_{crop}$ denote the image crop corresponding to bounding box $B^j_{crop}$. The ground-truth coherence signal $g_{coh}$ is computed as the maximum similarity between any crop and the text of the last message, which includes the final textual reasoning and answer. Formally,
\begin{align}
    g_{coh}=g_{loc} \cdot \max_j sim(R^j_{crop},[R_{last},A]) \in \{0,1\}.
\end{align}
Empirically, we find it best to determine similarity by asking an external VLM (Qwen 2.5 VL 7B \citep{qwen2.5-vl}) for a binary coherence judgment $g_{coh} \in \{0,1\}$, given the zoomed-in crops and last message.
Rather than requiring manual annotation, $g_{coh}$ offers a practical and scalable training signal, which we show empirically improves OOD selective prediction.
As a sanity check, we run a small human evaluation on 200 randomly chosen training samples and find that human annotations match the same $g_{coh}$ label in 92\% of cases.
Notably, if the localization is not correct, the coherence score is 0.
The inclusion of the coherence signal, $g_{coh}$, is motivated by the observation that, even after zooming in to the relevant region, the reasoner would in some scenarios provide an answer that was not correlated with the provided crop.
\Cref{fig:comparison_pipeline} shows an example of these labels.
In \Cref{sec:results}, \Cref{tab:coherence_ablation_id_ood} analyzes the effect of model size on coherence labeling.

\myparagraph{Training the selector}
For training the selector, all objectives are learned using binary cross-entropy (BCE) loss. 
The final loss becomes a weighted sum of the losses per objective, with $y$ indicating whether the answer is correct,
\begin{align}
    \mathcal{L} &= \lambda_{corr} \cdot \text{BCE}(c_{corr}, y) + \lambda_{loc} \cdot \text{BCE}(c_{loc}, g_{loc}) + \lambda_{coh} \cdot \text{BCE}(c_{coh}, g_{coh}).
    \label{eq:loss}
\end{align}
\(\lambda_{corr}\), \(\lambda_{loc}\), and \(\lambda_{coh}\) represent the scalar weights for the correctness, localization, and coherence objectives.
Tables and figures abbreviate these weights as $\lambda_c$, $\lambda_l$, and $\lambda_h$, respectively.
In \Cref{sec:results}, \Cref{tab:weight_ablation_id_ood} demonstrates that including all objectives is better than excluding any, and shows that our method is robust to variations in the weight values.

\myparagraph{Final confidence prediction}
At inference, we use the predicted visual evidence scores and the answer correctness confidence prediction, to obtain a final confidence value. 
These three scores are combined using the same weights as for training, to obtain the final confidence score $c_{\mathrm{sel}}$, i.e.\,
\begin{align}
    c_{\mathrm{sel}} = \lambda_{corr} \cdot c_{corr} + \lambda_{loc} \cdot c_{loc} + \lambda_{coh} \cdot c_{coh}.
\end{align}

%% file: figs/comparison_pipeline.tex
\begin{figure*}[t]
    \centering
    \includegraphics[width=\textwidth]{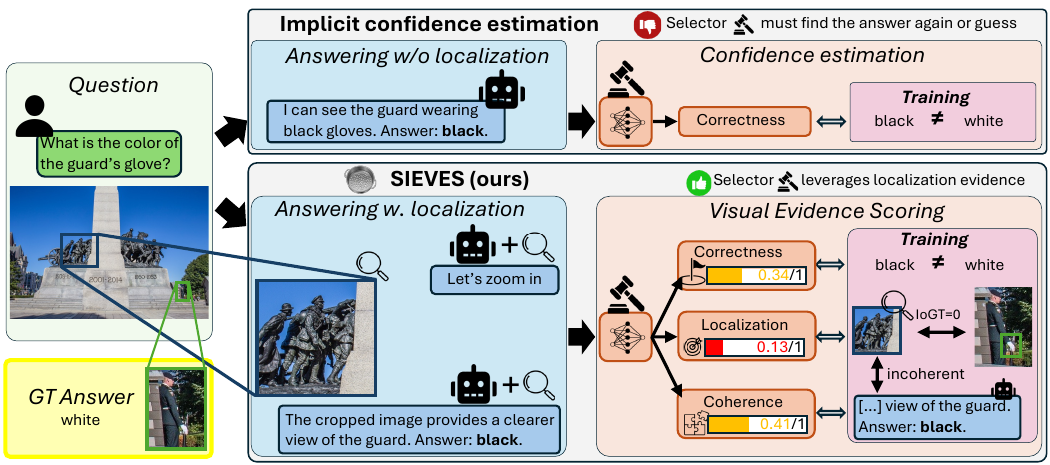}\vspace{-0.1cm}
    \caption{\textbf{Confidence from visual evidence.} \textbf{Top:} A correctness-only selector scores an answer from the image and text trace. \textbf{Bottom:} SIEVES also receives the reasoner's image crops and scores answer correctness, localization, and crop--answer coherence. Their weighted sum is the confidence used for abstention.}
    \label{fig:comparison_pipeline}
\end{figure*}

%% file: sec/4_experiments.tex
\section{Experimental setup}
\label{sec:experiments}
\subsection{Training setting}
Here we overview the training setting for the selector, details in \Cref{app:experimental_details}.

\myparagraph{Selector training data}
A selector is tasked with mapping a conversation tuple \{Q, I, R, A\} into a confidence scalar value \(c_{\mathrm{sel}} \in [0,1]\).
Therefore, we first collect training data by generating answers using Pixel-Reasoner \citep{suPixelReasonerIncentivizing2025} as the reasoner model.
Pixel-Reasoner \citep{suPixelReasonerIncentivizing2025} has been finetuned to use a zoom-in tool to improve high-resolution VQA accuracy.
As the source of these questions, we use the Thyme \citep{thyme} (high-resolution natural images) and TAT-DQA \citep{tat_dqa} (financial documents) datasets.
 All training labels can be generated automatically, as detailed in \Cref{app:overlap_and_labels}.

\myparagraph{Selector model}
We adopt a compact multimodal instruction-tuned model, specifically Gemma-3-4b-it \citep{gemma3_2025}, as the selector.
Being smaller than the reasoner(s), this allows us to demonstrate that by using a reasoner that provides visual evidence, the selector can be made smaller and hence more efficient. 
We add three separate value heads, one for each of the confidence scores \(c_{corr}, c_{loc}, c_{coh}\).
These value heads project the final hidden state of the last sequence element to a scalar value, so predicting several confidence scores adds no overhead to the standard single confidence selector.

\subsection{Evaluation}
We evaluate on five complementary OOD benchmarks summarized in \Cref{tab:benchmarks}, spanning fine-grained high-resolution perception, diagrams and tables, remote sensing, blind-user mobile photos, and human-adversarial visual QA.
This diversity stresses different failure modes and provides varied operating points for selective prediction, unlike traditionally used VQA benchmarks which tend to be in-distribution for recent models.
Some benchmarks explicitly benefit from zooming, while others such as VizWiz and AdVQA do not strictly require it.
We include both regimes because our goal is reliable selective prediction in diverse high-stakes OOD settings where abstention matters.

We report coverage at risk (C@r) at risk levels $r \in \{1,5,10,15,20,25,30\}\%$, together with the average $\overline{\text{C@r}}$ across those levels.
C@r measures how much of the data can be answered while keeping the error rate below a target risk level.
For small benchmarks, we generate up to five answers per question and compute metrics over the pooled question--answer pairs to reduce noise from answer generation.
We also report area under the risk--coverage curve (AURC), a more global summary of ranking quality across the full operating range.
For readability, \Cref{tab:main_results} reports the lowest risk level at which at least one method reaches double-digit coverage; \Cref{app:additional_results} contains results at all considered risk levels.
Additional benchmark details, repeated-answer evaluation protocol, and metric definitions are provided in \Cref{app:ood_benchmarks,app:selective_metrics}.

\input{tables/draft/main_results_v2.tex}

\subsection{Baselines}
In this paper, we explore whether (i) having a reasoner which localizes the relevant visual evidence by using a zooming tool helps reliability, and (ii) having a selector that explicitly evaluates the grounding of the answer improves estimating answer correctness.
Therefore, our baselines are chosen to test both of these hypotheses.
Firstly, we run our reasoners (Pixel-Reasoner, o3, and Gemini-3-Pro) with and without the zoom-in tool, and show whether the same abstention methods can answer more reliably when provided visual evidence.
Secondly, we compare SIEVES with various abstention methods.
 \Cref{tab:related_work_comparison} in the appendix summarizes the differences between these methods.

 \myparagraph{Intrinsic reasoner confidence}
We extract the log-probability of the final message, i.e. the last reasoning step and the final answer, and use it as the confidence score \citep{yoshikawa-okazaki-2023-selective,chen-etal-2023-adaptation}.
We experiment with different sections of the conversation to extract the log-probability from, and find this to perform best in general.
Notably, log-probabilities are not available for frontier proprietary models, so we cannot use this baseline for our experiments with o3 or Gemini-3-Pro.

\myparagraph{Zero-shot selector baselines}
Given the strong zero-shot capabilities of recent VLMs, we test whether they can estimate answer correctness without task-specific training.
Therefore we present two zero-shot baselines which repurpose an existing VLM to produce confidence scores.
In the zero-shot \textit{logit} selector, the model is asked to think step by step and then output yes or no to express whether it thinks the answer is correct.
We extract the log-probability of the yes token (normalized by the sum of the yes and no tokens) and use that as confidence score \citep{bhatt2024taughttoknow,varshney-baral-2023-post}.
Alternatively, we ask the zero-shot \textit{verbalized} selector to output a numerical confidence from 0 to 100 using tokens. %
For conciseness, \Cref{tab:main_results} only reports these baselines for the stronger setting where the answering model is required to provide visual evidence zero-shot.
Additional results answering using plain language CoT are provided in \Cref{app:additional_results}, where we show it achieves considerably lower coverage and AURC.

\myparagraph{Selection without localization estimation}
Lastly, we compare to a selector trained to predict only answer correctness.
This is our trained selector baseline: it uses the same Gemma-3-4B architecture, observable reasoner inputs, and single selector call as SIEVES, but is supervised only for answer correctness.
Thus, the selector and SIEVES are directly compute matched at inference; the additional localization and coherence value heads are scalar projections of the same final hidden state and add negligible overhead.
Notably, we run two versions of this baseline.
One is trained on answers generated by a reasoner without access to zoom-in tools, where no localization evidence is provided.
In this case the selector must rely on the reasoner's own textual reasoning to infer confidence, or solve the task again to estimate answer correctness.
The other version is trained on answers generated by a reasoner with access to zoom-in tools, where localization evidence is provided.

%% file: tables/draft/main_results_v2.tex
\begin{table}[!tb]
  \centering
  \setlength{\tabcolsep}{1.0pt}
  \resizebox{\textwidth}{!}{%
  \begin{tabular}{l cccc cccc cccc cccc cccc}
  \toprule
  & \multicolumn{4}{c}{V* Bench} & \multicolumn{4}{c}{HR-Bench-8k} & \multicolumn{4}{c}{MME-RW-L} & \multicolumn{4}{c}{VizWiz} & \multicolumn{4}{c}{AdVQA} \\
\cmidrule(lr){2-5}\cmidrule(lr){6-9}\cmidrule(lr){10-13}\cmidrule(lr){14-17}\cmidrule(lr){18-21}
Method & C@5 & $\overline{C@r}$ & AURC$\downarrow$ & Acc & C@1 & $\overline{C@r}$ & AURC$\downarrow$ & Acc & C@5 & $\overline{C@r}$ & AURC$\downarrow$ & Acc & C@20 & $\overline{C@r}$ & AURC$\downarrow$ & Acc & C@10 & $\overline{C@r}$ & AURC$\downarrow$ & Acc \\
  \midrule
  \multicolumn{21}{l}{\textbf{\emph{Pixel-Reasoner w/o localization}}} \\
  Logprobs & 0.2 & 52.7 & 14.7 & \multirow{2}{*}{79.6} & 0.0 & 6.2 & 30.3 & \multirow{2}{*}{65.6} & \underline{0.1} & 8.3 & 35.3 & \multirow{2}{*}{51.6} & 0.0 & 0.1 & 48.8 & \multirow{2}{*}{38.1} & 0.1 & 0.3 & 42.5 & \multirow{2}{*}{54.5} \\
  Selector & \underline{2.3} & \underline{54.6} & 15.7 &  & \underline{0.9} & 30.8 & 20.3 &  & 0.0 & \underline{17.8} & \textbf{30.4} &  & \underline{1.6} & \underline{3.5} & 45.6 &  & 0.0 & \underline{7.7} & 33.7 &  \\
  \midrule
  \multicolumn{21}{l}{\textbf{\emph{Pixel-Reasoner w/ localization}}} \\
  Logprobs & 0.1 & 50.2 & 15.7 & \multirow{5}{*}{80.5} & 0.3 & 28.9 & 22.6 & \multirow{5}{*}{68.4} & 0.0 & 0.7 & 43.5 & \multirow{5}{*}{51.3} & 0.1 & 0.2 & 51.3 & \multirow{5}{*}{38.0} & 0.0 & 0.0 & 40.3 & \multirow{5}{*}{58.1} \\
  Zero-shot verb. & 0.0 & 42.9 & 17.2 &  & 0.0 & 13.3 & 28.9 &  & 0.0 & 0.0 & 43.5 &  & 0.0 & 0.0 & 50.8 &  & 0.0 & 0.0 & 40.7 &  \\
  Zero-shot logit & 0.0 & 47.1 & 15.6 &  & 0.0 & 32.8 & 21.3 &  & 0.0 & 4.7 & 36.4 &  & 0.0 & 0.0 & 53.1 &  & 0.0 & 0.0 & 38.6 &  \\
  Selector & 1.4 & 53.1 & \underline{14.1} &  & 0.0 & \textbf{42.7} & \textbf{17.4} &  & 0.0 & 16.9 & 31.9 &  & 0.1 & 1.5 & \underline{44.2} &  & \underline{0.2} & 6.3 & \underline{33.0} &  \\
  SIEVES (ours) & \textbf{9.7} & \textbf{59.5} & \textbf{12.8} &  & \textbf{2.3} & \underline{41.1} & \underline{18.1} &  & \textbf{4.0} & \textbf{18.7} & \underline{31.0} &  & \textbf{7.6} & \textbf{6.6} & \textbf{42.3} &  & \textbf{2.8} & \textbf{9.3} & \textbf{32.2} &  \\
  \midrule
  \midrule
  \multicolumn{21}{l}{\textbf{\emph{o3 w/o localization}}} \\
  Selector & 0.1 & 21.8 & 23.4 & 71.8 & 0.3 & 33.3 & 19.2 & 72.1 & \underline{1.1} & 13.4 & 29.3 & 56.9 & \underline{6.3} & \underline{3.8} & 40.9 & 45.8 & 0.0 & 19.4 & 25.2 & 68.0 \\
  \midrule
  \multicolumn{21}{l}{\textbf{\emph{o3 w/ localization}}} \\
  Zero-shot verb. & 0.0 & 64.7 & \underline{9.1} & \multirow{4}{*}{85.9} & 0.0 & 54.2 & 13.5 & \multirow{4}{*}{82.0} & 0.0 & 0.0 & 36.1 & \multirow{4}{*}{58.7} & 0.0 & 0.0 & 41.6 & \multirow{4}{*}{46.9} & 0.0 & 27.5 & 24.9 & \multirow{4}{*}{72.4} \\
  Zero-shot logit & \underline{1.7} & \underline{65.1} & 9.9 &  & 0.0 & 56.3 & 13.1 &  & 0.0 & 4.8 & 35.3 &  & 0.0 & 0.0 & 39.0 &  & 0.0 & 30.6 & 21.9 &  \\
  Selector & 0.0 & 61.0 & 9.7 &  & \underline{1.0} & \underline{61.7} & \underline{10.5} &  & 0.1 & \underline{16.6} & \underline{28.6} &  & 0.0 & 3.7 & \underline{35.2} &  & \underline{3.8} & \underline{33.3} & \underline{20.1} &  \\
  SIEVES (ours) & \textbf{23.1} & \textbf{70.8} & \textbf{8.4} &  & \textbf{8.0} & \textbf{67.7} & \textbf{9.3} &  & \textbf{1.7} & \textbf{17.5} & \textbf{28.4} &  & \textbf{11.7} & \textbf{12.4} & \textbf{33.3} &  & \textbf{8.2} & \textbf{36.5} & \textbf{19.6} &  \\
  \midrule
  \midrule
  \multicolumn{21}{l}{\textbf{\emph{Gemini-3-Pro w/o localization}}} \\
  Selector & 8.8 & 68.4 & 8.8 & 86.4 & \underline{5.8} & 74.5 & 5.4 & 85.6 & \underline{5.1} & \textbf{34.7} & \underline{21.9} & 60.9 & \underline{3.8} & \underline{4.2} & 41.9 & 44.1 & 15.0 & 55.0 & 12.8 & 80.3 \\
  \midrule
  \multicolumn{21}{l}{\textbf{\emph{Gemini-3-Pro w/ localization}}} \\
  Zero-shot verb. & 88.3 & 84.0 & \underline{3.1} & \multirow{4}{*}{94.3} & 0.0 & 71.4 & 6.4 & \multirow{4}{*}{90.1} & 0.0 & 9.3 & 28.2 & \multirow{4}{*}{63.2} & 0.0 & 0.0 & 46.2 & \multirow{4}{*}{43.6} & 0.0 & 42.9 & 17.0 & \multirow{4}{*}{81.4} \\
  Zero-shot logit & \underline{93.5} & \underline{84.8} & \textbf{2.9} &  & 1.4 & 79.9 & 4.9 &  & 0.0 & 19.5 & 26.4 &  & 0.0 & 0.0 & 47.7 &  & 0.0 & 42.9 & 17.3 &  \\
  Selector & 40.7 & 77.6 & 3.5 &  & 4.5 & \underline{80.1} & \textbf{3.5} &  & 1.2 & 28.2 & 22.8 &  & 0.0 & 2.7 & \underline{39.5} &  & \underline{32.5} & \underline{58.9} & \underline{11.9} &  \\
  SIEVES (ours) & \textbf{95.0} & \textbf{85.6} & \underline{3.1} &  & \textbf{14.4} & \textbf{80.8} & \underline{4.4} &  & \textbf{12.2} & \underline{32.3} & \textbf{21.5} &  & \textbf{8.1} & \textbf{8.7} & \textbf{37.2} &  & \textbf{33.3} & \textbf{59.8} & \textbf{11.5} &  \\
  \bottomrule
  \end{tabular}%
  }%
\vspace{5pt}
\caption{
  \textbf{Selective prediction across reasoners and OOD benchmarks.}
  C@$r$: coverage at risk; $\overline{C@r}$: average coverage across risk levels \{1,5,10,15,20,25,30\}\%;
  AURC; Acc: answer accuracy (\%), per reasoner, equal across abstention methods.
  ``Selector'' is the compute-matched trained correctness-only selector (one Gemma-3-4B call).
  SIEVES is trained on Pixel-Reasoner answers only; and generalizes to o3 and Gemini-3-Pro and these OOD benchmarks.
  \textbf{Bold} indicates best result per reasoner, \underline{underlined} second best.
}
  \label{tab:main_results}
\vspace{-20pt}
\end{table}

%% file: sec/5_results.tex
\section{Results}
\label{sec:results}
\input{figs/qual_example_coherence.tex}
\myparagraph{Visual evidence facilitates answer selection}
Requiring localization for every answer simplifies selective prediction by producing explicit visual evidence.
Across our benchmarks, we observe higher coverage and lower AURC in \Cref{tab:main_results} when Pixel-Reasoner has access to the zoom-in tool and provides visual evidence.
Notably, although zooming in has been predominantly used as means to improve answer accuracy \citep{suPixelReasonerIncentivizing2025,zhengDeepEyesIncentivizingThinking2025}, we observe here that even when accuracy improvements are modest, the coverage gains obtained by the selector having access to visual evidence are much greater.
This is most evident when comparing $C@10$ for Gemini-3-Pro on AdVQA.
Even for the same standard selector, the coverage doubles from 15.0 to 32.5 with a mere 1.1\% accuracy improvement.
For SIEVES, this is even more pronounced, going from 1.6 C@20 for the correctness-only selector without localization in VizWiz, to 7.6 when using SIEVES, even though the reasoner accuracy actually decreases from 38.1 to 38.0.
Importantly, to fully exploit the visual evidence produced by the reasoner, the selector must explicitly estimate the localization quality of each answer. 
This is evidenced by two correctness-only selectors with identical architectures but different training data: 
the selector trained on chains from the localized, tool-augmented answerer occasionally performs worse than the one trained on chains from the non-localized answerer, e.g.\ on MME-RealWorld-Lite and V* Bench (\Cref{tab:main_results}).
Adding visual evidence scoring reverts this trend.
 At the risk levels reported in \Cref{tab:main_results}, SIEVES exceeds the compute-matched selector in all 15 reasoner--benchmark settings with localization, while using the same budget of one reasoner call and one Gemma-3-4B selector call.
It also improves average coverage in 14/15 settings and AURC in 13/15 settings.

However, when explicitly looking at the generalization to a stronger reasoner, o3 and Gemini-3-Pro in \Cref{tab:main_results}, we observe that the selector which receives answers with localization transfers more effectively.
This further confirms that visual evidence helps generalizing to OOD reasoners, as selectors can overfit textual expressions of confidence which are model-specific.

\myparagraph{Localization estimation enables generalization}
Estimating localization quality alongside answer correctness can maximally exploit the localization evidence provided by the reasoner, further improving the generalization beyond correctness-only selectors.
By explicitly requiring that the selector evaluates the grounding of the answer, the selector can discount overconfident but weakly grounded answers and learn to prefer well-grounded ones at the same expressed reasoner confidence.
\Cref{tab:main_results} showcases SIEVES as the most consistent method, offering improved C@r and AURC.

\Cref{fig:qual_example_coherence} shows high-stakes qualitative examples where, by explicitly scoring the quality of visual evidence, SIEVES correctly accepts or abstains when the implicit selector fails.

Notably, zero-shot selectors perform very poorly in general.
We hypothesize that, although these models are strong zero-shot answerers, estimating comparable confidence scores across the entire distribution of question-answer pairs is a challenging task, for which they have not been trained.

\myparagraph{SIEVES selector trained on Pixel-Reasoner generalizes out-of-the-box to o3 and Gemini-3-Pro}
An advantage of our design, which takes only model outputs and not model internals like hidden activations or log-probabilities, is that it can be applied to any reasoner.
This includes frontier models served through proprietary APIs, which generally do not provide access to such internal signals.
\Cref{tab:main_results} confirms that this selector design can generalize to stronger reasoners, such as o3 and Gemini-3-Pro.
We perform no adaptation for the o3 or Gemini-3-Pro reasoners.
They receive the same prompt as the Pixel-Reasoner, and we use the same selector checkpoints which are trained and tuned for abstaining on Pixel-Reasoner answers.
Not only does the coverage at relevant risk levels not decrease under the change of reasoner, but it significantly increases in most cases.
When selecting among these answers provided by o3 or Gemini-3-Pro with access to tools, the SIEVES selector shows an advantage over the correctness-only selector.
\Cref{fig:ood_avg_cov_panels} highlights that at low risk tolerances, SIEVES consistently achieves highest coverage, often by large margins.

The accuracy results indicate that both zoom-in tool use and stronger reasoners tend to yield higher answer correctness.
It is worth noting that the increase in coverage that our SIEVES selector provides is beyond the increase in accuracy provided by these stronger reasoner settings.
This means that SIEVES is able to exploit the improved visual evidence provided by the stronger reasoner with access to zoom-in tools.
When analyzing the behaviors of the reasoners with tools on V* Bench, we observe that o3 zooms in more (3.0 crops on average vs 1.1) and better (85\% of answers have mIoGT $\geq 0.75$ vs 37\%).
\Cref{app:localization_target} analyzes this, including qualitative analysis of correct and incorrect localization and answering.
This illustrates another generalization dimension, where o3 and Gemini-3-Pro generate significantly higher number of crops on average, since Pixel-Reasoner was explicitly trained to zoom-in just once, and yet our selector trained on Pixel-Reasoner can generalize to o3 and Gemini-3-Pro.
Similarly, the correctness-only selector is able to generalize better to o3 and Gemini-3-Pro when observing localization evidence, which was not the case with the in-domain Pixel-Reasoner answers.
This can be explained by the fact that, when presented only textual reasoning, the selector often relies on the verbally expressed confidence from the reasoner, which varies across reasoners.
In contrast, the visual evidence can be directly assessed by the selector, and better visual evidence further helps this selector to achieve higher coverage.
SIEVES also improves selective prediction when the abstention threshold is fixed at inference, as shown in \Cref{tab:fixed_threshold_transfer}.

\myparagraph{SIEVES improves with each of the fine-grained evidence scoring losses}
\Cref{tab:weight_ablation_id_ood} ablates the weights $\lambda_{corr}, \lambda_{loc}, \lambda_{coh}$.
The main paper reports the Pixel-Reasoner ablations, while the corresponding o3 and Gemini-3-Pro results are in \Cref{tab:ablation_o3_g3p}.
Including all objectives is better than excluding any, and the framework is robust to changes in the weights.
\Cref{tab:coherence_ablation_id_ood} varies the VLM used to annotate crop-image coherence during training.
We find that a larger labeler offers little benefit, whereas a smaller model (here the selector's base model) can also be used with a modest decrease in performance.
\Cref{tab:bbox_ablation_id_ood} shows that automatic bounding-box annotation with Gemini-3-Flash enables scalable SIEVES training.

\input{tables/draft/combined_ablation_id_ood.tex}

\input{tables/supp/fixed_threshold_transfer.tex}

\myparagraph{SIEVES improves consistency-based test-time scaling}
Sampling several answers and measuring their agreement uses the answerer's knowledge, but repeating long multimodal reasoning is expensive.
SIEVES without test-time scaling provides competitive selective prediction at about one-fifth the cost of five-sample consistency (\Cref{tab:consistency_cost_rebuttal}).
Weighting each answer's contribution to agreement using SIEVES improves over both standard consistency and correctness-selector-weighted consistency, for only 0.3\% extra cost.

\input{tables/draft/consistency_cost_rebuttal.tex}
\FloatBarrier

%% file: figs/qual_example_coherence.tex
\begin{figure*}[tb]
    \centering
    \begin{minipage}[t]{0.48\textwidth}
        \centering
        \includegraphics[width=\textwidth]{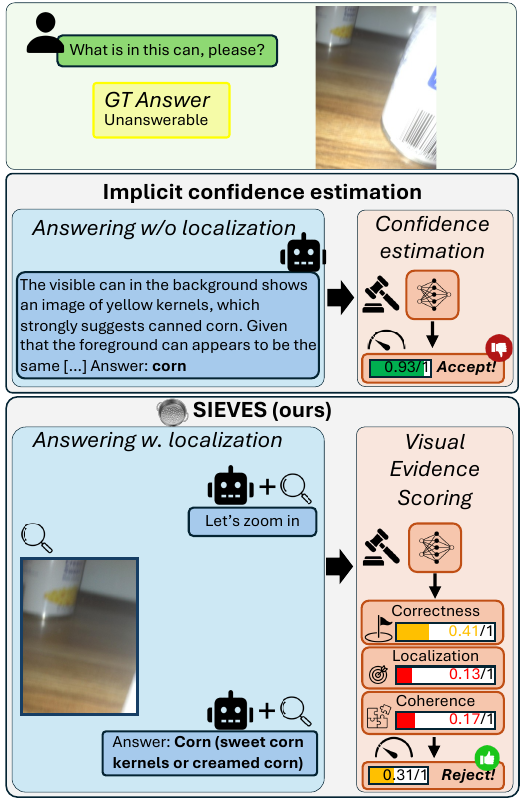}
    \end{minipage}
    \hfill
    \begin{minipage}[t]{0.48\textwidth}
        \centering
        \includegraphics[width=\textwidth]{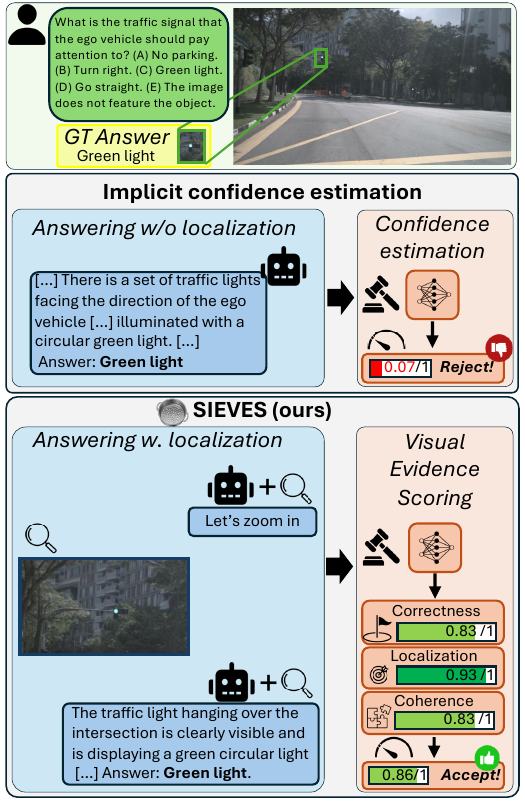}
    \end{minipage}\vspace{-0.2cm}
    \caption{\textbf{Examples where visual-evidence scores correct a selector error.} \textbf{Left:} SIEVES abstains because o3's crop shows a background can and the VizWiz image does not support the requested foreground label. \textbf{Right:} SIEVES accepts Gemini-3-Pro's correct answer because its crop shows the green traffic light.}
    \label{fig:qual_example_coherence}
\vspace{-15pt}
\end{figure*}

%% file: tables/draft/combined_ablation_id_ood.tex
\begin{table*}[tb]
\centering
\newcommand{\hlcell}[1]{\cellcolor{gray!15}\raisebox{-0.1ex}{#1}}
\newcommand{\hlnum}[1]{\cellcolor{gray!15}{\vphantom{Ay}#1}}
\captionsetup[subtable]{skip=1pt}
\caption{\textbf{Ablations.} We study modifications to the {\setlength{\fboxsep}{1pt}\protect\colorbox{gray!15}{\strutless default SIEVES configuration}}, reporting AURC and average coverage ($\overline{\mathrm{C@r}}$). The OOD results are averaged over 5 benchmarks, Thyme MC holdout set is ID. This table reports Pixel-Reasoner; \Cref{tab:ablation_o3_g3p} reports o3 and Gemini-3-Pro. (a) The default gives the lowest OOD AURC and ties for the highest mean coverage. (b) We compare coherence labels from three VLMs. (c) Automatic boxes give results close to reference boxes. \textbf{Bold} indicates best result, \underline{underlined} second best.}
\label{tab:ablation_id_ood}
\begin{subtable}[t]{0.52\textwidth}
\centering
\caption{\textbf{Objective weight ablation.}}
\label{tab:weight_ablation_id_ood}
\setlength{\tabcolsep}{3pt}
\resizebox{\textwidth}{!}{%
\renewcommand{\arraystretch}{0.445}
\begin{tabular}{rrr rr rr}
\toprule
 & & & \multicolumn{2}{c}{ID} & \multicolumn{2}{c}{OOD} \\
\cmidrule(lr){4-5}\cmidrule(lr){6-7}
$\lambda_{c}$ & $\lambda_{l}$ & $\lambda_{h}$ & $\overline{C@r}\uparrow$ & AURC$\downarrow$ & $\overline{C@r}\uparrow$ & AURC$\downarrow$ \\
\midrule
1.0 & 0.0 & 0.0 & \textbf{26.7} & \underline{25.8}& 24.1 & 28.1  \\
0.0 & 1.0 & 0.0 & 3.0 & 38.4& 17.3 & 33.0  \\
0.0 & 0.0 & 1.0 & 9.1 & 34.9& 21.4 & 29.9  \\
0.6 & 0.4 & 0.0 & 24.3 & 27.0& \textbf{27.0} & \underline{27.4}  \\
0.6 & 0.0 & 0.4 & 21.4 & 28.0& 25.2 & 27.6  \\
\cmidrule(l){1-7}
\hlnum{0.6} & \hlnum{0.3} & \hlnum{0.1} & \hlnum{\underline{26.6}} & \hlnum{\textbf{25.7}}& \hlnum{\textbf{27.0}} & \hlnum{\textbf{27.3}}  \\
0.6 & 0.2 & 0.2 & 23.5 & 27.0& \underline{25.6} & 27.5  \\
0.3 & 0.3 & 0.3 & 19.4 & 29.6& 24.2 & 28.9  \\
0.2 & 0.4 & 0.4 & 14.1 & 33.1& 21.9 & 30.2  \\
\bottomrule
\end{tabular}%
}%
\end{subtable}
\hfill
\begin{minipage}[t]{0.45\textwidth}
\centering
\begin{subtable}[t]{\textwidth}
\centering
\caption{\textbf{Crop-answer coherence annotator.}}
\vspace{-2pt}
\label{tab:coherence_ablation_id_ood}
\setlength{\tabcolsep}{2pt}
\resizebox{\textwidth}{!}{%
\begin{tabular}{l rr rr}
\toprule
 & \multicolumn{2}{c}{ID} & \multicolumn{2}{c}{OOD} \\
\cmidrule(lr){2-3}\cmidrule(lr){4-5}
Model & $\overline{C@r}\uparrow$ & AURC$\downarrow$ & $\overline{C@r}\uparrow$ & AURC$\downarrow$ \\
\midrule
\hlcell{Qwen 2.5 VL 7B} & \hlcell{\underline{26.6}} & \hlcell{\underline{25.7}}& \hlcell{\textbf{27.0}} & \hlcell{\textbf{27.3}}  \\
Gemma 3 4B & 24.4 & 27.1& 25.5 & 28.1  \\
Qwen 2.5 VL 72B & \textbf{26.8} & \textbf{25.5}& 25.5 & \underline{27.8}  \\
\bottomrule
\end{tabular}%
}%
\end{subtable}
\vspace{1.4em}
\begin{subtable}[t]{\textwidth}
\centering
\vspace{4pt}
\caption{\textbf{Loc. bounding-box annotator.}}
\vspace{-2pt}
\label{tab:bbox_ablation_id_ood}
\setlength{\tabcolsep}{2pt}
\resizebox{\textwidth}{!}{%
\begin{tabular}{l rr rr}
\toprule
 & \multicolumn{2}{c}{ID} & \multicolumn{2}{c}{OOD} \\
\cmidrule(lr){2-3}\cmidrule(lr){4-5}
BBox annot. & $\overline{C@r}\uparrow$ & AURC$\downarrow$ & $\overline{C@r}\uparrow$ & AURC$\downarrow$ \\
\midrule
\hlcell{Human} & \hlcell{26.6} & \hlcell{25.7}& \hlcell{27.0} & \hlcell{27.3}  \\
Gemini-3-Flash & 24.2 & 27.5& 26.3 & 27.6  \\
\bottomrule
\end{tabular}%
}%
\end{subtable}
\end{minipage}
\vspace{-10pt}
\vspace{-10pt}
\end{table*}

%% file: tables/supp/fixed_threshold_transfer.tex
\begin{table}[tb]
\centering
\caption{\textbf{SIEVES improves reliability with fixed abstention thresholds transferred from a small held-out set.}
We choose thresholds on held-out Thyme-MC to maximize effective reliability $\Phi_2=(1\cdot N_{\rm correct}-2\cdot N_{\rm wrong}+0\cdot N_{\rm abstain})/N$ and transfer them unchanged to five OOD benchmarks answered by o3.
Each correct answer earns $+1$, each mistake costs $2$, and each abstention contributes $0$; $N$ is the total number of questions.}
\label{tab:fixed_threshold_transfer}
\setlength{\tabcolsep}{4pt}
\resizebox{\linewidth}{!}{%
\begin{tabular}{lrrrrrr}
\toprule
Method & V* & HR-Bench & MME-RW-L & VizWiz & AdVQA & Mean $\Phi_2$ \\
\midrule
Selector & 6.49 & 10.42 & 5.54 & 1.94 & 8.40 & 6.56 \\
SIEVES & \textbf{15.60} & \textbf{17.93} & \textbf{7.04} & \textbf{6.25} & \textbf{14.26} & \textbf{12.22} \\
\bottomrule
\end{tabular}}
\end{table}

%% file: tables/draft/consistency_cost_rebuttal.tex
\begin{table}[tb]
\centering\small
\caption{\textbf{SIEVES improves consistency-based abstention at the same five-sample inference budget.}
SIEVES without test-time scaling provides competitive reliability at lower cost and latency; SIEVES-weighted consistency helps further for only 0.3\% extra cost.
Results average over available reasoners from Pixel-Reasoner, o3, and Gemini-3-Pro; \Cref{tab:consistency_baseline} gives the per-reasoner breakdown.
R and S denote reasoner and trained selector calls, and Gen.\ S denotes a generative selector.
Costs are USD per 100 questions, and latency assumes parallel calls.
}
\label{tab:consistency_cost_rebuttal}
\label{tab:cost_latency}
\setlength{\tabcolsep}{3pt}\renewcommand{\arraystretch}{1.15}
\resizebox{\linewidth}{!}{%
\begin{tabular}{ll cc cc ccc ccc ccc ccc}
\toprule
 & & \multicolumn{2}{c}{\textbf{Cost (\$)}} & \multicolumn{2}{c}{\textbf{Latency (s)}} &
\multicolumn{3}{c}{\textbf{V* Bench}} & \multicolumn{3}{c}{\textbf{HR-Bench-8k}} &
\multicolumn{3}{c}{\textbf{MME-RW-L}} & \multicolumn{3}{c}{\textbf{VizWiz}} \\
\cmidrule(lr){3-4}\cmidrule(lr){5-6}\cmidrule(lr){7-9}\cmidrule(lr){10-12}\cmidrule(lr){13-15}\cmidrule(lr){16-18}
\textbf{Abstention method} & \textbf{Calls} & Ans. & Sel. & Ans. & Sel. &
C@5 & $\overline{C@r}$ & AURC$\downarrow$ & C@1 & $\overline{C@r}$ & AURC$\downarrow$ &
C@5 & $\overline{C@r}$ & AURC$\downarrow$ & C@20 & $\overline{C@r}$ & AURC$\downarrow$ \\
\midrule
Zero-shot logit & 1R+1 Gen.\ S & 1.845 & 0.0117 & 16.4 & 4.7 & 31.7 & 65.7 & 9.5 & 0.5 & 56.3 & 13.1 & 0.0 & 9.7 & 32.7 & 0.0 & 0.0 & 46.6 \\
SIEVES & 1R+1S & 1.845 & 0.0070 & 16.4 & 0.5 & 42.6 & 72.0 & 8.1 & 5.2 & 54.4 & 13.7 & 2.9 & 18.1 & 29.7 & 7.6 & 6.6 & 42.3 \\
Consistency & 5R & 9.826 & N/A & 28.8 & N/A & 59.4 & 77.2 & 5.2 & 0.0 & 55.9 & 12.5 & 0.0 & 4.0 & 33.0 & 0.0 & 1.6 & 45.7 \\
\quad + Selector weights & 5R+5S & 9.826 & 0.0358 & 28.8 & 0.5 & 65.8 & 79.3 & 5.5 & 11.6 & 65.1 & 10.3 & 8.6 & 26.2 & 26.0 & 7.0 & 6.8 & 42.8 \\
\quad \textbf{+ SIEVES weights} & 5R+5S & 9.826 & 0.0358 & 28.8 & 0.5 & \textbf{69.2} & \textbf{84.6} & \textbf{4.0} & \textbf{17.8} & \textbf{69.8} & \textbf{8.5} & \textbf{9.4} & \textbf{28.1} & \textbf{25.2} & \textbf{9.0} & \textbf{7.1} & \textbf{42.2} \\
\bottomrule
\end{tabular}}
\end{table}

%% file: sec/6_conclusion.tex
\section{Conclusion, limitations, and future work}
\label{sec:conclusion}
We introduce SIEVES for generalizable selective prediction in challenging visual question answering tasks.
 SIEVES requires answering models to provide visual evidence that supports their answers, i.e. image crops.
The SIEVES selector uses this evidence to assess how well localized the relevant visual elements are, and whether the crops are coherent with the reasoning and final answer.
 By explicitly scoring the quality of visual evidence, SIEVES filters out overconfident but weakly grounded answers.
On a diverse set of challenging OOD VQA benchmarks, SIEVES achieves up to six times higher coverage at low risk when compared with non-grounding baselines.
Additionally, because the SIEVES selector only consumes model outputs, not internals such as log-probabilities or hidden states, we showed that it can generalize from being trained on a weaker open-source model to evaluated using a proprietary frontier model such as o3 and Gemini-3-Pro.
Selective prediction will be increasingly important for safely automating end-to-end tasks, and we hope our work can inspire the use of visual evidence to improve reliability in visual tasks. %

%% file: sec/6_limitations.tex
\par
\label{sec:limitations}

\myparagraph{Limitations}
SIEVES improves over the correctness-only selector, but absolute coverage remains low at strict risk levels, especially for weaker reasoners and harder datasets.
This limits full automation, as a large percentage of tasks remain unsolved by the automated system, and are instead deferred to an expert or unanswered.
Additionally, our experiments cover VQA and use spatial crops as evidence.
An ideal selector would generalize to any type of evidence, like code snippets or academic references.

%% file: sec/6_future_work.tex
\par
\myparagraph{Future work}
\label{sec:future_work}
SIEVES uses image crops as evidence, but the same principle could apply wherever an answer or action can be checked against observable evidence.
Improving coverage while maintaining low risk is especially important in critical settings such as medicine and law.
In medicine, recent work assesses whether medical sources or patient information support generated claims and diagnoses using zero- or few-shot prompted LLMs, without a task-trained selector \citep{wu2025medicalreferences,ren2026medicalconfidence}.
In our visual setting, zero-shot confidence selectors perform substantially worse than our selector trained with fine-grained evidence signals, motivating trained evidence selectors for medical confidence prediction.
In law, systems already retrieve and cite legal evidence \citep{louis2024legalqa}, yet commercial legal RAG systems still hallucinate on 17--33\% of queries, including cases where a cited authority does not support the stated proposition \citep{magesh2025hallucinationfree}.
A trained evidence selector could provide higher coverage at a fixed low risk by scoring cited legal evidence and abstaining when it is weak.
In coding, current agent-safety evaluators and automated permission modules use prompted LLMs or separate model classifiers to assess whether actions are risky or authorized \citep{ruan2024agentrisks,yuan2024rjudge,ji2026permissiongate,trebacz2026autoreview,hughes2026automode}.
They do not assess whether task evidence, such as a failing test or error trace, the relevant code, and a passing test after a patch, supports the proposed edit or the need to execute a given command.
A SIEVES-style selector could be trained with fine-grained signals for the relevance and coherence of this evidence, and use them to approve an edit, request more evidence, or defer to an expert before running a catastrophic command.

%% file: sec/X_suppl.tex
\clearpage
\appendix
\setcounter{page}{1}
\renewcommand{\thepage}{A\arabic{page}}
\makeatletter
\setlength{\@fptop}{0pt}
\setlength{\@dblfptop}{0pt}
\makeatother
\begin{center}
  \Large\bfseries Supplementary Material
\end{center}
\vspace{1em}

We briefly describe how the appendix is organized.
Section~\ref{app:selective_metrics} defines the selective-prediction metrics and repeated-answer evaluation protocol.
Section~\ref{app:experimental_details} provides training details, automatic-label generation, and train--test overlap checks.
Section~\ref{app:further_analysis} studies optional zooming and broad crops.
Section~\ref{app:localization_target} studies localization targets, supervision choices, and localization-related failure modes.
Section~\ref{app:additional_results} provides baseline comparisons, cost and latency, threshold transfer, coverage at all risk levels, and additional qualitative examples.
Section~\ref{app:ood_benchmarks} describes the OOD benchmarks.
Section~\ref{app:prompt_templates} presents the annotation and judging prompts.

\section{Selective prediction metrics and repeated-answer evaluation}
\label{app:selective_metrics}

\myparagraph{Coverage at risk (C@r)}
Selective prediction seeks to maximize coverage (fraction of answered, i.e., non-abstained, questions) at a user-chosen risk level (error rate on answered examples) \citep{geifmanSelectiveClassificationDeep2017,geifman2019selectivenet}.
For sample \(i\), let \(A^i\) be the predicted answer and \(y_i\) the ground-truth answer, both represented in language space.
Let \(c_{\mathrm{sel}}^i \in [0,1]\) denote the final scalar confidence prediction for answer \(A^i\), obtained from the individual confidence scores \(\{c_{corr}, c_{loc}, c_{coh}\}\) as described in \Cref{sec:method}, and let \(\tau \in [0,1]\) be the abstention threshold.
The coverage at risk \(r\) is
\[
    \text{C@}r \triangleq \text{C@R}(r) = \max_{\tau} \; \text{C}(\tau) \quad \text{s.t.} \quad \text{R}(\tau) \le r,
\]
where
\begin{align}
    C(\tau) &= \frac{1}{N}\sum_{i=1}^N \indicator[c_{\mathrm{sel}}^i \ge \tau], \\
    R(\tau) &= \frac{\sum_{i=1}^N \indicator[A^i \ne y_i] \, \indicator[c_{\mathrm{sel}}^i \ge \tau]}{\sum_{i=1}^N \indicator[c_{\mathrm{sel}}^i \ge \tau]}.
\end{align}

\myparagraph{Area under the risk--coverage curve (AURC)}
Reporting coverage at a single risk level provides a snapshot of the performance of a confidence prediction method.
Alternatively, one can aggregate the error rate (i.e. risk) across all coverage levels, sorting samples by confidence.
The continuous version of AURC is given by:
\[
\text{AURC} = \int_{0}^{1} R(C) \, dC = \int_{\tau_\text{max}}^{\tau_\text{min}} R(\tau) \, \Bigl|\frac{dC(\tau)}{d\tau}\Bigr| \, d\tau
\]
In practice, we compute the AURC by sorting all samples by confidence, \(c_{[1]} \ge c_{[2]} \ge \dots \ge c_{[N]}\), and then computing risk at each sample (which is equivalent to each coverage level), summing over all samples, starting from the highest confidence ones.
We define the risk at the k-th ordered sample as:
\[
    R_{[k]} = \frac{1}{k} \sum_{i=1}^{k} \indicator[A^{[i]} \ne y_{[i]}]
\]
from which the discrete AURC follows:
\[
\text{AURC} = \frac{1}{N} \sum_{k=1}^{N} R_{[k]}.
\]
While AURC is a convenient single-number summary of ranking quality over the entire coverage range, if a desired risk level is known, coverage at risk can be more relevant for deployment.

\myparagraph{Repeated answers and aggregation}
As described in \Cref{sec:experiments}, we    often  sample five answers per question to    measure answer-sampling variance.
We  generate five answers    for all three reasoners on V* Bench, which has only 191 questions.
 Pixel-Reasoner    also has five runs on HR-Bench-8k, MME-RealWorld-Lite, and VizWiz; o3 has five runs  on HR-Bench-8k and    MME-RealWorld-Lite.
Other settings have one matched run due to inference cost.
Low-risk coverage on a small dataset can change because of only a few answers.
The main evaluation pools the independently sampled answer traces across runs to reduce answer-sampling noise.
As a consistency check, we also evaluate each run separately, giving one answer per question in each run.
\Cref{tab:single_answer_vstar,tab:single_answer_hrbench8k,tab:single_answer_mme_realworld,tab:single_answer_vizwiz,tab:single_answer_advqa} report the mean and sample standard deviation across runs.
These standard deviations measure answer-sampling variability on the fixed benchmark, rather than uncertainty from sampling another test set.
The pooled main results are in \Cref{tab:main_results}.
The per-run tables use uniform answer extraction from the stored model outputs.
Both trained selectors are evaluated on the full benchmarks.

To formalize the difference between aggregating and averaging over repetitions, let \(q \in \{1,\dots,M\}\) index questions and \(t \in \{1,\dots,T\}\) index sampled answers per question, with \(T=5\), confidence \(c_{\mathrm{sel}}^{q,t}\), prediction \(A^{q,t}\), and ground-truth answer \(y_q\).
   The pooled  evaluation applies the   coverage definition to the full set of \(M T\)    question--answer  pairs:
\begin{align}
    C_{\mathrm{pool}}(\tau) &= \frac{1}{M T}\sum_{q=1}^{M}\sum_{t=1}^{T}\indicator[c_{\mathrm{sel}}^{q,t} \ge \tau].
\end{align}
The pooled operating threshold is therefore
\[
    \tau^\star_{\mathrm{pool}}(r) = \arg\max_{\tau} C_{\mathrm{pool}}(\tau)
    \quad \text{s.t.} \quad R_{\mathrm{pool}}(\tau) \le r,
\]
and pooled C@r is \(C_{\mathrm{pool}}(\tau^\star_{\mathrm{pool}}(r))\).

For comparison, an \emph{averaged per-repetition} alternative evaluates each repetition $t$ separately:
\begin{align}
    C_t(\tau_t) &= \frac{1}{M}\sum_{q=1}^{M}\indicator[c_{\mathrm{sel}}^{q,t} \ge \tau_t].
\end{align}
For each repetition \(t\), we choose its own operating threshold
\[
    \tau^\star_t(r) = \arg\max_{\tau} C_t(\tau)
    \quad \text{s.t.} \quad R_t(\tau) \le r,
\]
and the averaged per-repetition summary   is
\[
    \frac{1}{T}\sum_{t=1}^{T} C_t(\tau^\star_t(r)),
\]
with    sample standard deviations reported  across the \(T\) repetitions.

\input{tables/supp/single_answer_full.tex}

\section{Experimental details}
\label{app:experimental_details}

To improve the share of answers with localization, we modify the reasoner's prompt to force it to zoom-in at least once.
Since the questions in the training data are open-ended, when there is no string match between the ground truth and predicted answer, we prompt Qwen3-8B \citep{yang2025qwen3} using the ground truth and predicted answer to evaluate whether the predicted answer is correct.

We finetune all linear layers of the backbone using LoRA \citep{hu2021lora} with rank 512.
We train with 10\% label smoothing.
We select the best checkpoint for each risk level using a multiple-choice held-out set from Thyme \citep{thyme}.

All selectors are trained for 74880 steps with batch size 8, using LLaMA-Factory \citep{zheng2024llamafactory}.
 Each selector is trained on one node with eight H100 GPUs (80 GB per GPU) and finishes in under 15 hours, or under 120 GPU-hours per run.
This is a one-time training cost; the same checkpoint is used across OOD datasets and reasoners.
 To reduce the memory footprint, we use gradient checkpointing \citep{chen2016training} and flash attention 2.0 \citep{dao2023flashattention2}.
We use cosine learning rate scheduling with a warmup period of 211 steps and a maximum learning rate of 1e-6.

Thyme is a high-resolution natural image QA, where objects are often small and localized, comprising 8083 question-image pairs.
Since no validation or test splits are provided for Thyme, we construct a holdout set by randomly choosing 750 of the questions (approximately 10\%).
TAT-DQA is a document QA dataset based on financial reports, annotated by experts, which contains 16,558 question-answer pairs, out of which 13,251 are included in the training split.
We generate 5 responses from the reasoner model to each training question.

To increase the frequency of answers with localization, we modify the reasoner's prompt to force it to zoom-in at least once, and only accept the final answer if a zoom-in crop is generated or the maximum number of turns reached, whichever comes first.
\Cref{app:optional_zoom} also evaluates optional cropping.
This guarantees that for 94\% of the questions in Thyme and 77\% in TAT-DQA, a zoom-in crop is generated, with 51\% and 31\% respectively having a mIoGT $\geq 0.75$.

For the model selection we employ Thyme's holdout set, cast into a multiple-choice (MC) format for more similarity with MC benchmarks.
To construct this MC version, we pad the ground-truth answer with an incorrect response from Pixel-Reasoner (if a wrong one was generated), and GPT-5 \citep{openai-gpt-5} generated additional distractor options, until 4 answers per question are obtained.

To generate multiple choices for the Thyme hold out set, we draw incorrect answers (hard negatives) from Pixel-Reasoner.
If these are not enough to reach 4 choices (including the ground-truth), we generate distractor options with GPT-5.
The exact prompt templates used for distractor options generation, answering with localization, correctness judging, coherence labeling, and localization annotation are shown in \Cref{app:prompt_templates}.

\subsection{Automatic labels and train--test overlap}
\label{app:overlap_and_labels}

All coherence labels are generated with Qwen2.5-VL-7B.
The supervision and loss-weight ablations use the same pooled protocol as the main evaluation.
In the fully automatic localization setting, 63.74\% of the training traces use boxes extracted from TAT-DQA document structure, and Gemini-3-Flash generates boxes for the remaining natural images.
This natural-image box generation cost about \$21 in total.
Neither labeler is called at test time.
\Cref{tab:automatic_supervision} shows that this automatic setting remains close to the setting that uses reference boxes.

We check 20,277 unique training QA pairs against 17,092 evaluation QA pairs for exact train--test overlap.
One evaluation example shares an image with training data (0.006\%).
There are no shared questions, QA pairs, or question--image pairs.
One image appears in both Thyme and MME-RealWorld-Lite, but with different questions and answers: Thyme asks for the word on the second line (``PANAMA''), while MME asks for the full ship inscription (``MSC NEW YORK PANAMA'').

\input{tables/supp/automatic_supervision.tex}

\section{   Further analysis and implementation details }
  \label{app:further_analysis}

   \subsection{Optional zooming and broad crops}
\label{app:optional_zoom}

The main setup requests at least one crop before the final answer to expose visual evidence.
This is a data-collection choice, not a requirement that every VQA question needs zooming.
We replay the stored Pixel-Reasoner traces while allowing an answer before the first crop.
The prompt and sampled model outputs are unchanged; only the stopping rule differs.
Of 36 ,   547 traces, 6,230 (17.05\%) end earlier under the optional policy.
Among the 6,183 early traces with comparable parsed answers, 997 (16.12\%) change their answer.
\Cref{tab:optional_zoom} reports selective-prediction results.

 We also test VizWiz and AdVQA examples for which the smallest returned crop covers more than 25\% or 50\% of the image.
These questions do not depend on a tiny target region.
\Cref{tab:broad_crop_scope} shows that SIEVES still improves mean coverage on both benchmarks and improves AURC in three of four comparisons.

\input{tables/supp/optional_zoom.tex}

\input{tables/supp/broad_crop_scope.tex}

\section{Localization target and threshold analyses}
\label{app:localization_target}
\label{app:ablation}

\subsection{Additional supervision choices}

\Cref{tab:additional_label_ablation} compares binary localization with continuous mIoGT regression.
Binary localization is better on the held-out ID set and the OOD mean.

For coherence label creation, we also considered judging the complete reasoning trace against all returned crops. Compared with judging the final reasoning step and answer against the best-supporting crop, this changes 8.0\% of the coherence labels across the training and validation examples, after applying the localization requirement.
The final-step variant performed better on the held-out ID set used for model choice, so we retained it.

\input{tables/supp/additional_label_ablation.tex}

 \subsection{Why IoGT instead of IoU}
\label{app:iogt_vs_iou}
The localization target should measure whether the relevant object is visible in the crop used by the reasoner.
This differs from object detection, where a predicted box is expected to tightly match the object.
   In our setting, a useful crop may intentionally include context around a small object, so its IoU with the object box can be very small even when the crop contains the full object.
 This makes IoU a poor target for deciding whether the model looked at the relevant visual evidence.

  \begin{table}[tb]
\centering
\caption{
    \textbf{IoU is highly skewed for zoom-in crops.}
    We recover IoU from the stored inverse-log IoU signal on the SIEVES training data.
    Even permissive IoU thresholds label only a small fraction of crops as localized, because crops are designed to make small objects visible rather than tightly segment them.
}
\label{tab:iou_skew}
\setlength{\tabcolsep}{6pt}
\begin{tabular}{lccc}
\toprule
Split & IoU $\geq 0.25$ & IoU $\geq 0.50$ & IoU $\geq 0.75$ \\
\midrule
Train & 3.36\% & 0.61\% & 0.04\% \\
Holdout & 1.37\% & 0.72\% & 0.00\% \\
\bottomrule
\end{tabular}
\end{table}

By contrast, IoGT directly measures object containment:
\(\text{IoGT}=1\) when the crop contains the entire ground-truth object, regardless of how much surrounding context the crop includes.
   This is the behavior we want from a  zoom-in    tool.

\subsection{Ablating threshold for binarizing localization}
\label{app:iogt_threshold_ablation}
\input{tables/draft/sr_threshold_ablation.tex}
Here, we also ablate the mIoGT threshold used to binarize the localization target, which then propagates to the coherence target:
\begin{align}
    g_{loc} &= \indicator[\text{mIoGT} \geq t], \\
    g_{coh} &= g_{loc} \cdot \max_j sim(R^j_{crop},[R_{last},A]),
\end{align}
where $t \in \{0.25, 0.50, 0.75\}$. For Pixel-Reasoner with localization, \Cref{tab:sr_threshold_ablation} compares these three choices on the Thyme tuning split and across the five OOD benchmarks.
The selected value $t=0.75$ is best in-distribution, achieving the highest average coverage and the lowest AURC.
This transfers to OOD, where it also gives the highest average coverage and the lowest AURC.

\subsection{Visual evidence quality on V* Bench}
\label{app:vstar_visual_evidence_quality}
 One possible concern  with IoGT  is that a reasoner could satisfy    the target  with a very broad crop.
We therefore analyze the actual V* Bench crops from Pixel-Reasoner and o3 with localization, using the same five repeated answers as in the main evaluation.
   \Cref{tab:vstar_crop_quality} shows that the crops are not broad: the median final crop covers only 4.74\% of the image  for o3 and 2.67\% for Pixel-Reasoner, and no final crop covers more than 25\% of the image.

  The key difference is instead  whether the crop  actually  contains the relevant object   .
For o3, the median final crop contains the entire ground-truth object, whereas Pixel-Reasoner often produces a crop that is too tight or mislocalized.

\begin{table}[tb]
\centering
\caption{
    \textbf{V* Bench visual-evidence quality.}
    We report statistics for the final valid crop on V* Bench, pooled over the five repeated answers used in the main evaluation.
    A valid crop trace has at least one nonzero-area predicted crop; its percentage is computed over all answer traces.
    ``Crop/image'' is the crop area divided by the image area.
    ``Object/crop'' is the union area of the ground-truth objects divided by the crop area.
    ``GT recall'' is the fraction of the ground-truth object covered by the crop.
}
\label{tab:vstar_crop_quality}
\setlength{\tabcolsep}{4pt}
\begin{tabular}{lccccc}
\toprule
\shortstack{Reasoner} & \shortstack{Valid crop\\traces} & \shortstack{Median\\crop/image} & \shortstack{Median\\object/crop} & \shortstack{Median\\GT recall} & \shortstack{Crop $>$25\%\\image} \\
\midrule
Pixel-Reasoner & 90.7\% & 2.67\% & 1.38\% & 54.74\% & 0.00\% \\
 o3    & 98.0\% & 4.74\% & 1.15\% & 100.00\% & 0.00\% \\
\bottomrule
\end{tabular}
\end{table}

\subsection{Failure modes and stratified localization behavior}
\label{app:localization_failure_modes}
\begin{figure}[tb]
    \centering
    \includegraphics[width=0.92\linewidth]{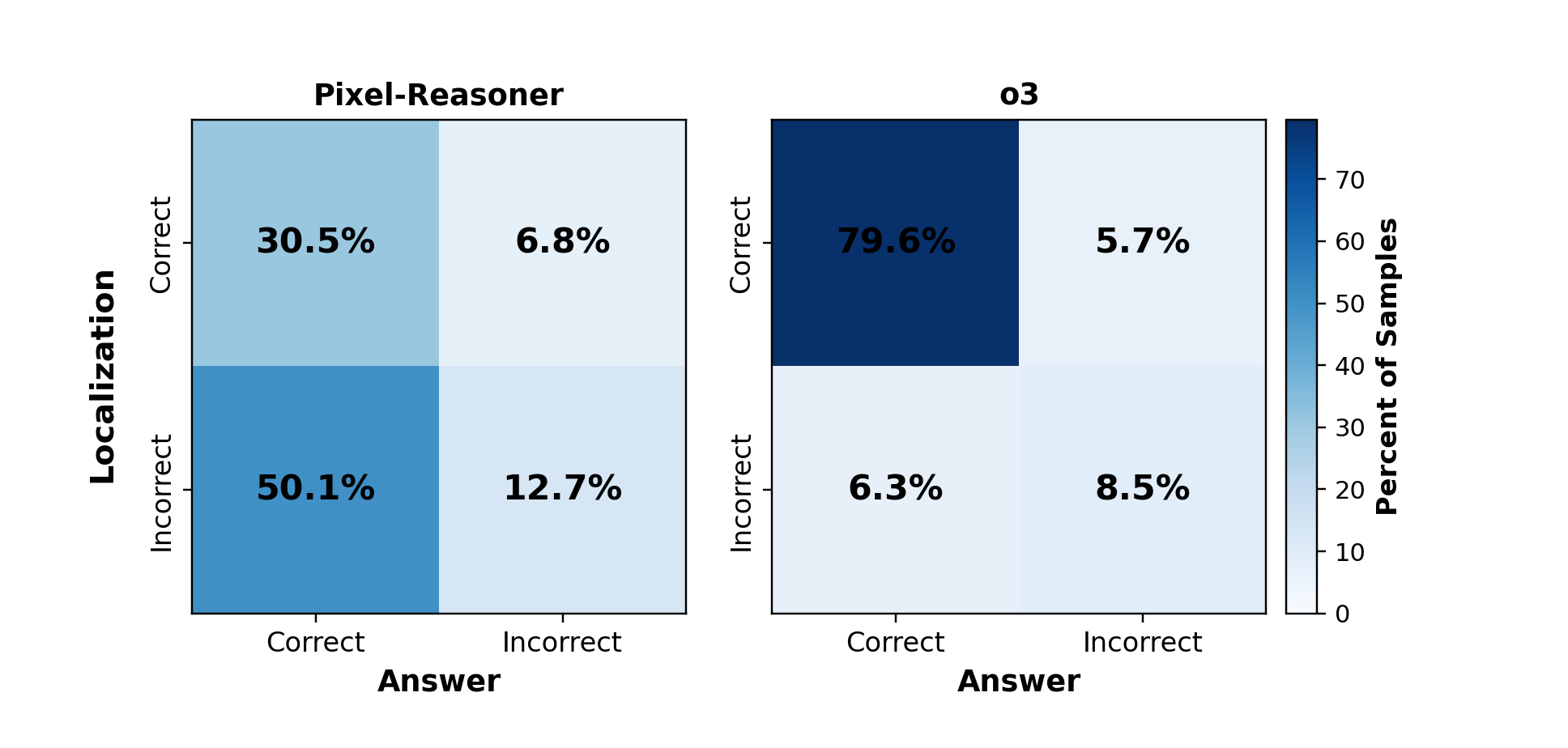}
    \caption{
        \textbf{Answer correctness vs.\ localization correctness on V* Bench.}
        We pool the five repeated answers used in the main evaluation for Pixel-Reasoner and o3 with localization.
        Localization is correct when mIoGT is at least 0.75.
        The answer-correct column sums to the corresponding V* Bench answer accuracy in \Cref{tab:main_results}.
        o3's accuracy gains coincide with substantially better localization, while Pixel-Reasoner frequently answers correctly despite weak localization evidence.
    }
    \label{fig:vstar_loc_answer_heatmap}
\end{figure}

\Cref{fig:vstar_loc_answer_heatmap} stratifies V* Bench answers by answer correctness and localization correctness.
This highlights two qualitatively different regimes.
Pixel-Reasoner often answers correctly even when the crop does not meet the localization target, indicating that answer correctness alone does not guarantee that  the    provided visual evidence is reliable.
In contrast,  o3   's improvements are concentrated in the upper-left cell: it is much more often correct while also localizing the relevant evidence.
This supports the need to score visual evidence explicitly rather than relying only on final-answer correctness.
Additional qualitative examples of incorrect localization and weak crop-answer grounding are shown in \Cref{fig:qual_example_coherence_mme}.

\begin{figure}[tb]
    \centering
    \includegraphics[width=0.92\linewidth]{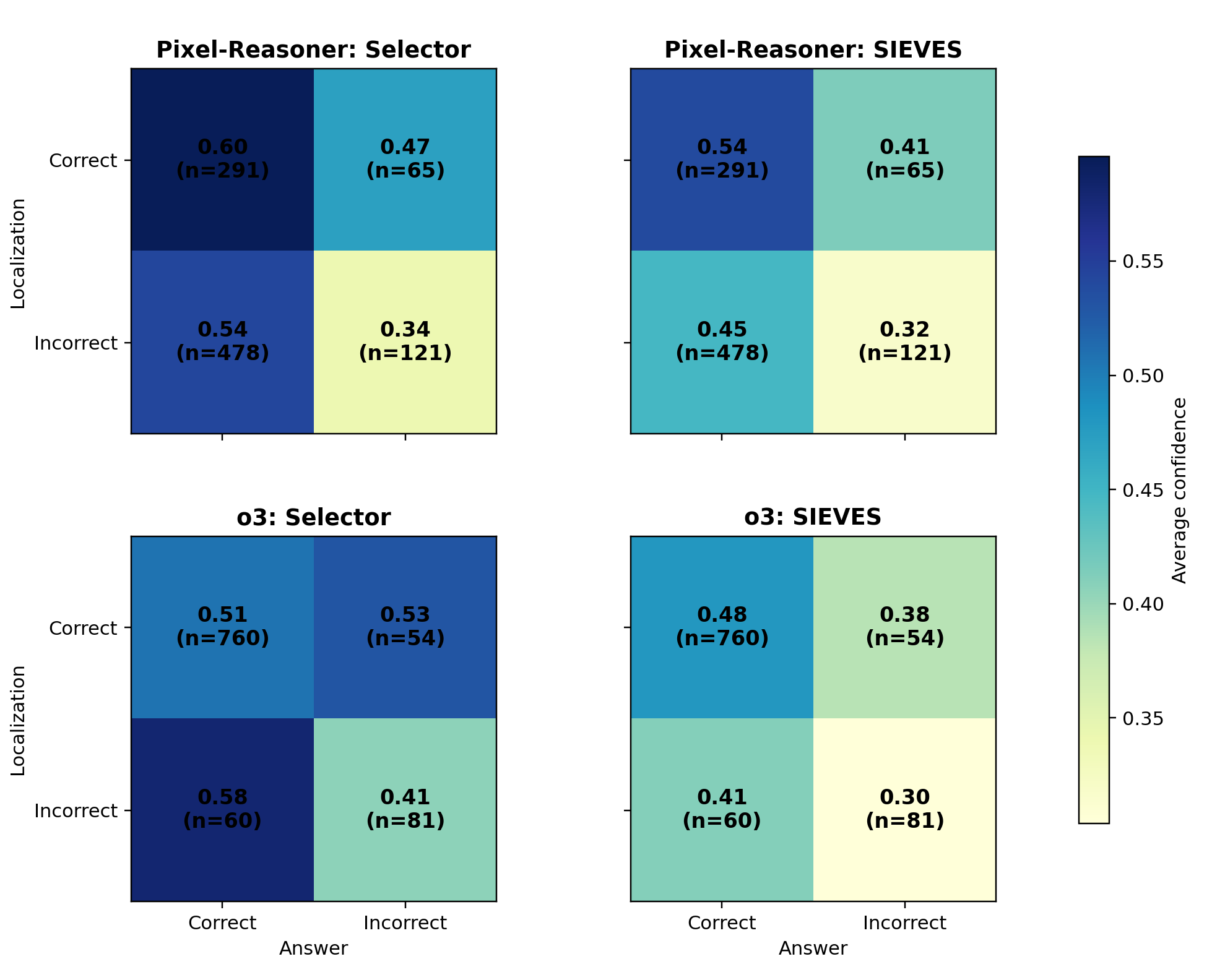}
    \caption{
        \textbf{SIEVES confidence better reflects answer and evidence quality.}
        Each cell shows mean confidence and the number of examples in parentheses.
        For o3, the correctness-only selector is more confident in incorrect answers with correct localization than in answers where both are correct.
        SIEVES instead assigns its highest confidence when both the answer and localization are correct.
    }
    \label{fig:vstar_confidence_by_loc_answer}
\end{figure}

   The answer-correct/localization-incorrect cell is not a target success case for SIEVES:
even when the final answer is correct, the trace does not provide reliable visual evidence for trusting that answer.
We nevertheless analyze this regime because it diagnoses whether a visual-evidence selector simply discards all such answers or can still retain some when the remaining evidence is strong.
\Cref{fig:vstar_confidence_by_loc_answer,tab:vstar_confidence_by_loc_answer} compare the correctness-only selector with SIEVES using the method-specific V*  C@5    checkpoints.
SIEVES assigns the highest selection rate when both the answer and localization are correct, but it can still accept some traces when only one of these signals is correct.
In contrast, traces with both incorrect answers and incorrect localization are selected least often.
This is consistent with our goal: the selector should prefer answers supported by visible evidence, without requiring answer correctness and localization correctness to be perfectly correlated.

   \input{tables/draft/vstar_confidence_by_loc_answer.tex}

  \Cref{tab:vstar_confidence_by_loc_answer_filtered} repeats the same analysis after removing traces with no localization attempt or no parseable answer.
The trend remains the same: SIEVES selects the answer-correct/localization-correct slice most often, but still accepts some traces when either the answer or the localization signal is correct, and selects the both-incorrect slice least often.
In this filtered subset,  the    o3 correctness-only selector even assigns higher average confidence to the answer-incorrect/localization-incorrect slice than to the answer-correct/localization-correct slice.

\input{tables/draft/vstar_confidence_by_loc_answer_filtered.tex}

 For completeness, \Cref{tab:ablation_o3_g3p} reports the same selector-design ablations on the two stronger OOD-only reasoners, o3 and Gemini-3-Pro.

\begin{figure}[tb]
    \centering
    \includegraphics[width=\linewidth]{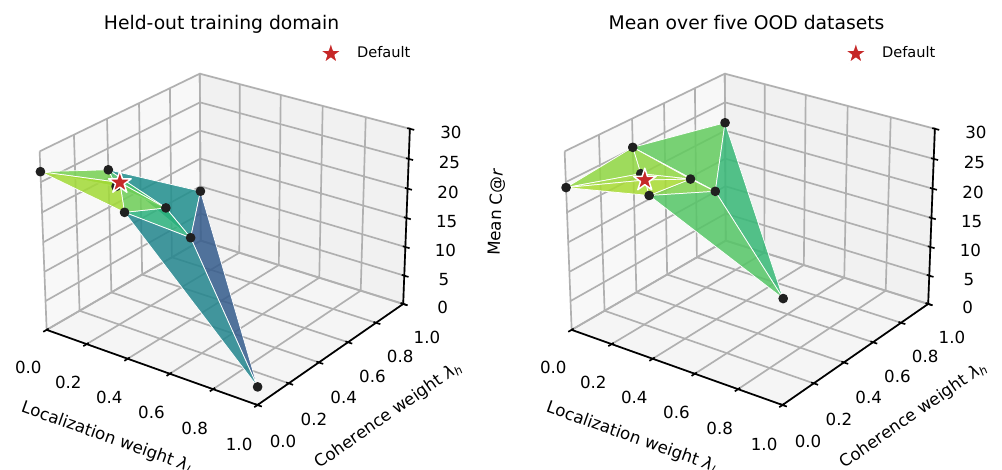}
    \caption{\textbf{Coverage across the tested objective weights.}
    Each point is one row of \Cref{tab:weight_ablation_id_ood}; the surface only helps show the trend between tested points.
    Point positions use weights normalized to sum to one.
    Several nearby mixtures perform similarly.
    The default $(\lambda_c,\lambda_l,\lambda_h)=(0.6,0.3,0.1)$ gives the best OOD AURC in the table.}
    \label{fig:objective_weight_surface}
\end{figure}

\input{tables/supp/ablation_o3_g3p.tex}

\section{Additional results}
\label{app:additional_results}
 
  \subsection{Qualitative examples}
 \Cref{fig:qual_example_coherence_mme} shows additional examples where SIEVES correctly accepts or abstains when the implicit selector fails.

\input{figs/qual_example_coherence_mme.tex}

\subsection{Baseline comparisons, cost, and latency}
\label{app:cost_and_baselines}

\input{tables/related_work_comparison.tex}

\Cref{tab:cost_latency} separates answering cost from confidence-scoring cost.
Costs are estimated from visible token counts and public prices.
Pixel-Reasoner uses the hosted Qwen2.5-7B rate as a size-based cost proxy (\$0.04/\$0.10 per million input/output tokens); this text-model rate may underestimate vision-model hosting cost.
The o3 rates are \$2/\$8, Gemini-3-Pro rates are \$2/\$12, and Gemma-3-4B rates are \$0.05/\$0.10 per million input/output tokens.
For text, we use the exact local tokenizers for open models, \texttt{o200k\_base} for o3, and about four characters per token for Gemini.
Each selector crop adds 256 input tokens; o3 uses 75 base tokens plus 150 per 512-pixel tile, and Gemini uses 1,120 tokens per image.
Multi-turn reasoners are charged for the accumulated context at every turn.
For closed models, hidden reasoning tokens are unavailable and are not counted.
Latency uses provider time-to-first-token and throughput values and assumes independent calls run in parallel.
The Gemma pooling selectors generate no output tokens.
Cost is averaged within each of the eight reasoner--benchmark settings, then across settings.

The selector has the same Gemma-3-4B backbone, visible inputs, and one-pass budget as SIEVES.
We add two black-box comparisons.
First, consistency scores five stochastic answers by their agreement, so it uses five full reasoner calls.
Second, ReCoVERR asks auxiliary questions until it finds enough evidence or reaches its budget.
\Cref{tab:recoverr_comparison} compares ReCoVERR and SIEVES on the same Pixel-Reasoner answers from V* Bench.
SIEVES has better C@5, mean coverage, and AURC at much lower estimated cost and latency.
\Cref{tab:recoverr_extended} adds completed searches across all five benchmarks using Gemma for evidence confidence.
These searches allow up to 100 auxiliary questions, keep the original answer fixed, and measure whether evidence search improves its confidence ranking.
We evaluate two ReCoVERR configurations: InstructBLIP with GPT-3.5 and Flan-T5, and Pixel-Reasoner with Gemma-3-4B.
Both gather evidence through auxiliary visual questions using the benchmark image, without external object detections or image captions.
We use the available GPT-3.5 endpoint for the first configuration.
Answers that do not meet the recovery criterion retain their direct confidence score.

\input{tables/draft/recoverr_comparison.tex}
\input{tables/supp/recoverr_extended.tex}

  \subsection{Additional baselines and more coverage levels on OOD benchmarks}
 When evaluating on OOD benchmarks, the SIEVES selector consistently shows an advantage over the correctness-only selector and other baselines.
The overall trends are generally consistent across risk levels.
We  additionally evaluate five-sample consistency-based abstention in
\Cref{tab:consistency_baseline}.
This baseline assigns each stochastic answer the frequency with which its
parsed answer occurs among five samples.
The aggregate consistency comparison averages all three reasoners on V* Bench, Pixel-Reasoner and o3 on HR-Bench-8k and MME-RW-L, and Pixel-Reasoner only on VizWiz, where it is the only reasoner with five sampled answers per question.
For Pixel-Reasoner it obtains zero coverage at every demanding risk level shown
in the main table, although its global V* Bench AURC is competitive.
For the more accurate o3 and Gemini-3-Pro reasoners it is highly competitive on
V* Bench, but it requires five full reasoner calls and remains weak at the
the risk levels reported in Table~\ref{tab:main_results} on HR-Bench and MME-RW-L.
\Cref{tab:selective_test_time_scaling} holds the five sampled answers
fixed and compares agreement-based abstention with SIEVES-weighted voting.
Weighting each answer by its SIEVES scores improves target-risk coverage, mean
coverage, and AURC in 6/7 settings.
The exception is o3 on V* Bench, where the high base accuracy and agreement
make plain consistency particularly effective.
We  also compare to a \textit{verbalized} zero-shot selector, which is asked to output a confidence score between 0 and 100 in natural language \citep{groot2024overconfidence,verbalized_confidence_triggers_self_verification_2025,on_verbalized_confidence_llms_2025}.
\Cref{tab:paper_vstar_full_baselines,tab:paper_hrbench_full_baselines,tab:paper_mme_realworld_lite_baselines,tab:paper_vizwiz_baselines,tab:paper_advqa_baselines} show coverage at all risk levels for the OOD benchmarks.
As for the ablation results, low risk levels can be less informative, e.g. in VizWiz, if the task is challenging for the answerer.
Similarly, coverage at high risk levels are saturated for V* Bench.

\input{tables/supp/baselines/consistency_baseline.tex}

\input{tables/supp/baselines/selective_test_time_scaling.tex}

 \input{tables/supp/baselines/vtsar_baselines.tex}
\input{tables/supp/baselines/hrbench_baselines.tex}
\input{tables/supp/baselines/mme_realworld_lite_baselines.tex}
\input{tables/supp/baselines/vizwiz_baselines.tex}
\input{tables/supp/baselines/advqa_baselines.tex}

\subsection{Frozen-threshold transfer}
\label{app:threshold_transfer}

C@r and AURC test how well a score ranks answers.
They do not test whether one raw threshold stays calibrated after a distribution shift.
For this stricter test, we choose one threshold for each method on 3,695 held-out Thyme-MC Pixel-Reasoner traces and apply it unchanged to five datasets answered by o3.
No target labels are used to choose the threshold.

We report effective reliability $\Phi_c=(N_{\rm correct}-cN_{\rm wrong})/N$, where an accepted correct answer is worth $+1$, an accepted error costs $c$, and an abstention is worth zero.
\Cref{tab:fixed_threshold_transfer} reports $c=2$, for which SIEVES improves all five targets.
Across costs $c\in\{1,2,4,5,10\}$, SIEVES has higher mean transferred utility at four costs; the selector is slightly better at $c=1$.
At $c=100$, both methods can have negative mean utility and the selector is better.
Thus, the fixed-threshold results support score transfer in many settings but do not establish OOD risk control.

\input{tables/supp/threshold_transfer_full.tex}

\Cref{tab:threshold_transfer_all_reasoners} also reports the aggregate over Pixel-Reasoner, o3, and Gemini-3-Pro.
At costs $c=2,4,5$, SIEVES has higher mean effective reliability than the correctness-only selector with or without localization.
\input{tables/supp/threshold_transfer_all_reasoners.tex}

  \subsection{In-distribution results}
 \input{tables/supp/baselines/thyme_750_mc_baselines.tex}
We also report the coverage at different risk levels for the hold-out sets of the training dataset in \Cref{tab:paper_thyme_750_mc_baselines}. %
The SIEVES selector achieves the best coverage at lower risk, being a close second to correctness-only selector for lower-stakes scenarios (i.e. higher risk).
However, the differences in performance are not as pronounced as in some OOD cases, again highlighting the generalization capabilities of explicitly estimating localization quality.

\section{Details on the OOD benchmarks}
\label{app:ood_benchmarks}
\input{tables/benchmarks.tex}

For VizWiz and AdVQA, we use the validation splits, since the test splits are not public and the evaluation server for the test set does not support selective prediction metrics.
These two benchmarks are open-ended, so we use a hard accuracy metric with the same LLM judge used during training.

\myparagraph{V* Bench}
\citet{wu2023v} target fine-grained perception in high-resolution images, focusing on attribute recognition (``What is the material of the glove?'') and spatial relationships (``Is the red suitcase on the left or right side of the red stool?'').
Images are 2246 $\times$ 1582 on average, sourced from SA-1B \citep{kirillov2023segment}, and relevant objects represent a small percentage of image area.

\myparagraph{HR-Bench-8k}
\citet{hrbench_wang2024} extend the V* Bench by increasing both the size and resolution by 4.
It contains 800 questions on ``Fine-grained Single-instance Perception'' (aligns with V* Bench's attribute recognition), and ``Fine-grained Cross-instance Perception'' (aligns with V* Bench's spatial relationship reasoning).
We use the 8K resolution version.
The domain of the images is broader.
Gathered from DIV8K \citep{Gu2019DIV8KD8}, questions pertain not only natural images but various domains like map and chart analysis.
This benchmark is more challenging and less sensitive to noise (due to the size) than V* Bench, while still explicitly stresses tiny-object recognition and reading fine details that only appear when zooming.

\myparagraph{MME-RealWorld-Lite}
\citet{mme_realworld_2024} focus on real-world tasks that require perceiving small or far-away elements in high-resolution imagery.
The Lite split contains 1,782 unique multiple-choice questions sampled from the full set for faster evaluation, while preserving the benchmark's breadth.
The benchmark organizes tasks into five top-level domains, Remote Sensing (RS), Autonomous Driving (AD), Monitoring (MO), Diagram and Table (DT), and Optical Character Recognition (OCR), and up to 43 sub-tasks beneath them.
Images are 2{,}000 $\times$ 1{,}500 pixels on average, with relevant objects frequently being small.
However, the difficulty of this benchmark (top model in the public leaderboard barely reaches 60\% accuracy) can be attributed mostly to the nature of the questions, which are much more out of distribution for models compared to the other common VQA benchmarks.

\myparagraph{VizWiz}
\citet{gurari2018vizwiz} aim to evaluate models on a real-world setting they are often not trained for: answering questions pertaining images taken by blind users, relating to their daily life.
These images are often blurry or even don't contain the object in question (32\% of questions in the validation set are marked as non-answerable), and the textual questions are more conversational, as they are transcriptions of enunciated questions by the blind users.
Unlike the prior three benchmarks we presented, this benchmark is open-ended.
Each visual question has ten crowd answers, and items are labeled answerable vs non-answerable.
The validation split contains 4,319 examples, of which 2,934 (68\%) are answerable.
We evaluate on validation rather than test, since test labels are not public and the official server does not support selective-prediction metrics.
   We  do not use train or validation splits for model development.
For evaluation to be consistent with that of our other benchmarks, we follow the stricter ``hard'' accuracy: we select the most frequent human answer per item and count a prediction correct only if it matches such ground truth answer, either by exact match or with the same LLM judge we use during training.
For questions which are annotated as unanswerable, any response is considered incorrect, as the optimal strategy is to abstain.
Although nominally high resolution (1224 $\times$ 1224 pixels on average), unlike in the other benchmarks, zooming in is generally not required or expected to be beneficial.
Nonetheless, the OOD nature of this benchmark, together with the high-stakes nature of the tasks, makes it a valuable testbed for our grounding-aware selective prediction framework.

\myparagraph{AdVQA}
\citet{sheng2021advqa} introduce an adversarially collected VQA benchmark where crowd workers interact with a strong model and craft valid questions that fool it.
This ``stress test'' distribution is substantially harder than prior benchmarks (e.g.\ VQAv2 \citep{goyal2017making}), and highlights brittle reasoning and spurious priors.
The validation split (v1.0\footnote{\url{https://adversarialvqa.org/download.html}}) contains 10,000 questions.
Following our setup for VizWiz described above, we use the validation split using a hard accuracy metric.
AdVQA is the largest of the benchmarks we use by total question count and is particularly useful for selective prediction because it is intentionally designed to expose failure cases where abstention should be preferred.

\clearpage
\section{Prompt templates}
\label{app:prompt_templates}

The exact prompts referenced in the experimental details are collected here for completeness.

\myparagraph{Thyme multiple-choice distractor generation prompt}
We use this prompt to convert Thyme into a multiple-choice setting by generating hard distractors conditioned on the question, the relevant crop, and optionally a wrong model answer.
In the main experiments, this chat template is queried with GPT-5.
The full message sequence is defined in \Cref{fig:supp_thyme_mc_prompt}.
\input{figs/supp_thyme_mc_prompt.tex}

\myparagraph{Reasoner with localization prompt}
To encourage explicit visual evidence, we augment the reasoner prompt with the instruction block in \Cref{fig:supp_forced_zoom_prompt}.
This tells the model to use the crop tool before giving its first final answer, so the resulting    trace  contains localization evidence .
The reasoners emit plain-text \texttt{crop\_image\_normalized} calls with \texttt{bbox\_2d} corner coordinates in $[0,1]$ and a one-based \texttt{target\_image} index.
The executor converts the box to pixels, clips it to the image bounds, and returns the crop as another image in the conversation.
The spatial-evaluation files express boxes in $[0,1000]$; these are converted coordinates for measuring overlap, rather than the coordinates requested in the reasoner's tool schema .
\input{figs/supp_forced_zoom_prompt.tex}

\myparagraph{Correctness judge prompt}
For open-ended questions, when normalized string match fails, we run the prompt in \Cref{fig:supp_correctness_judge_prompt} to decide whether the prediction should still count as correct.
This makes correctness labels more robust to valid rewordings and semantically equivalent answers.
\input{figs/supp_correctness_judge_prompt.tex}

\myparagraph{Grounding-coherence labeling prompt}
We use the prompt in \Cref{fig:supp_grounding_coherence_prompt} to create the crop-answer coherence labels used for selector training.
The judge receives the question, the reasoner's final answer, and one predicted crop, and decides whether the crop contains enough evidence and whether the answer is supported by that crop.
\input{figs/supp_grounding_coherence_prompt.tex}

\myparagraph{Localization labeling prompt}
For automatic localization annotation on Thyme, we prompt Gemini-3-Flash with the two-stage pipeline defined in \Cref{fig:supp_gemini_flash_bbox_prompt}.
The first stage extracts the target object phrase from the question, and the second localizes it in the full image.
\input{figs/supp_gemini_flash_bbox_prompt.tex}
\FloatBarrier

%% file: tables/supp/single_answer_full.tex
\begin{table}[tb]
\centering
\small
\caption{\textbf{Single-answer results on V* Bench (191 questions).} Values are mean $\pm$ sample SD across repetitions; $n=1$ rows have no SD. Trained selectors use the fixed ID-selected checkpoint for each metric.}
\label{tab:single_answer_vstar}
\label{tab:repeated_answer_pooling}
\setlength{\tabcolsep}{2pt}
\resizebox{\linewidth}{!}{%
\begin{tabular}{lcrrrrrrrrrr}
\toprule
Method & $n$ & C@1 & C@5 & C@10 & C@15 & C@20 & C@25 & C@30 & $\overline{C@r}$ & AURC & Acc. \\
\midrule
\multicolumn{12}{l}{\textbf{Pixel-Reasoner}} \\
Selector, no crops & 5 & 2.5 $\pm$ 0.9 & 2.5 $\pm$ 0.9 & 26.1 $\pm$ 17.8 & 67.2 $\pm$ 19.2 & 95.4 $\pm$ 6.5 & 100.0 $\pm$ 0.0 & 100.0 $\pm$ 0.0 & 56.2 $\pm$ 5.6 & 15.3 $\pm$ 3.6 & 79.6 $\pm$ 1.8 \\
Selector & 5 & 2.9 $\pm$ 3.1 & 8.8 $\pm$ 11.5 & 20.6 $\pm$ 8.6 & 66.6 $\pm$ 14.8 & 97.2 $\pm$ 6.0 & 100.0 $\pm$ 0.0 & 100.0 $\pm$ 0.0 & 56.6 $\pm$ 2.3 & 14.1 $\pm$ 0.8 & 80.5 $\pm$ 1.6 \\
SIEVES & 5 & 5.5 $\pm$ 3.3 & 16.3 $\pm$ 9.1 & 35.2 $\pm$ 18.3 & 77.1 $\pm$ 12.6 & 99.4 $\pm$ 1.1 & 100.0 $\pm$ 0.0 & 100.0 $\pm$ 0.0 & 61.9 $\pm$ 3.5 & 12.7 $\pm$ 1.8 & 80.5 $\pm$ 1.6 \\
ZS logit & 5 & 0.9 $\pm$ 2.1 & 2.3 $\pm$ 5.2 & 9.1 $\pm$ 10.1 & 54.0 $\pm$ 33.3 & 98.7 $\pm$ 2.5 & 100.0 $\pm$ 0.0 & 100.0 $\pm$ 0.0 & 52.2 $\pm$ 7.0 & 15.5 $\pm$ 2.7 & 80.5 $\pm$ 1.6 \\
ZS verbalized & 5 & 0.0 $\pm$ 0.0 & 0.0 $\pm$ 0.0 & 0.0 $\pm$ 0.0 & 17.6 $\pm$ 39.3 & 98.4 $\pm$ 2.2 & 100.0 $\pm$ 0.0 & 100.0 $\pm$ 0.0 & 45.1 $\pm$ 5.8 & 17.1 $\pm$ 2.0 & 80.5 $\pm$ 1.6 \\
Logprobs & 5 & 2.3 $\pm$ 1.8 & 2.3 $\pm$ 1.8 & 13.1 $\pm$ 11.8 & 52.1 $\pm$ 36.3 & 98.8 $\pm$ 1.8 & 100.0 $\pm$ 0.0 & 100.0 $\pm$ 0.0 & 52.7 $\pm$ 7.0 & 15.6 $\pm$ 2.9 & 80.5 $\pm$ 1.6 \\
\midrule
\multicolumn{12}{l}{\textbf{o3}} \\
Selector, no crops & 5 & 0.5 $\pm$ 0.5 & 0.5 $\pm$ 0.5 & 0.5 $\pm$ 0.5 & 2.1 $\pm$ 2.0 & 10.5 $\pm$ 15.3 & 46.1 $\pm$ 13.1 & 91.0 $\pm$ 20.1 & 21.6 $\pm$ 4.9 & 23.6 $\pm$ 3.5 & 71.8 $\pm$ 1.5 \\
Selector & 5 & 0.3 $\pm$ 0.5 & 1.9 $\pm$ 2.0 & 50.7 $\pm$ 24.0 & 98.4 $\pm$ 2.2 & 100.0 $\pm$ 0.0 & 100.0 $\pm$ 0.0 & 100.0 $\pm$ 0.0 & 64.5 $\pm$ 3.7 & 9.8 $\pm$ 1.4 & 85.9 $\pm$ 1.5 \\
SIEVES & 5 & 12.9 $\pm$ 8.7 & 22.5 $\pm$ 11.6 & 75.7 $\pm$ 15.9 & 98.4 $\pm$ 2.2 & 100.0 $\pm$ 0.0 & 100.0 $\pm$ 0.0 & 100.0 $\pm$ 0.0 & 72.8 $\pm$ 4.8 & 8.4 $\pm$ 1.4 & 85.9 $\pm$ 1.5 \\
ZS logit & 5 & 5.8 $\pm$ 5.2 & 12.1 $\pm$ 10.7 & 80.5 $\pm$ 3.4 & 93.3 $\pm$ 9.2 & 100.0 $\pm$ 0.0 & 100.0 $\pm$ 0.0 & 100.0 $\pm$ 0.0 & 70.2 $\pm$ 1.8 & 9.2 $\pm$ 2.7 & 85.9 $\pm$ 1.5 \\
ZS verbalized & 5 & 0.0 $\pm$ 0.0 & 15.3 $\pm$ 34.2 & 65.9 $\pm$ 36.9 & 93.6 $\pm$ 8.8 & 100.0 $\pm$ 0.0 & 100.0 $\pm$ 0.0 & 100.0 $\pm$ 0.0 & 67.8 $\pm$ 7.3 & 8.3 $\pm$ 2.0 & 85.9 $\pm$ 1.5 \\
\midrule
\multicolumn{12}{l}{\textbf{Gemini-3-Pro}} \\
Selector, no crops & 5 & 7.0 $\pm$ 6.6 & 22.3 $\pm$ 16.7 & 62.8 $\pm$ 29.6 & 100.0 $\pm$ 0.0 & 100.0 $\pm$ 0.0 & 100.0 $\pm$ 0.0 & 100.0 $\pm$ 0.0 & 70.3 $\pm$ 6.4 & 8.8 $\pm$ 3.1 & 86.5 $\pm$ 1.1 \\
Selector & 5 & 8.1 $\pm$ 3.6 & 77.0 $\pm$ 29.5 & 100.0 $\pm$ 0.0 & 100.0 $\pm$ 0.0 & 100.0 $\pm$ 0.0 & 100.0 $\pm$ 0.0 & 100.0 $\pm$ 0.0 & 83.6 $\pm$ 3.9 & 3.6 $\pm$ 1.2 & 94.3 $\pm$ 1.4 \\
SIEVES & 5 & 14.7 $\pm$ 9.6 & 84.4 $\pm$ 20.6 & 100.0 $\pm$ 0.0 & 100.0 $\pm$ 0.0 & 100.0 $\pm$ 0.0 & 100.0 $\pm$ 0.0 & 100.0 $\pm$ 0.0 & 85.6 $\pm$ 2.9 & 3.1 $\pm$ 1.0 & 94.3 $\pm$ 1.4 \\
ZS logit & 5 & 22.9 $\pm$ 25.2 & 91.3 $\pm$ 7.2 & 100.0 $\pm$ 0.0 & 100.0 $\pm$ 0.0 & 100.0 $\pm$ 0.0 & 100.0 $\pm$ 0.0 & 100.0 $\pm$ 0.0 & 87.7 $\pm$ 3.9 & 2.9 $\pm$ 1.1 & 94.3 $\pm$ 1.4 \\
ZS verbalized & 5 & 0.0 $\pm$ 0.0 & 90.5 $\pm$ 5.6 & 100.0 $\pm$ 0.0 & 100.0 $\pm$ 0.0 & 100.0 $\pm$ 0.0 & 100.0 $\pm$ 0.0 & 100.0 $\pm$ 0.0 & 84.4 $\pm$ 0.8 & 3.4 $\pm$ 0.9 & 94.3 $\pm$ 1.4 \\
\bottomrule
\end{tabular}}
\end{table}

\begin{table}[tb]
\centering
\small
\caption{\textbf{Single-answer results on HR-Bench-8k (800 questions).} Values are mean $\pm$ sample SD across repetitions; $n=1$ rows have no SD. Trained selectors use the fixed ID-selected checkpoint for each metric.}
\label{tab:single_answer_hrbench8k}
\setlength{\tabcolsep}{2pt}
\resizebox{\linewidth}{!}{%
\begin{tabular}{lcrrrrrrrrrr}
\toprule
Method & $n$ & C@1 & C@5 & C@10 & C@15 & C@20 & C@25 & C@30 & $\overline{C@r}$ & AURC & Acc. \\
\midrule
\multicolumn{12}{l}{\textbf{Pixel-Reasoner}} \\
Selector, no crops & 5 & 1.8 $\pm$ 1.0 & 3.0 $\pm$ 3.2 & 7.0 $\pm$ 8.2 & 20.0 $\pm$ 11.9 & 44.7 $\pm$ 6.5 & 64.5 $\pm$ 4.1 & 81.0 $\pm$ 2.9 & 31.7 $\pm$ 3.4 & 20.4 $\pm$ 1.3 & 65.5 $\pm$ 1.0 \\
Selector & 5 & 0.9 $\pm$ 0.8 & 4.4 $\pm$ 3.9 & 21.6 $\pm$ 11.3 & 42.9 $\pm$ 6.9 & 60.8 $\pm$ 5.4 & 78.9 $\pm$ 5.2 & 95.0 $\pm$ 2.7 & 43.5 $\pm$ 2.2 & 17.3 $\pm$ 1.5 & 68.3 $\pm$ 1.0 \\
SIEVES & 5 & 2.6 $\pm$ 0.8 & 7.8 $\pm$ 2.1 & 19.3 $\pm$ 4.7 & 35.8 $\pm$ 6.9 & 54.3 $\pm$ 6.4 & 73.5 $\pm$ 5.0 & 94.8 $\pm$ 4.5 & 41.1 $\pm$ 2.8 & 18.1 $\pm$ 1.1 & 68.3 $\pm$ 1.0 \\
ZS logit & 5 & 0.7 $\pm$ 0.9 & 0.7 $\pm$ 0.9 & 2.2 $\pm$ 2.4 & 14.2 $\pm$ 14.0 & 43.0 $\pm$ 15.6 & 73.9 $\pm$ 3.1 & 94.2 $\pm$ 3.0 & 32.7 $\pm$ 4.7 & 21.2 $\pm$ 1.6 & 68.3 $\pm$ 1.0 \\
ZS verbalized & 5 & 0.0 $\pm$ 0.0 & 0.0 $\pm$ 0.0 & 0.0 $\pm$ 0.0 & 0.0 $\pm$ 0.0 & 0.0 $\pm$ 0.0 & 0.0 $\pm$ 0.0 & 75.6 $\pm$ 42.3 & 10.8 $\pm$ 6.0 & 29.0 $\pm$ 1.0 & 68.3 $\pm$ 1.0 \\
Logprobs & 5 & 0.3 $\pm$ 0.3 & 0.3 $\pm$ 0.3 & 0.3 $\pm$ 0.3 & 1.3 $\pm$ 2.2 & 2.9 $\pm$ 5.8 & 32.5 $\pm$ 20.4 & 84.5 $\pm$ 11.3 & 17.4 $\pm$ 3.6 & 28.1 $\pm$ 1.2 & 68.3 $\pm$ 1.0 \\
\midrule
\multicolumn{12}{l}{\textbf{o3}} \\
Selector, no crops & 5 & 1.3 $\pm$ 0.9 & 1.4 $\pm$ 1.1 & 5.3 $\pm$ 3.8 & 13.7 $\pm$ 4.9 & 32.4 $\pm$ 10.8 & 83.0 $\pm$ 9.1 & 100.0 $\pm$ 0.0 & 33.9 $\pm$ 2.9 & 19.2 $\pm$ 0.6 & 72.1 $\pm$ 1.0 \\
Selector & 5 & 3.0 $\pm$ 3.0 & 5.8 $\pm$ 7.2 & 41.1 $\pm$ 9.8 & 88.6 $\pm$ 9.0 & 100.0 $\pm$ 0.0 & 100.0 $\pm$ 0.0 & 100.0 $\pm$ 0.0 & 62.6 $\pm$ 3.4 & 10.4 $\pm$ 1.2 & 82.1 $\pm$ 1.0 \\
SIEVES & 5 & 8.2 $\pm$ 1.5 & 22.2 $\pm$ 1.2 & 51.5 $\pm$ 8.5 & 91.2 $\pm$ 5.4 & 100.0 $\pm$ 0.0 & 100.0 $\pm$ 0.0 & 100.0 $\pm$ 0.0 & 67.6 $\pm$ 1.6 & 9.3 $\pm$ 0.8 & 82.1 $\pm$ 1.0 \\
ZS logit & 5 & 0.7 $\pm$ 0.7 & 2.9 $\pm$ 5.5 & 17.1 $\pm$ 16.9 & 86.7 $\pm$ 2.3 & 100.0 $\pm$ 0.0 & 100.0 $\pm$ 0.0 & 100.0 $\pm$ 0.0 & 58.2 $\pm$ 2.8 & 12.7 $\pm$ 1.4 & 82.1 $\pm$ 1.0 \\
ZS verbalized & 5 & 0.0 $\pm$ 0.0 & 0.0 $\pm$ 0.0 & 0.0 $\pm$ 0.0 & 87.8 $\pm$ 2.4 & 100.0 $\pm$ 0.0 & 100.0 $\pm$ 0.0 & 100.0 $\pm$ 0.0 & 55.4 $\pm$ 0.3 & 13.2 $\pm$ 1.0 & 82.1 $\pm$ 1.0 \\
\midrule
\multicolumn{12}{l}{\textbf{Gemini-3-Pro}} \\
Selector, no crops & 1 & 5.8 & 31.4 & 85.0 & 100.0 & 100.0 & 100.0 & 100.0 & 74.6 & 5.5 & 85.9 \\
Selector & 1 & 4.5 & 56.5 & 100.0 & 100.0 & 100.0 & 100.0 & 100.0 & 80.1 & 3.5 & 90.1 \\
SIEVES & 1 & 14.4 & 50.9 & 100.0 & 100.0 & 100.0 & 100.0 & 100.0 & 80.8 & 4.4 & 90.1 \\
ZS logit & 1 & 1.4 & 58.0 & 100.0 & 100.0 & 100.0 & 100.0 & 100.0 & 79.9 & 4.9 & 90.1 \\
ZS verbalized & 1 & 0.0 & 0.0 & 100.0 & 100.0 & 100.0 & 100.0 & 100.0 & 71.4 & 6.4 & 90.1 \\
\bottomrule
\end{tabular}}
\end{table}

\begin{table}[tb]
\centering
\small
\caption{\textbf{Single-answer results on MME-RealWorld-Lite (1,782 questions).} Values are mean $\pm$ sample SD across repetitions; $n=1$ rows have no SD. Trained selectors use the fixed ID-selected checkpoint for each metric.}
\label{tab:single_answer_mme_realworld}
\setlength{\tabcolsep}{2pt}
\resizebox{\linewidth}{!}{%
\begin{tabular}{lcrrrrrrrrrr}
\toprule
Method & $n$ & C@1 & C@5 & C@10 & C@15 & C@20 & C@25 & C@30 & $\overline{C@r}$ & AURC & Acc. \\
\midrule
\multicolumn{12}{l}{\textbf{Pixel-Reasoner}} \\
Selector, no crops & 5 & 0.1 $\pm$ 0.1 & 0.1 $\pm$ 0.1 & 4.9 $\pm$ 5.7 & 17.7 $\pm$ 2.3 & 26.6 $\pm$ 1.5 & 34.5 $\pm$ 1.4 & 44.7 $\pm$ 1.6 & 18.4 $\pm$ 1.3 & 30.4 $\pm$ 0.4 & 51.6 $\pm$ 0.3 \\
Selector & 5 & 0.2 $\pm$ 0.1 & 0.3 $\pm$ 0.5 & 4.9 $\pm$ 1.4 & 13.2 $\pm$ 2.7 & 26.0 $\pm$ 1.1 & 33.9 $\pm$ 1.2 & 41.1 $\pm$ 1.9 & 17.1 $\pm$ 0.8 & 31.9 $\pm$ 0.4 & 51.3 $\pm$ 0.8 \\
SIEVES & 5 & 1.0 $\pm$ 1.1 & 5.3 $\pm$ 1.9 & 10.3 $\pm$ 2.1 & 17.9 $\pm$ 0.9 & 26.0 $\pm$ 1.5 & 32.5 $\pm$ 2.0 & 41.6 $\pm$ 2.0 & 19.2 $\pm$ 0.9 & 31.0 $\pm$ 0.5 & 51.3 $\pm$ 0.8 \\
ZS logit & 5 & 0.0 $\pm$ 0.0 & 0.0 $\pm$ 0.0 & 0.0 $\pm$ 0.0 & 0.0 $\pm$ 0.0 & 1.3 $\pm$ 1.9 & 5.4 $\pm$ 4.1 & 28.2 $\pm$ 5.0 & 5.0 $\pm$ 0.8 & 36.3 $\pm$ 0.3 & 51.3 $\pm$ 0.8 \\
ZS verbalized & 5 & 0.0 $\pm$ 0.0 & 0.0 $\pm$ 0.0 & 0.0 $\pm$ 0.0 & 0.0 $\pm$ 0.0 & 0.0 $\pm$ 0.0 & 0.0 $\pm$ 0.0 & 0.0 $\pm$ 0.0 & 0.0 $\pm$ 0.0 & 43.5 $\pm$ 1.0 & 51.3 $\pm$ 0.8 \\
Logprobs & 5 & 0.2 $\pm$ 0.4 & 0.2 $\pm$ 0.4 & 0.4 $\pm$ 0.6 & 1.0 $\pm$ 1.5 & 2.3 $\pm$ 3.0 & 7.2 $\pm$ 4.2 & 14.3 $\pm$ 4.1 & 3.6 $\pm$ 1.7 & 39.5 $\pm$ 1.7 & 51.3 $\pm$ 0.8 \\
\midrule
\multicolumn{12}{l}{\textbf{o3}} \\
Selector, no crops & 5 & 0.7 $\pm$ 0.5 & 1.6 $\pm$ 0.2 & 2.4 $\pm$ 0.5 & 5.2 $\pm$ 1.2 & 9.3 $\pm$ 1.8 & 32.2 $\pm$ 2.1 & 45.4 $\pm$ 2.7 & 13.8 $\pm$ 0.6 & 29.3 $\pm$ 0.8 & 56.9 $\pm$ 1.1 \\
Selector & 5 & 1.6 $\pm$ 0.6 & 0.4 $\pm$ 0.2 & 10.2 $\pm$ 1.4 & 14.8 $\pm$ 1.5 & 18.5 $\pm$ 3.2 & 26.6 $\pm$ 3.5 & 48.9 $\pm$ 6.6 & 17.3 $\pm$ 1.9 & 28.6 $\pm$ 1.3 & 58.9 $\pm$ 0.9 \\
SIEVES & 5 & 1.0 $\pm$ 0.2 & 4.0 $\pm$ 1.7 & 8.9 $\pm$ 2.4 & 15.5 $\pm$ 2.2 & 22.2 $\pm$ 1.8 & 31.0 $\pm$ 2.9 & 45.1 $\pm$ 2.8 & 18.2 $\pm$ 1.5 & 28.3 $\pm$ 0.8 & 58.9 $\pm$ 0.9 \\
ZS logit & 5 & 0.0 $\pm$ 0.0 & 0.0 $\pm$ 0.0 & 0.0 $\pm$ 0.0 & 0.0 $\pm$ 0.0 & 1.3 $\pm$ 1.8 & 12.9 $\pm$ 3.6 & 22.8 $\pm$ 3.1 & 5.3 $\pm$ 1.0 & 35.2 $\pm$ 0.8 & 58.9 $\pm$ 0.9 \\
ZS verbalized & 5 & 0.0 $\pm$ 0.0 & 0.0 $\pm$ 0.0 & 0.0 $\pm$ 0.0 & 0.0 $\pm$ 0.0 & 0.0 $\pm$ 0.0 & 0.0 $\pm$ 0.0 & 0.0 $\pm$ 0.0 & 0.0 $\pm$ 0.0 & 36.1 $\pm$ 0.9 & 58.9 $\pm$ 0.9 \\
\midrule
\multicolumn{12}{l}{\textbf{Gemini-3-Pro}} \\
Selector, no crops & 1 & 2.4 & 5.1 & 24.2 & 37.4 & 47.1 & 57.0 & 70.2 & 34.8 & 21.9 & 61.3 \\
Selector & 1 & 0.7 & 1.2 & 18.2 & 26.8 & 37.3 & 48.5 & 64.4 & 28.2 & 22.9 & 63.2 \\
SIEVES & 1 & 3.4 & 12.2 & 17.2 & 31.6 & 39.6 & 53.5 & 68.9 & 32.3 & 21.5 & 63.2 \\
ZS logit & 1 & 0.0 & 0.0 & 0.0 & 0.0 & 28.6 & 44.9 & 63.0 & 19.5 & 26.4 & 63.2 \\
ZS verbalized & 1 & 0.0 & 0.0 & 0.0 & 0.0 & 0.0 & 0.0 & 65.0 & 9.3 & 28.2 & 63.2 \\
\bottomrule
\end{tabular}}
\end{table}

\begin{table}[tb]
\centering
\small
\caption{\textbf{Single-answer results on VizWiz (4,319 questions).} Values are mean $\pm$ sample SD across repetitions; $n=1$ rows have no SD. Trained selectors use the fixed ID-selected checkpoint for each metric.}
\label{tab:single_answer_vizwiz}
\setlength{\tabcolsep}{2pt}
\resizebox{\linewidth}{!}{%
\begin{tabular}{lcrrrrrrrrrr}
\toprule
Method & $n$ & C@1 & C@5 & C@10 & C@15 & C@20 & C@25 & C@30 & $\overline{C@r}$ & AURC & Acc. \\
\midrule
\multicolumn{12}{l}{\textbf{Pixel-Reasoner}} \\
Selector, no crops & 5 & 0.1 $\pm$ 0.1 & 0.1 $\pm$ 0.1 & 0.4 $\pm$ 0.3 & 1.1 $\pm$ 0.4 & 2.3 $\pm$ 1.6 & 5.0 $\pm$ 5.0 & 16.7 $\pm$ 2.0 & 3.7 $\pm$ 0.9 & 45.6 $\pm$ 0.5 & 38.1 $\pm$ 0.4 \\
Selector & 5 & 0.1 $\pm$ 0.1 & 0.1 $\pm$ 0.1 & 0.3 $\pm$ 0.2 & 1.2 $\pm$ 0.8 & 0.2 $\pm$ 0.3 & 1.2 $\pm$ 1.6 & 9.6 $\pm$ 1.0 & 1.8 $\pm$ 0.3 & 44.2 $\pm$ 0.3 & 38.0 $\pm$ 0.1 \\
SIEVES & 5 & 0.1 $\pm$ 0.1 & 0.2 $\pm$ 0.4 & 0.4 $\pm$ 0.8 & 1.9 $\pm$ 2.1 & 6.6 $\pm$ 3.3 & 13.8 $\pm$ 2.4 & 22.1 $\pm$ 1.8 & 6.5 $\pm$ 1.3 & 42.3 $\pm$ 0.5 & 38.0 $\pm$ 0.1 \\
ZS logit & 5 & 0.0 $\pm$ 0.0 & 0.0 $\pm$ 0.0 & 0.0 $\pm$ 0.0 & 0.0 $\pm$ 0.0 & 0.0 $\pm$ 0.0 & 0.0 $\pm$ 0.0 & 0.0 $\pm$ 0.0 & 0.0 $\pm$ 0.0 & 53.1 $\pm$ 0.8 & 38.0 $\pm$ 0.1 \\
ZS verbalized & 5 & 0.0 $\pm$ 0.0 & 0.0 $\pm$ 0.0 & 0.0 $\pm$ 0.0 & 0.0 $\pm$ 0.0 & 0.0 $\pm$ 0.0 & 0.0 $\pm$ 0.0 & 0.0 $\pm$ 0.0 & 0.0 $\pm$ 0.0 & 50.8 $\pm$ 0.5 & 38.0 $\pm$ 0.1 \\
Logprobs & 5 & 0.1 $\pm$ 0.1 & 0.1 $\pm$ 0.1 & 0.1 $\pm$ 0.1 & 0.1 $\pm$ 0.1 & 0.1 $\pm$ 0.1 & 0.2 $\pm$ 0.2 & 0.7 $\pm$ 0.8 & 0.2 $\pm$ 0.2 & 50.6 $\pm$ 1.1 & 38.0 $\pm$ 0.1 \\
\midrule
\multicolumn{12}{l}{\textbf{o3}} \\
Selector, no crops & 1 & 0.0 & 0.0 & 0.0 & 0.0 & 6.3 & 5.4 & 14.6 & 3.8 & 40.9 & 45.8 \\
Selector & 1 & 0.1 & 0.0 & 0.2 & 5.5 & 0.0 & 1.3 & 19.1 & 3.8 & 35.2 & 46.9 \\
SIEVES & 1 & 0.2 & 0.0 & 0.0 & 7.2 & 11.7 & 27.2 & 40.5 & 12.4 & 33.3 & 46.9 \\
ZS logit & 1 & 0.0 & 0.0 & 0.0 & 0.0 & 0.0 & 0.0 & 0.0 & 0.0 & 39.0 & 46.9 \\
ZS verbalized & 1 & 0.0 & 0.0 & 0.0 & 0.0 & 0.0 & 0.0 & 0.0 & 0.0 & 41.6 & 46.9 \\
\midrule
\multicolumn{12}{l}{\textbf{Gemini-3-Pro}} \\
Selector, no crops & 1 & 0.0 & 0.0 & 0.0 & 0.7 & 3.8 & 6.5 & 18.4 & 4.2 & 41.9 & 44.1 \\
Selector & 1 & 0.0 & 0.1 & 1.0 & 4.2 & 0.0 & 4.5 & 9.4 & 2.7 & 39.5 & 43.6 \\
SIEVES & 1 & 0.0 & 0.0 & 0.0 & 2.0 & 8.1 & 21.3 & 29.2 & 8.7 & 37.2 & 43.6 \\
ZS logit & 1 & 0.0 & 0.0 & 0.0 & 0.0 & 0.0 & 0.0 & 0.0 & 0.0 & 47.7 & 43.6 \\
ZS verbalized & 1 & 0.0 & 0.0 & 0.0 & 0.0 & 0.0 & 0.0 & 0.0 & 0.0 & 46.2 & 43.6 \\
\bottomrule
\end{tabular}}
\end{table}

\begin{table}[tb]
\centering
\small
\caption{\textbf{Single-answer results on AdVQA (10,000 questions).} Values are mean $\pm$ sample SD across repetitions; $n=1$ rows have no SD. Trained selectors use the fixed ID-selected checkpoint for each metric.}
\label{tab:single_answer_advqa}
\setlength{\tabcolsep}{2pt}
\resizebox{\linewidth}{!}{%
\begin{tabular}{lcrrrrrrrrrr}
\toprule
Method & $n$ & C@1 & C@5 & C@10 & C@15 & C@20 & C@25 & C@30 & $\overline{C@r}$ & AURC & Acc. \\
\midrule
\multicolumn{12}{l}{\textbf{Pixel-Reasoner}} \\
Selector, no crops & 1 & 0.0 & 0.0 & 0.0 & 0.0 & 1.0 & 19.0 & 33.6 & 7.7 & 33.7 & 54.5 \\
Selector & 1 & 0.0 & 0.0 & 0.2 & 0.9 & 3.0 & 11.2 & 28.8 & 6.3 & 33.0 & 58.1 \\
SIEVES & 1 & 0.0 & 1.3 & 2.8 & 3.5 & 7.5 & 15.7 & 34.1 & 9.3 & 32.2 & 58.1 \\
ZS logit & 1 & 0.0 & 0.0 & 0.0 & 0.0 & 0.0 & 0.0 & 0.0 & 0.0 & 38.6 & 58.1 \\
ZS verbalized & 1 & 0.0 & 0.0 & 0.0 & 0.0 & 0.0 & 0.0 & 0.0 & 0.0 & 40.7 & 58.1 \\
Logprobs & 1 & 0.0 & 0.0 & 0.0 & 0.0 & 0.1 & 0.1 & 0.1 & 0.1 & 40.3 & 58.1 \\
\midrule
\multicolumn{12}{l}{\textbf{o3}} \\
Selector, no crops & 1 & 0.0 & 0.0 & 0.0 & 2.5 & 7.6 & 40.8 & 85.2 & 19.4 & 25.2 & 68.0 \\
Selector & 1 & 0.0 & 0.1 & 3.8 & 15.8 & 30.1 & 83.1 & 100.0 & 33.3 & 20.1 & 72.5 \\
SIEVES & 1 & 0.1 & 0.1 & 8.2 & 18.5 & 43.6 & 84.9 & 100.0 & 36.5 & 19.6 & 72.5 \\
ZS logit & 1 & 0.0 & 0.0 & 0.0 & 0.0 & 38.0 & 75.9 & 100.0 & 30.6 & 21.9 & 72.5 \\
ZS verbalized & 1 & 0.0 & 0.0 & 0.0 & 0.0 & 0.0 & 92.4 & 100.0 & 27.5 & 24.9 & 72.5 \\
\midrule
\multicolumn{12}{l}{\textbf{Gemini-3-Pro}} \\
Selector, no crops & 1 & 0.1 & 0.1 & 15.0 & 69.6 & 100.0 & 100.0 & 100.0 & 55.0 & 12.8 & 80.3 \\
Selector & 1 & 0.0 & 0.1 & 32.5 & 79.9 & 100.0 & 100.0 & 100.0 & 58.9 & 11.9 & 81.4 \\
SIEVES & 1 & 0.1 & 6.1 & 33.3 & 78.9 & 100.0 & 100.0 & 100.0 & 59.8 & 11.5 & 81.4 \\
ZS logit & 1 & 0.0 & 0.0 & 0.0 & 0.0 & 100.0 & 100.0 & 100.0 & 42.9 & 17.3 & 81.4 \\
ZS verbalized & 1 & 0.0 & 0.0 & 0.0 & 0.0 & 100.0 & 100.0 & 100.0 & 42.9 & 17.0 & 81.4 \\
\bottomrule
\end{tabular}}
\end{table}

%% file: tables/supp/automatic_supervision.tex
\begin{table}[tb]
\centering
\caption{\textbf{Automatic selector supervision.}
The selector has no evidence targets.
The reference-box row uses a mix of document-derived and reference boxes.
The final row uses automatic boxes for every training example.
Coherence labels are automatic in both SIEVES rows.}
\label{tab:automatic_supervision}
\setlength{\tabcolsep}{5pt}
\resizebox{\linewidth}{!}{%
\begin{tabular}{lrrr}
\toprule
Training setup & ID $\overline{C@r}$ & OOD $\overline{C@r}$ & OOD AURC$\downarrow$ \\
\midrule
Selector & 26.7 & 24.1 & 28.1 \\
SIEVES, reference boxes & 26.6 & \textbf{27.0} & \textbf{27.3} \\
SIEVES, automatic boxes & 24.2 & 26.3 & 27.6 \\
\bottomrule
\end{tabular}}
\end{table}

%% file: tables/supp/optional_zoom.tex
\begin{table}[tb]
\centering
\caption{\textbf{Optional versus requested zooming with Pixel-Reasoner.}
The paired replay keeps the prompt and stored samples fixed and changes only whether an answer may be returned before a crop.
Higher $\overline{C@r}$ and lower AURC are better.}
\label{tab:optional_zoom}
\setlength{\tabcolsep}{4pt}
\resizebox{\linewidth}{!}{%
\begin{tabular}{lrrrrrrrr}
\toprule
& \multicolumn{2}{c}{V* Bench} & \multicolumn{2}{c}{HR-Bench-8k} & \multicolumn{2}{c}{VizWiz} & \multicolumn{2}{c}{AdVQA} \\
\cmidrule(lr){2-3}\cmidrule(lr){4-5}\cmidrule(lr){6-7}\cmidrule(lr){8-9}
Zoom policy & $\overline{C@r}$ & AURC & $\overline{C@r}$ & AURC & $\overline{C@r}$ & AURC & $\overline{C@r}$ & AURC \\
\midrule
Optional & 54.7 & 14.07 & 40.7 & 18.53 & 1.6 & 46.80 & 8.6 & 32.68 \\
Requested & \textbf{60.3} & \textbf{11.68} & \textbf{42.1} & \textbf{17.94} & \textbf{3.8} & \textbf{44.55} & \textbf{9.3} & \textbf{32.37} \\
\bottomrule
\end{tabular}}
\end{table}

%% file: tables/supp/broad_crop_scope.tex
\begin{table}[tb]
\centering
\caption{\textbf{Tasks that do not require a tight crop.}
Results are averaged over Pixel-Reasoner, o3, and Gemini-3-Pro on the same questions.
The subset is defined by the smallest returned crop as a share of image area.}
\label{tab:broad_crop_scope}
\setlength{\tabcolsep}{3.5pt}
\resizebox{\linewidth}{!}{%
\begin{tabular}{clrrrrrr}
\toprule
& & \multicolumn{3}{c}{VizWiz} & \multicolumn{3}{c}{AdVQA} \\
\cmidrule(lr){3-5}\cmidrule(lr){6-8}
Min. crop area & Method & C@20 & $\overline{C@r}$ & AURC & C@10 & $\overline{C@r}$ & AURC \\
\midrule
\multirow{2}{*}{$>25\%$} & Selector, no tool & 10.9 & 11.0 & 34.0 & \textbf{13.6} & 34.7 & 21.5 \\
& SIEVES, tool & \textbf{21.9} & \textbf{17.0} & \textbf{30.1} & 11.7 & \textbf{38.1} & \textbf{19.9} \\
\midrule
\multirow{2}{*}{$>50\%$} & Selector, no tool & 19.2 & 16.6 & 30.1 & \textbf{19.3} & 35.3 & \textbf{20.2} \\
& SIEVES, tool & \textbf{28.5} & \textbf{22.1} & \textbf{26.8} & 15.4 & \textbf{39.8} & 20.3 \\
\bottomrule
\end{tabular}}
\end{table}

%% file: tables/supp/additional_label_ablation.tex
\begin{table}[tb]
\centering
\caption{\textbf{Localization label ablation with Pixel-Reasoner.}
Binary localization matches the decision of whether the needed region is visible.}
\label{tab:additional_label_ablation}
\setlength{\tabcolsep}{4pt}
\resizebox{\linewidth}{!}{%
\begin{tabular}{lrrrr}
\toprule
Label design & ID $\overline{C@r}$ & ID AURC & OOD $\overline{C@r}$ & OOD AURC \\
\midrule
Continuous mIoGT + MSE & 26.5 & 25.8 & 26.2 & 27.9 \\
Binary mIoGT $\geq0.75$ + BCE & \textbf{26.6} & \textbf{25.7} & \textbf{27.0} & \textbf{27.3} \\
\bottomrule
\end{tabular}}
\end{table}

%% file: tables/draft/sr_threshold_ablation.tex
\begin{table}[tb]
\centering
\caption{\textbf{Spatial recall threshold ablation.} We vary the mIoGT threshold $t$ in $g_{loc}=\indicator[\text{mIoGT}\geq t]$, used to define the ground truth label indicating whether localization is correct or not for a reasoning trace. Reported are average coverage ($\overline{\mathrm{C@r}}$; $\uparrow$) and AURC ($\downarrow$), on the ID tuning set (Thyme) and averaged across five OOD benchmarks, for Pixel-Reasoner with localization. The row highlighted in gray denotes \protect\colorbox{gray!15}{the final SIEVES setting ($t=0.75$)}, as in the main text.}
\label{tab:sr_threshold_ablation}
\setlength{\tabcolsep}{4pt}
\begin{tabular}{l cc cc}
\toprule
 & \multicolumn{2}{c}{ID} & \multicolumn{2}{c}{OOD} \\
\cmidrule(lr){2-3}\cmidrule(lr){4-5}
$t$ & $\overline{\mathrm{C@r}}\uparrow$ & AURC$\downarrow$ & $\overline{\mathrm{C@r}}\uparrow$ & AURC$\downarrow$ \\
\midrule
0.25 & \underline{26.5} & \underline{25.8} & 25.6 & 27.8 \\
0.50 & 24.0 & 27.1 & \underline{26.9} & \underline{27.7} \\
\cellcolor{gray!15} 0.75 & \cellcolor{gray!15} \textbf{26.6} & \cellcolor{gray!15} \textbf{25.7} & \cellcolor{gray!15} \textbf{27.0} & \cellcolor{gray!15} \textbf{27.3} \\
\bottomrule
\end{tabular}
\end{table}

%% file: tables/draft/vstar_confidence_by_loc_answer.tex
\begin{table}[tb]
\centering
\caption{\textbf{SIEVES preferentially accepts correct answers with correct localization on V* Bench.} Columns report mean confidence and the percentage accepted within each group at 5\% risk.}
\label{tab:vstar_confidence_by_loc_answer}
\small
\setlength{\tabcolsep}{3pt}
\begin{tabular}{@{}llrrrr@{}}
\toprule
Reasoner & Cell & \shortstack{Selector\\conf.} & \shortstack{SIEVES\\conf.} & \shortstack{Selector\\selected} & \shortstack{SIEVES\\selected} \\
\midrule
Pixel-Reasoner & Loc. corr. / Ans. corr. & 0.60 & 0.54 & 3.1\% & 16.8\% \\
 & Loc. corr. / Ans. incorr. & 0.47 & 0.41 & 0.0\% & 1.5\% \\
 & Loc. incorr. / Ans. corr. & 0.54 & 0.45 & 1.0\% & 8.4\% \\
 & Loc. incorr. / Ans. incorr. & 0.34 & 0.32 & 0.0\% & 2.5\% \\
\midrule
o3 & Loc. corr. / Ans. corr. & 0.51 & 0.48 & 0.0\% & 26.7\% \\
 & Loc. corr. / Ans. incorr. & 0.53 & 0.38 & 0.0\% & 14.8\% \\
 & Loc. incorr. / Ans. corr. & 0.58 & 0.41 & 0.0\% & 11.7\% \\
 & Loc. incorr. / Ans. incorr. & 0.41 & 0.30 & 0.0\% & 3.7\% \\
\bottomrule
\end{tabular}
\end{table}

%% file: tables/draft/vstar_confidence_by_loc_answer_filtered.tex
\begin{table}[tb]
\centering
\caption{\textbf{SIEVES preferentially accepts correct answers with correct localization on V* Bench.} We include examples with a crop and a parseable answer. Columns report mean confidence and the percentage accepted within each group at 5\% risk.}
\label{tab:vstar_confidence_by_loc_answer_filtered}
\small
\setlength{\tabcolsep}{3pt}
\begin{tabular}{@{}llrrrr@{}}
\toprule
Reasoner & Cell & \shortstack{Selector\\conf.} & \shortstack{SIEVES\\conf.} & \shortstack{Selector\\selected} & \shortstack{SIEVES\\selected} \\
\midrule
Pixel-Reasoner & Loc. corr. / Ans. corr. & 0.60 & 0.54 & 3.1\% & 16.8\% \\
 & Loc. corr. / Ans. incorr. & 0.47 & 0.41 & 0.0\% & 1.5\% \\
 & Loc. incorr. / Ans. corr. & 0.54 & 0.46 & 1.2\% & 9.4\% \\
 & Loc. incorr. / Ans. incorr. & 0.36 & 0.34 & 0.0\% & 2.9\% \\
\midrule
o3 & Loc. corr. / Ans. corr. & 0.51 & 0.48 & 0.0\% & 26.7\% \\
 & Loc. corr. / Ans. incorr. & 0.57 & 0.40 & 0.0\% & 16.0\% \\
 & Loc. incorr. / Ans. corr. & 0.58 & 0.41 & 0.0\% & 12.3\% \\
 & Loc. incorr. / Ans. incorr. & 0.53 & 0.39 & 0.0\% & 5.1\% \\
\bottomrule
\end{tabular}
\end{table}

%% file: tables/supp/ablation_o3_g3p.tex
\begin{table*}[tb]
\centering
\newcommand{\hlcellSupp}[1]{\cellcolor{gray!15}\raisebox{-0.1ex}{#1}}
\newcommand{\hlnumSupp}[1]{\cellcolor{gray!15}{\vphantom{Ay}#1}}
\captionsetup[subtable]{skip=1pt}
\begin{subtable}[t]{0.52\textwidth}
\centering
\caption{\textbf{Objective weight ablation.}}
\vspace{-2pt}
\setlength{\tabcolsep}{3pt}
\resizebox{0.78\textwidth}{!}{%
\renewcommand{\arraystretch}{0.85}
\begin{tabular}{c rrr rr}
\toprule
 & & & & \multicolumn{2}{c}{OOD} \\
\cmidrule(lr){5-6}
$f$ & $\lambda_{c}$ & $\lambda_{l}$ & $\lambda_{h}$ & $\overline{C@r}\uparrow$ & AURC$\downarrow$ \\
\midrule
\multirow{9}{*}{o3} & 1.0 & 0.0 & 0.0 & 35.3 & 20.8 \\
& 0.0 & 1.0 & 0.0 & 37.5 & 20.7 \\
& 0.0 & 0.0 & 1.0 & 39.2 & \underline{19.3} \\
& 0.6 & 0.4 & 0.0 & 38.4 & 20.4 \\
& 0.6 & 0.0 & 0.4 & 39.8 & 20.4 \\
\cmidrule(l){2-6}
& \hlnumSupp{0.6} & \hlnumSupp{0.3} & \hlnumSupp{0.1} & \hlnumSupp{41.0} & \hlnumSupp{19.8} \\
& 0.6 & 0.2 & 0.2 & 37.7 & 21.3 \\
& 0.3 & 0.3 & 0.3 & \textbf{42.6} & \textbf{19.0} \\
& 0.2 & 0.4 & 0.4 & \underline{41.3} & 19.6 \\
\midrule \midrule
\multirow{9}{*}{G3P} & 1.0 & 0.0 & 0.0 & 49.5 & 16.2 \\
& 0.0 & 1.0 & 0.0 & 36.4 & 22.9 \\
& 0.0 & 0.0 & 1.0 & 40.6 & 19.8 \\
& 0.6 & 0.4 & 0.0 & 52.1 & 16.5 \\
& 0.6 & 0.0 & 0.4 & 53.6 & 16.6 \\
\cmidrule(l){2-6}
& \hlnumSupp{0.6} & \hlnumSupp{0.3} & \hlnumSupp{0.1} & \hlnumSupp{53.4} & \hlnumSupp{\underline{15.5}} \\
& 0.6 & 0.2 & 0.2 & \textbf{54.7} & 15.7 \\
& 0.3 & 0.3 & 0.3 & \underline{54.1} & \textbf{15.2} \\
& 0.2 & 0.4 & 0.4 & 51.6 & 16.3 \\
\bottomrule
\end{tabular}%
}%
\end{subtable}
\hfill
\begin{minipage}[t]{0.45\textwidth}
\centering
\begin{subtable}[t]{\textwidth}
\centering
\caption{\textbf{Crop-answer coherence annotator.}}
\setlength{\tabcolsep}{2pt}
\resizebox{\textwidth}{!}{%
\begin{tabular}{c l rr}
\toprule
$f$ & Model & $\overline{C@r}\uparrow$ & AURC$\downarrow$ \\
\midrule
\multirow{3}{*}{o3} & \hlcellSupp{Qwen 2.5 VL 7B} & \hlcellSupp{\underline{41.0}} & \hlcellSupp{\textbf{19.8}} \\
& Gemma 3 4B & \textbf{41.2} & \underline{20.3} \\
& Qwen 2.5 VL 72B & 38.5 & 21.0 \\
\midrule \midrule
\multirow{3}{*}{G3P} & \hlcellSupp{Qwen 2.5 VL 7B} & \hlcellSupp{53.4} & \hlcellSupp{\textbf{15.5}} \\
& Gemma 3 4B & \underline{53.6} & 16.2 \\
& Qwen 2.5 VL 72B & \textbf{54.7} & \underline{15.8} \\
\bottomrule
\end{tabular}%
}%
\end{subtable}
\vspace{1.4em}
\begin{subtable}[t]{\textwidth}
\centering
\vspace{10pt}
\caption{\textbf{Loc. bounding-box annotator.}}
\setlength{\tabcolsep}{2pt}
\resizebox{\textwidth}{!}{%
\begin{tabular}{c l rr}
\toprule
$f$ & BBox annot. & $\overline{C@r}\uparrow$ & AURC$\downarrow$ \\
\midrule
\multirow{2}{*}{o3} & \hlcellSupp{Human} & \hlcellSupp{41.0} & \hlcellSupp{19.8} \\
& Gemini-3-Flash & 40.0 & 21.0 \\
\midrule \midrule
\multirow{2}{*}{G3P} & \hlcellSupp{Human} & \hlcellSupp{53.4} & \hlcellSupp{15.5} \\
& Gemini-3-Flash & 52.1 & 16.2 \\
\bottomrule
\end{tabular}%
}%
\end{subtable}
\end{minipage}
\vspace{0.6em}
\caption{\textbf{Additional ablations on stronger OOD reasoners.}
We report the same selector-design ablations as in \Cref{tab:ablation_id_ood}, but for o3 and Gemini-3-Pro only.
G3P denotes Gemini-3-Pro. Since these are evaluated purely out-of-distribution, we report OOD metrics only.
\textbf{Bold} indicates best result per reasoner and subtable, \underline{underlined} second best, and the gray row marks the default SIEVES configuration.}
\label{tab:ablation_o3_g3p}
\vspace{-6pt}
\end{table*}

%% file: figs/qual_example_coherence_mme.tex
\begin{figure*}[tb]
    \centering
    \begin{minipage}[t]{0.48\textwidth}
        \centering
        \includegraphics[width=\textwidth]{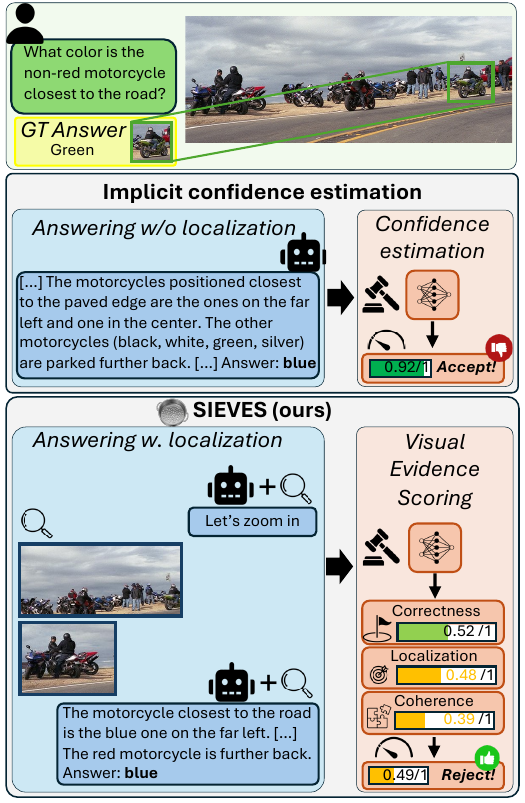}
    \end{minipage}
    \hfill
    \begin{minipage}[t]{0.48\textwidth}
        \centering
        \includegraphics[width=\textwidth]{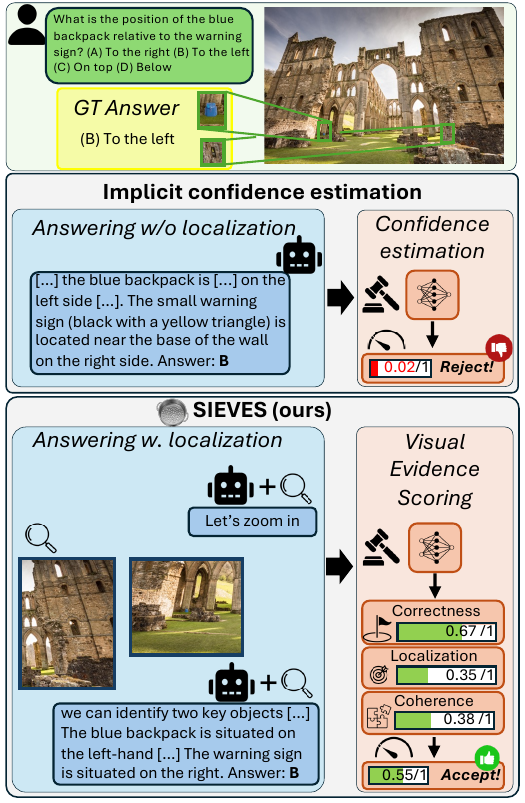}
    \end{minipage}
    \caption{\textbf{More selector errors corrected by visual evidence.} \textbf{Left:} SIEVES abstains because the answer refers to the wrong motorcycle. \textbf{Right:} SIEVES accepts the correct relative-position answer because the crop supports it.}
    \label{fig:qual_example_coherence_mme}
\end{figure*}

%% file: tables/related_work_comparison.tex
\begin{table}[tb]
\centering
\caption{\textbf{Comparison with selective-prediction methods.}
``Internal states'' means that a method needs hidden states, logits, or model-specific features.
Calls count full reasoner calls per question; ReCoVERR may ask up to 100 auxiliary questions.}
\label{tab:related_work_comparison}
\setlength{\tabcolsep}{3pt}
\resizebox{\linewidth}{!}{%
\begin{tabular}{lcccc}
\toprule
Method & Internal states & Reasoner calls & Learned selector & Scores visual evidence \\
\midrule
Reliable VQA \citep{whitehead2022reliable} & yes & 1 & yes & no \\
Learning from Peers \citep{dancette2024selective} & yes & 1 & yes & no \\
HARMONY \citep{mushtaq2025harmony} & yes & 1 & yes & no \\
Consistency \citep{khan2024uncertainty} & no & $>1$ & no & no \\
ReCoVERR \citep{srinivasan2024recovrr} & no & $1+$ auxiliary & no & indirect \\
SIEVES & no & 1 & yes & yes \\
\bottomrule
\end{tabular}}
\end{table}

%% file: tables/draft/recoverr_comparison.tex
\begin{table}[tb]
\centering
\caption{\textbf{Black-box comparison on V* Bench using the same Pixel-Reasoner answers.}
We evaluate one answer per question from a single repetition on all 191 V* Bench questions.
ReCoVERR may ask up to 100 auxiliary questions.
}
\label{tab:recoverr_comparison}
\setlength{\tabcolsep}{4pt}
\resizebox{\linewidth}{!}{%
\begin{tabular}{llrrrrr}
\toprule
Method & Models used for confidence & Cost $\times100$ & Latency & C@5 & $\overline{C@r}$ & AURC$\downarrow$ \\
\midrule
ReCoVERR & InstructBLIP + GPT-3.5/Flan-T5 & \$0.379 & 51 s & 1.6 & 44.1 & 24.04 \\
ReCoVERR & Pixel-Reasoner + Gemma-3-4B & \$0.795 & $\sim$2.5 min & 13.6 & 54.2 & 14.05 \\
SIEVES & Pixel-Reasoner + Gemma-3-4B & \textbf{\$0.0070} & \textbf{0.5 s} & \textbf{21.5} & \textbf{62.9} & \textbf{11.71} \\
\bottomrule
\end{tabular}}
\end{table}

%% file: tables/supp/recoverr_extended.tex
\begin{table}[tb]
\centering
\caption{\textbf{ReCoVERR search across five benchmarks.} We use a single repetition of Pixel-Reasoner answers. Gemma supplies direct confidence and evidence checks; Pixel-Reasoner answers auxiliary questions. Search allows up to ten rounds with ten questions per round. Scores come from a search run at 5\% risk on V* and 1\% elsewhere; we evaluate their rankings at the reported target risk. VizWiz uses 4,317 traces.}
\label{tab:recoverr_extended}
\setlength{\tabcolsep}{4pt}
\resizebox{\linewidth}{!}{%
\begin{tabular}{lrrrrrrr}
\toprule
 & & \multicolumn{3}{c}{Direct Gemma score} & \multicolumn{3}{c}{ReCoVERR search} \\
\cmidrule(lr){3-5}\cmidrule(lr){6-8}
Benchmark & Questions & C@$r$ & $\overline{C@r}$ & AURC & C@$r$ & $\overline{C@r}$ & AURC \\
\midrule
V* Bench & 191 & 0.0 & 42.9 & 15.7 & 0.0 & 54.5 & 15.4 \\
HR-Bench-8k & 800 & 2.9 & 34.0 & 20.9 & 0.6 & 28.6 & 24.5 \\
MME-RealWorld-Lite & 1782 & 3.5 & 11.8 & 35.3 & 0.2 & 0.3 & 41.6 \\
VizWiz & 4317 & 0.7 & 1.2 & 47.9 & 0.3 & 0.3 & 54.8 \\
AdVQA & 10000 & 0.0 & 3.4 & 36.1 & 0.0 & 0.4 & 36.4 \\
\bottomrule
\end{tabular}}
\end{table}

%% file: tables/supp/baselines/consistency_baseline.tex
\begin{table}[t]
  \centering
  \setlength{\tabcolsep}{2.5pt}
  \resizebox{\textwidth}{!}{%
  \begin{tabular}{l ccc ccc ccc ccc}
    \toprule
    & \multicolumn{3}{c}{V* Bench}
    & \multicolumn{3}{c}{HR-Bench-8k}
    & \multicolumn{3}{c}{MME-RW-L}
    & \multicolumn{3}{c}{VizWiz} \\
    \cmidrule(lr){2-4}\cmidrule(lr){5-7}\cmidrule(lr){8-10}\cmidrule(lr){11-13}
    Reasoner and method
    & C@5 & $\overline{C@r}$ & AURC$\downarrow$
    & C@1 & $\overline{C@r}$ & AURC$\downarrow$
    & C@5 & $\overline{C@r}$ & AURC$\downarrow$
    & C@20 & $\overline{C@r}$ & AURC$\downarrow$ \\
    \midrule
    Pixel-Reasoner, consistency & 0.0 & 54.1 & 12.2 & 0.0 & 33.7 & 18.9 & 0.0 & 0.0 & 36.8 & 0.0 & 1.6 & 45.7 \\
    Pixel-Reasoner, SIEVES      & 9.7 & 59.5 & 12.8 & 2.3 & 41.1 & 18.1 & 4.0 & 18.7 & 31.0 & 7.6 & 6.6 & 42.3 \\
    \midrule
    o3, consistency             & 81.3 & 92.2 & 2.2 & 0.0 & 78.1 & 6.0 & 0.0 & 8.1 & 29.1 & -- & -- & -- \\
    o3, SIEVES                  & 23.1 & 70.8 & 8.4 & 8.0 & 67.7 & 9.3 & 1.7 & 17.5 & 28.4 & 11.7 & 12.4 & 33.3 \\
    \midrule
    Gemini-3-Pro, consistency   & 96.9 & 85.3 & 1.4 & -- & -- & -- & -- & -- & -- & -- & -- & -- \\
    Gemini-3-Pro, SIEVES        & 95.0 & 85.6 & 3.1 & 14.4 & 80.8 & 4.4 & 12.2 & 32.3 & 21.5 & 8.1 & 8.7 & 37.2 \\
    \bottomrule
  \end{tabular}%
  }
  \caption{\textbf{Five-sample consistency-based abstention.}
  The confidence of each stochastic answer is the fraction of the five samples
  producing the same parsed answer. Consistency uses five full
  reasoner calls; SIEVES uses one reasoner call and one smaller 4B selector
  call. Due to inference cost, o3 has one VizWiz repetition and Gemini-3-Pro
  has one repetition outside V* Bench; these consistency entries are marked
  with dashes. AdVQA has one repetition per reasoner and is omitted.}
  \label{tab:consistency_baseline}
\end{table}

%% file: tables/supp/baselines/selective_test_time_scaling.tex
\begin{table}[t]
  \centering
  \setlength{\tabcolsep}{3pt}
  \resizebox{\textwidth}{!}{%
  \begin{tabular}{ll ccc ccc}
    \toprule
    & & \multicolumn{3}{c}{Agreement / self-consistency}
    & \multicolumn{3}{c}{SIEVES-weighted self-consistency} \\
    \cmidrule(lr){3-5}\cmidrule(lr){6-8}
    Reasoner & Benchmark ($r$)
    & C@$r$ & $\overline{C@r}$ & AURC$\downarrow$
    & C@$r$ & $\overline{C@r}$ & AURC$\downarrow$ \\
    \midrule
    Pixel-Reasoner & V* Bench (5\%)    & 0.0 & 54.1 & 12.1 & \textbf{28.4} & \textbf{70.2} & \textbf{8.6} \\
    Pixel-Reasoner & HR-Bench-8k (1\%) & 0.0 & 34.2 & 18.9 & \textbf{7.9} & \textbf{59.5} & \textbf{12.1} \\
    Pixel-Reasoner & MME-RW-L (5\%)    & 0.0 & 0.0 & 36.3 & \textbf{11.5} & \textbf{27.2} & \textbf{26.3} \\
    Pixel-Reasoner & VizWiz (20\%)      & 0.0 & 1.7 & 43.2 & \textbf{12.2} & \textbf{9.2} & \textbf{38.7} \\
    \midrule
    o3 & V* Bench (5\%)    & \textbf{91.1} & \textbf{96.0} & \textbf{1.4} & 85.8 & 89.3 & 2.1 \\
    o3 & HR-Bench-8k (1\%) & 0.0 & 80.4 & 5.2 & \textbf{30.5} & \textbf{84.2} & \textbf{3.6} \\
    o3 & MME-RW-L (5\%)    & 0.0 & 8.6 & 28.5 & \textbf{7.6} & \textbf{30.2} & \textbf{22.7} \\
    \bottomrule
  \end{tabular}%
  }
  \caption{\textbf{SIEVES complements five-sample test-time scaling.}
  Both methods use the same five stochastic reasoner outputs.
  Agreement selects the majority answer and uses its frequency as confidence
  (five reasoner calls, no selector calls).
  SIEVES-weighted self-consistency sums the SIEVES scores of samples producing
  each answer, selects the largest sum, and uses that sum divided by five as
  confidence (five reasoner and five 4B selector calls).
  SIEVES weighting improves all three selective metrics in 6/7 settings; plain
  agreement is stronger for o3 on V* Bench.}
  \label{tab:selective_test_time_scaling}
\end{table}

%% file: tables/supp/baselines/vtsar_baselines.tex
\begin{table*}[t]
    \caption{\textbf{V* Bench baselines.} \zoomInTool ~ indicates whether the reasoner $f$ explicitly localizes the answer or not. $\overline{C@r}$: average coverage across risk levels 1--30\%.}
    \label{tab:paper_vstar_full_baselines}
    \centering
    \resizebox{\textwidth}{!}{%
    \begin{tabular}{lllrrrrrrrrrr}
    \toprule
    \multirow{2}{*}{$f$} & \multirow{2}{*}{\zoomInTool} & \multirow{2}{*}{Abstention Method} & \multirow{2}{*}{Acc$\uparrow$} & \multicolumn{7}{c}{Coverage at Risk (C@r) $\uparrow$} & \multirow{2}{*}{$\overline{C@r}\uparrow$} & \multirow{2}{*}{AURC$\downarrow$} \\
    &  &  &  & @1 & @5 & @10 & @15 & @20 & @25 & @30 &  &  \\
    \midrule
    \multirow{9}{*}{Pixel-Reasoner} & \multirow{4}{*}{\xmark} & Logprobs & \multirow{4}{*}{79.6} & 0.2 & 0.2 & 9.7 & 61.5 & 97.5 & 100.0 & 100.0 & 52.7 & 14.7 \\
     &  & Zero-shot verb. &  & 0.0 & 0.0 & 0.0 & 0.0 & 93.8 & 100.0 & 100.0 & 42.0 & 17.3 \\
     &  & Zero-shot logit &  & 0.0 & 0.0 & 0.0 & 0.0 & 98.0 & 100.0 & 100.0 & 42.6 & 16.9 \\
     &  & Selector &  & 1.9 & 2.3 & 8.4 & 73.0 & 96.3 & 100.0 & 100.0 & 54.6 & 15.7 \\
    \cmidrule(l){2-13}
     & \multirow{5}{*}{\cmark} & Logprobs & \multirow{5}{*}{80.5} & 0.1 & 0.1 & 9.8 & 41.3 & 100.0 & 100.0 & 100.0 & 50.2 & 15.7 \\
     &  & Zero-shot verb. &  & 0.0 & 0.0 & 0.0 & 0.0 & 100.0 & 100.0 & 100.0 & 42.9 & 17.2 \\
     &  & Zero-shot logit &  & 0.0 & 0.0 & 5.5 & 24.2 & 100.0 & 100.0 & 100.0 & 47.1 & 15.6 \\
     &  & Selector &  & 1.5 & 1.4 & 8.4 & 60.4 & 100.0 & 100.0 & 100.0 & 53.1 & 14.1 \\
     &  & SIEVES (ours) &  & 1.9 & 9.7 & 31.7 & 73.4 & 100.0 & 100.0 & 100.0 & 59.5 & 12.8 \\
    \midrule
    \midrule
    \multirow{7}{*}{o3} & \multirow{3}{*}{\xmark} & Zero-shot verb. & \multirow{3}{*}{71.8} & 0.0 & 0.0 & 0.0 & 0.0 & 0.0 & 0.0 & 100.0 & 14.3 & 26.2 \\
     &  & Zero-shot logit &  & 0.0 & 0.0 & 0.0 & 0.0 & 23.6 & 77.3 & 100.0 & 28.7 & 22.7 \\
     &  & Selector &  & 0.1 & 0.1 & 0.1 & 0.1 & 0.6 & 51.5 & 100.0 & 21.8 & 23.4 \\
    \cmidrule(l){2-13}
     & \multirow{4}{*}{\cmark} & Zero-shot verb. & \multirow{4}{*}{85.9} & 0.0 & 0.0 & 84.4 & 84.4 & 84.4 & 100.0 & 100.0 & 64.7 & 9.1 \\
     &  & Zero-shot logit &  & 1.7 & 1.7 & 84.2 & 84.2 & 84.2 & 100.0 & 100.0 & 65.1 & 9.9 \\
     &  & Selector &  & 0.0 & 0.0 & 26.7 & 100.0 & 100.0 & 100.0 & 100.0 & 61.0 & 9.7 \\
     &  & SIEVES (ours) &  & 3.4 & 23.1 & 69.3 & 100.0 & 100.0 & 100.0 & 100.0 & 70.8 & 8.4 \\
    \midrule
    \midrule
    \multirow{7}{*}{Gemini-3-Pro} & \multirow{3}{*}{\xmark} & Zero-shot verb. & \multirow{3}{*}{86.4} & 0.0 & 0.0 & 0.0 & 100.0 & 100.0 & 100.0 & 100.0 & 57.1 & 12.0 \\
     &  & Zero-shot logit &  & 0.0 & 0.0 & 0.0 & 100.0 & 100.0 & 100.0 & 100.0 & 57.1 & 13.8 \\
     &  & Selector &  & 0.6 & 8.8 & 69.1 & 100.0 & 100.0 & 100.0 & 100.0 & 68.4 & 8.8 \\
    \cmidrule(l){2-13}
     & \multirow{4}{*}{\cmark} & Zero-shot verb. & \multirow{4}{*}{94.3} & 0.0 & 88.3 & 100.0 & 100.0 & 100.0 & 100.0 & 100.0 & 84.0 & 3.1 \\
     &  & Zero-shot logit &  & 0.0 & 93.5 & 100.0 & 100.0 & 100.0 & 100.0 & 100.0 & 84.8 & 2.9 \\
     &  & Selector &  & 2.3 & 40.7 & 100.0 & 100.0 & 100.0 & 100.0 & 100.0 & 77.6 & 3.5 \\
     &  & SIEVES (ours) &  & 4.3 & 95.0 & 100.0 & 100.0 & 100.0 & 100.0 & 100.0 & 85.6 & 3.1 \\
    \bottomrule
    \end{tabular}%
    }%
\end{table*}

%% file: tables/supp/baselines/hrbench_baselines.tex
\begin{table*}[t]
    \caption{\textbf{HR-Bench-8k baselines.} \zoomInTool ~ indicates whether the reasoner $f$ explicitly localizes the answer or not. $\overline{C@r}$: average coverage across risk levels 1--30\%.}
    \label{tab:paper_hrbench_full_baselines}
    \centering
    \resizebox{\textwidth}{!}{%
    \begin{tabular}{lllrrrrrrrrrr}
    \toprule
    \multirow{2}{*}{$f$} & \multirow{2}{*}{\zoomInTool} & \multirow{2}{*}{Abstention Method} & \multirow{2}{*}{Acc$\uparrow$} & \multicolumn{7}{c}{Coverage at Risk (C@r) $\uparrow$} & \multirow{2}{*}{$\overline{C@r}\uparrow$} & \multirow{2}{*}{AURC$\downarrow$} \\
    &  &  &  & @1 & @5 & @10 & @15 & @20 & @25 & @30 &  &  \\
    \midrule
    \multirow{9}{*}{Pixel-Reasoner} & \multirow{4}{*}{\xmark} & Logprobs & \multirow{4}{*}{65.6} & 0.0 & 0.0 & 0.3 & 0.4 & 3.9 & 9.1 & 29.7 & 6.2 & 30.3 \\
     &  & Zero-shot verb. &  & 0.0 & 0.0 & 0.0 & 0.0 & 0.0 & 0.0 & 0.0 & 0.0 & 30.6 \\
     &  & Zero-shot logit &  & 0.0 & 0.0 & 0.0 & 0.0 & 0.0 & 47.1 & 75.8 & 17.6 & 25.9 \\
     &  & Selector &  & 0.9 & 1.8 & 2.9 & 20.2 & 44.2 & 64.9 & 80.7 & 30.8 & 20.3 \\
    \cmidrule(l){2-13}
     & \multirow{5}{*}{\cmark} & Logprobs & \multirow{5}{*}{68.4} & 0.3 & 0.5 & 0.7 & 7.0 & 38.0 & 63.3 & 92.8 & 28.9 & 22.6 \\
     &  & Zero-shot verb. &  & 0.0 & 0.0 & 0.0 & 0.0 & 0.0 & 0.0 & 93.4 & 13.3 & 28.9 \\
     &  & Zero-shot logit &  & 0.0 & 0.0 & 2.4 & 12.1 & 47.3 & 73.9 & 93.9 & 32.8 & 21.3 \\
     &  & Selector &  & 1.0 & 3.5 & 19.4 & 44.5 & 58.3 & 78.4 & 94.5 & 42.7 & 17.4 \\
     &  & SIEVES (ours) &  & 2.3 & 7.1 & 19.0 & 35.0 & 55.5 & 73.0 & 95.6 & 41.1 & 18.1 \\
    \midrule
    \midrule
    \multirow{7}{*}{o3} & \multirow{3}{*}{\xmark} & Zero-shot verb. & \multirow{3}{*}{72.1} & 0.0 & 0.0 & 0.0 & 0.0 & 0.0 & 84.3 & 100.0 & 26.3 & 24.6 \\
     &  & Zero-shot logit &  & 0.0 & 0.0 & 0.0 & 0.0 & 12.6 & 82.0 & 100.0 & 27.8 & 22.4 \\
     &  & Selector &  & 0.3 & 0.8 & 3.3 & 12.7 & 30.3 & 86.0 & 100.0 & 33.3 & 19.2 \\
    \cmidrule(l){2-13}
     & \multirow{4}{*}{\cmark} & Zero-shot verb. & \multirow{4}{*}{82.0} & 0.0 & 0.0 & 0.0 & 89.7 & 89.7 & 100.0 & 100.0 & 54.2 & 13.5 \\
     &  & Zero-shot logit &  & 0.0 & 0.0 & 16.5 & 88.7 & 88.7 & 100.0 & 100.0 & 56.3 & 13.1 \\
     &  & Selector &  & 1.0 & 0.8 & 37.9 & 92.2 & 100.0 & 100.0 & 100.0 & 61.7 & 10.5 \\
     &  & SIEVES (ours) &  & 8.4 & 22.1 & 50.5 & 93.0 & 100.0 & 100.0 & 100.0 & 67.7 & 9.3 \\
    \midrule
    \midrule
    \multirow{7}{*}{Gemini-3-Pro} & \multirow{3}{*}{\xmark} & Zero-shot verb. & \multirow{3}{*}{85.6} & 0.0 & 0.0 & 0.0 & 100.0 & 100.0 & 100.0 & 100.0 & 57.1 & 10.9 \\
     &  & Zero-shot logit &  & 0.0 & 0.0 & 80.0 & 100.0 & 100.0 & 100.0 & 100.0 & 68.6 & 10.1 \\
     &  & Selector &  & 5.8 & 31.4 & 84.6 & 100.0 & 100.0 & 100.0 & 100.0 & 74.5 & 5.4 \\
    \cmidrule(l){2-13}
     & \multirow{4}{*}{\cmark} & Zero-shot verb. & \multirow{4}{*}{90.1} & 0.0 & 0.0 & 100.0 & 100.0 & 100.0 & 100.0 & 100.0 & 71.4 & 6.4 \\
     &  & Zero-shot logit &  & 1.4 & 58.0 & 100.0 & 100.0 & 100.0 & 100.0 & 100.0 & 79.9 & 4.9 \\
     &  & Selector &  & 4.5 & 56.5 & 100.0 & 100.0 & 100.0 & 100.0 & 100.0 & 80.1 & 3.5 \\
     &  & SIEVES (ours) &  & 14.4 & 50.9 & 100.0 & 100.0 & 100.0 & 100.0 & 100.0 & 80.8 & 4.4 \\
    \bottomrule
    \end{tabular}%
    }%
\end{table*}

%% file: tables/supp/baselines/mme_realworld_lite_baselines.tex
\begin{table*}[t]
    \caption{\textbf{MME-RealWorld-Lite baselines.} \zoomInTool ~ indicates whether the reasoner $f$ explicitly localizes the answer or not. $\overline{C@r}$: average coverage across risk levels 1--30\%.}
    \label{tab:paper_mme_realworld_lite_baselines}
    \centering
    \resizebox{\textwidth}{!}{%
    \begin{tabular}{lllrrrrrrrrrr}
    \toprule
    \multirow{2}{*}{$f$} & \multirow{2}{*}{\zoomInTool} & \multirow{2}{*}{Abstention Method} & \multirow{2}{*}{Acc$\uparrow$} & \multicolumn{7}{c}{Coverage at Risk (C@r) $\uparrow$} & \multirow{2}{*}{$\overline{C@r}\uparrow$} & \multirow{2}{*}{AURC$\downarrow$} \\
    &  &  &  & @1 & @5 & @10 & @15 & @20 & @25 & @30 &  &  \\
    \midrule
    \multirow{9}{*}{Pixel-Reasoner} & \multirow{4}{*}{\xmark} & Logprobs & \multirow{4}{*}{51.6} & 0.1 & 0.1 & 1.1 & 4.5 & 7.0 & 16.7 & 28.4 & 8.3 & 35.3 \\
     &  & Zero-shot verb. &  & 0.0 & 0.0 & 0.0 & 0.0 & 0.0 & 0.0 & 0.0 & 0.0 & 37.0 \\
     &  & Zero-shot logit &  & 0.0 & 0.0 & 0.0 & 0.0 & 0.0 & 0.0 & 27.7 & 4.0 & 37.7 \\
     &  & Selector &  & 0.0 & 0.0 & 0.0 & 18.1 & 27.1 & 34.8 & 44.8 & 17.8 & 30.4 \\
    \cmidrule(l){2-13}
     & \multirow{5}{*}{\cmark} & Logprobs & \multirow{5}{*}{51.3} & 0.0 & 0.0 & 0.2 & 0.2 & 0.2 & 1.4 & 3.0 & 0.7 & 43.5 \\
     &  & Zero-shot verb. &  & 0.0 & 0.0 & 0.0 & 0.0 & 0.0 & 0.0 & 0.0 & 0.0 & 43.5 \\
     &  & Zero-shot logit &  & 0.0 & 0.0 & 0.0 & 0.0 & 0.0 & 4.0 & 28.7 & 4.7 & 36.4 \\
     &  & Selector &  & 0.0 & 0.0 & 4.0 & 13.5 & 25.9 & 33.8 & 41.0 & 16.9 & 31.9 \\
     &  & SIEVES (ours) &  & 0.0 & 4.0 & 9.9 & 17.9 & 25.9 & 32.4 & 41.1 & 18.7 & 31.0 \\
    \midrule
    \midrule
    \multirow{7}{*}{o3} & \multirow{3}{*}{\xmark} & Zero-shot verb. & \multirow{3}{*}{56.9} & 0.0 & 0.0 & 0.0 & 0.0 & 0.0 & 0.0 & 0.0 & 0.0 & 36.8 \\
     &  & Zero-shot logit &  & 0.0 & 0.0 & 0.0 & 0.0 & 4.5 & 16.6 & 32.6 & 7.7 & 34.1 \\
     &  & Selector &  & 0.0 & 1.1 & 2.2 & 5.3 & 8.1 & 31.1 & 45.9 & 13.4 & 29.3 \\
    \cmidrule(l){2-13}
     & \multirow{4}{*}{\cmark} & Zero-shot verb. & \multirow{4}{*}{58.7} & 0.0 & 0.0 & 0.0 & 0.0 & 0.0 & 0.0 & 0.0 & 0.0 & 36.1 \\
     &  & Zero-shot logit &  & 0.0 & 0.0 & 0.0 & 0.0 & 0.0 & 10.8 & 23.1 & 4.8 & 35.3 \\
     &  & Selector &  & 1.2 & 0.1 & 9.6 & 14.6 & 16.7 & 26.4 & 47.9 & 16.6 & 28.6 \\
     &  & SIEVES (ours) &  & 0.7 & 1.7 & 7.7 & 14.4 & 21.7 & 30.9 & 45.2 & 17.5 & 28.4 \\
    \midrule
    \midrule
    \multirow{7}{*}{Gemini-3-Pro} & \multirow{3}{*}{\xmark} & Zero-shot verb. & \multirow{3}{*}{60.9} & 0.0 & 0.0 & 0.0 & 0.0 & 0.0 & 0.0 & 51.2 & 7.3 & 29.5 \\
     &  & Zero-shot logit &  & 0.0 & 0.0 & 0.0 & 12.7 & 25.8 & 43.5 & 57.9 & 20.0 & 26.9 \\
     &  & Selector &  & 2.4 & 5.1 & 24.2 & 37.4 & 47.2 & 56.8 & 70.1 & 34.7 & 21.9 \\
    \cmidrule(l){2-13}
     & \multirow{4}{*}{\cmark} & Zero-shot verb. & \multirow{4}{*}{63.2} & 0.0 & 0.0 & 0.0 & 0.0 & 0.0 & 0.0 & 65.0 & 9.3 & 28.2 \\
     &  & Zero-shot logit &  & 0.0 & 0.0 & 0.0 & 0.0 & 28.6 & 44.9 & 63.0 & 19.5 & 26.4 \\
     &  & Selector &  & 0.7 & 1.2 & 18.2 & 26.8 & 37.3 & 48.5 & 64.4 & 28.2 & 22.8 \\
     &  & SIEVES (ours) &  & 3.4 & 12.2 & 17.2 & 31.6 & 39.6 & 53.5 & 68.9 & 32.3 & 21.5 \\
    \bottomrule
    \end{tabular}%
    }%
\end{table*}

%% file: tables/supp/baselines/vizwiz_baselines.tex
\begin{table*}[t]
    \caption{\textbf{VizWiz baselines.} \zoomInTool ~ indicates whether the reasoner $f$ explicitly localizes the answer or not. $\overline{C@r}$: average coverage across risk levels 1--30\%.}
    \label{tab:paper_vizwiz_baselines}
    \centering
    \resizebox{\textwidth}{!}{%
    \begin{tabular}{lllrrrrrrrrrr}
    \toprule
    \multirow{2}{*}{$f$} & \multirow{2}{*}{\zoomInTool} & \multirow{2}{*}{Abstention Method} & \multirow{2}{*}{Acc$\uparrow$} & \multicolumn{7}{c}{Coverage at Risk (C@r) $\uparrow$} & \multirow{2}{*}{$\overline{C@r}\uparrow$} & \multirow{2}{*}{AURC$\downarrow$} \\
    &  &  &  & @1 & @5 & @10 & @15 & @20 & @25 & @30 &  &  \\
    \midrule
    \multirow{9}{*}{Pixel-Reasoner} & \multirow{4}{*}{\xmark} & Logprobs & \multirow{4}{*}{38.1} & 0.0 & 0.0 & 0.0 & 0.0 & 0.0 & 0.1 & 0.4 & 0.1 & 48.8 \\
     &  & Zero-shot verb. &  & 0.0 & 0.0 & 0.0 & 0.0 & 0.0 & 0.0 & 0.0 & 0.0 & 46.5 \\
     &  & Zero-shot logit &  & 0.0 & 0.0 & 0.0 & 0.0 & 0.0 & 0.0 & 0.0 & 0.0 & 48.1 \\
     &  & Selector &  & 0.0 & 0.0 & 0.1 & 0.7 & 1.6 & 5.4 & 16.5 & 3.5 & 45.6 \\
    \cmidrule(l){2-13}
     & \multirow{5}{*}{\cmark} & Logprobs & \multirow{5}{*}{38.0} & 0.0 & 0.0 & 0.0 & 0.0 & 0.1 & 0.3 & 1.1 & 0.2 & 51.3 \\
     &  & Zero-shot verb. &  & 0.0 & 0.0 & 0.0 & 0.0 & 0.0 & 0.0 & 0.0 & 0.0 & 50.8 \\
     &  & Zero-shot logit &  & 0.0 & 0.0 & 0.0 & 0.0 & 0.0 & 0.0 & 0.0 & 0.0 & 53.1 \\
     &  & Selector &  & 0.0 & 0.0 & 0.1 & 0.6 & 0.1 & 0.1 & 9.3 & 1.5 & 44.2 \\
     &  & SIEVES (ours) &  & 0.0 & 0.0 & 0.0 & 1.1 & 7.6 & 14.7 & 22.5 & 6.6 & 42.3 \\
    \midrule
    \midrule
    \multirow{7}{*}{o3} & \multirow{3}{*}{\xmark} & Zero-shot verb. & \multirow{3}{*}{45.8} & 0.0 & 0.0 & 0.0 & 0.0 & 0.0 & 0.0 & 0.0 & 0.0 & 43.1 \\
     &  & Zero-shot logit &  & 0.0 & 0.0 & 0.0 & 0.0 & 0.0 & 0.0 & 0.0 & 0.0 & 47.1 \\
     &  & Selector &  & 0.0 & 0.0 & 0.0 & 0.0 & 6.3 & 5.4 & 14.6 & 3.8 & 40.9 \\
    \cmidrule(l){2-13}
     & \multirow{4}{*}{\cmark} & Zero-shot verb. & \multirow{4}{*}{46.9} & 0.0 & 0.0 & 0.0 & 0.0 & 0.0 & 0.0 & 0.0 & 0.0 & 41.6 \\
     &  & Zero-shot logit &  & 0.0 & 0.0 & 0.0 & 0.0 & 0.0 & 0.0 & 0.0 & 0.0 & 39.0 \\
     &  & Selector &  & 0.1 & 0.0 & 0.2 & 5.5 & 0.0 & 1.3 & 19.1 & 3.7 & 35.2 \\
     &  & SIEVES (ours) &  & 0.2 & 0.0 & 0.0 & 7.2 & 11.7 & 27.2 & 40.5 & 12.4 & 33.3 \\
    \midrule
    \midrule
    \multirow{7}{*}{Gemini-3-Pro} & \multirow{3}{*}{\xmark} & Zero-shot verb. & \multirow{3}{*}{44.1} & 0.0 & 0.0 & 0.0 & 0.0 & 0.0 & 0.0 & 0.0 & 0.0 & 43.8 \\
     &  & Zero-shot logit &  & 0.0 & 0.0 & 0.0 & 0.0 & 0.0 & 0.0 & 0.0 & 0.0 & 45.0 \\
     &  & Selector &  & 0.0 & 0.0 & 0.0 & 0.7 & 3.8 & 6.5 & 18.4 & 4.2 & 41.9 \\
    \cmidrule(l){2-13}
     & \multirow{4}{*}{\cmark} & Zero-shot verb. & \multirow{4}{*}{43.6} & 0.0 & 0.0 & 0.0 & 0.0 & 0.0 & 0.0 & 0.0 & 0.0 & 46.2 \\
     &  & Zero-shot logit &  & 0.0 & 0.0 & 0.0 & 0.0 & 0.0 & 0.0 & 0.0 & 0.0 & 47.7 \\
     &  & Selector &  & 0.0 & 0.1 & 1.0 & 4.2 & 0.0 & 4.5 & 9.4 & 2.7 & 39.5 \\
     &  & SIEVES (ours) &  & 0.0 & 0.0 & 0.0 & 2.0 & 8.1 & 21.3 & 29.2 & 8.7 & 37.2 \\
    \bottomrule
    \end{tabular}%
    }%
\end{table*}

%% file: tables/supp/baselines/advqa_baselines.tex
\begin{table*}[t]
    \caption{\textbf{AdVQA baselines.} \zoomInTool ~ indicates whether the reasoner $f$ explicitly localizes the answer or not. $\overline{C@r}$: average coverage across risk levels 1--30\%.}
    \label{tab:paper_advqa_baselines}
    \centering
    \resizebox{\textwidth}{!}{%
    \begin{tabular}{lllrrrrrrrrrr}
    \toprule
    \multirow{2}{*}{$f$} & \multirow{2}{*}{\zoomInTool} & \multirow{2}{*}{Abstention Method} & \multirow{2}{*}{Acc$\uparrow$} & \multicolumn{7}{c}{Coverage at Risk (C@r) $\uparrow$} & \multirow{2}{*}{$\overline{C@r}\uparrow$} & \multirow{2}{*}{AURC$\downarrow$} \\
    &  &  &  & @1 & @5 & @10 & @15 & @20 & @25 & @30 &  &  \\
    \midrule
    \multirow{9}{*}{Pixel-Reasoner} & \multirow{4}{*}{\xmark} & Logprobs & \multirow{4}{*}{54.5} & 0.0 & 0.0 & 0.1 & 0.1 & 0.1 & 0.5 & 1.5 & 0.3 & 42.5 \\
     &  & Zero-shot verb. &  & 0.0 & 0.0 & 0.0 & 0.0 & 0.0 & 0.0 & 0.0 & 0.0 & 42.7 \\
     &  & Zero-shot logit &  & 0.0 & 0.0 & 0.0 & 0.0 & 0.0 & 0.0 & 0.0 & 0.0 & 42.2 \\
     &  & Selector &  & 0.0 & 0.0 & 0.0 & 0.0 & 1.0 & 19.0 & 33.6 & 7.7 & 33.7 \\
    \cmidrule(l){2-13}
     & \multirow{5}{*}{\cmark} & Logprobs & \multirow{5}{*}{58.1} & 0.0 & 0.0 & 0.0 & 0.0 & 0.1 & 0.1 & 0.1 & 0.0 & 40.3 \\
     &  & Zero-shot verb. &  & 0.0 & 0.0 & 0.0 & 0.0 & 0.0 & 0.0 & 0.0 & 0.0 & 40.7 \\
     &  & Zero-shot logit &  & 0.0 & 0.0 & 0.0 & 0.0 & 0.0 & 0.0 & 0.0 & 0.0 & 38.6 \\
     &  & Selector &  & 0.0 & 0.0 & 0.2 & 0.9 & 3.0 & 11.2 & 28.8 & 6.3 & 33.0 \\
     &  & SIEVES (ours) &  & 0.0 & 1.3 & 2.8 & 3.5 & 7.5 & 15.7 & 34.1 & 9.3 & 32.2 \\
    \midrule
    \midrule
    \multirow{7}{*}{o3} & \multirow{3}{*}{\xmark} & Zero-shot verb. & \multirow{3}{*}{68.0} & 0.0 & 0.0 & 0.0 & 0.0 & 0.0 & 0.0 & 0.0 & 0.0 & 31.8 \\
     &  & Zero-shot logit &  & 0.0 & 0.0 & 0.0 & 0.0 & 0.0 & 0.0 & 33.6 & 4.8 & 30.2 \\
     &  & Selector &  & 0.0 & 0.0 & 0.0 & 2.5 & 7.6 & 40.8 & 85.2 & 19.4 & 25.2 \\
    \cmidrule(l){2-13}
     & \multirow{4}{*}{\cmark} & Zero-shot verb. & \multirow{4}{*}{72.4} & 0.0 & 0.0 & 0.0 & 0.0 & 0.0 & 92.4 & 100.0 & 27.5 & 24.9 \\
     &  & Zero-shot logit &  & 0.0 & 0.0 & 0.0 & 0.0 & 38.0 & 75.9 & 100.0 & 30.6 & 21.9 \\
     &  & Selector &  & 0.0 & 0.1 & 3.8 & 15.8 & 30.1 & 83.1 & 100.0 & 33.3 & 20.1 \\
     &  & SIEVES (ours) &  & 0.1 & 0.1 & 8.2 & 18.5 & 43.6 & 84.9 & 100.0 & 36.5 & 19.6 \\
    \midrule
    \midrule
    \multirow{7}{*}{Gemini-3-Pro} & \multirow{3}{*}{\xmark} & Zero-shot verb. & \multirow{3}{*}{80.3} & 0.0 & 0.0 & 0.0 & 0.0 & 100.0 & 100.0 & 100.0 & 42.9 & 16.6 \\
     &  & Zero-shot logit &  & 0.0 & 0.0 & 0.0 & 0.0 & 100.0 & 100.0 & 100.0 & 42.9 & 17.4 \\
     &  & Selector &  & 0.1 & 0.1 & 15.0 & 69.6 & 100.0 & 100.0 & 100.0 & 55.0 & 12.8 \\
    \cmidrule(l){2-13}
     & \multirow{4}{*}{\cmark} & Zero-shot verb. & \multirow{4}{*}{81.4} & 0.0 & 0.0 & 0.0 & 0.0 & 100.0 & 100.0 & 100.0 & 42.9 & 17.0 \\
     &  & Zero-shot logit &  & 0.0 & 0.0 & 0.0 & 0.0 & 100.0 & 100.0 & 100.0 & 42.9 & 17.3 \\
     &  & Selector &  & 0.0 & 0.1 & 32.5 & 79.9 & 100.0 & 100.0 & 100.0 & 58.9 & 11.9 \\
     &  & SIEVES (ours) &  & 0.1 & 6.1 & 33.3 & 78.9 & 100.0 & 100.0 & 100.0 & 59.8 & 11.5 \\
    \bottomrule
    \end{tabular}%
    }%
\end{table*}

%% file: tables/supp/threshold_transfer_full.tex
\begin{table}[tb]
\centering
\caption{\textbf{Transferred thresholds at several error costs.} One threshold per method and cost is chosen on held-out Thyme-MC and applied unchanged to o3 answers. Coverage, realized error rate, and effective reliability $\Phi_c$ are reported in percent. Error rates after transfer may exceed the source risk level.}
\label{tab:threshold_transfer_full}
\setlength{\tabcolsep}{2pt}
\resizebox{\linewidth}{!}{%
\begin{tabular}{lr rrr rrr rrr rrr rrr r}
\toprule
 & & \multicolumn{3}{c}{V* Bench} & \multicolumn{3}{c}{HR-Bench-8k} & \multicolumn{3}{c}{MME-RW-L} & \multicolumn{3}{c}{VizWiz} & \multicolumn{3}{c}{AdVQA} & \\
\cmidrule(lr){3-5}\cmidrule(lr){6-8}\cmidrule(lr){9-11}\cmidrule(lr){12-14}\cmidrule(lr){15-17}
Method & $c$ & Cov. & Err.$\downarrow$ & $\Phi_c\uparrow$ & Cov. & Err.$\downarrow$ & $\Phi_c\uparrow$ & Cov. & Err.$\downarrow$ & $\Phi_c\uparrow$ & Cov. & Err.$\downarrow$ & $\Phi_c\uparrow$ & Cov. & Err.$\downarrow$ & $\Phi_c\uparrow$ & Mean $\Phi_c\uparrow$ \\
\midrule
Selector & 1 & 43.0 & 14.4 & 30.7 & 34.7 & 12.8 & 25.8 & 27.7 & 25.5 & 13.6 & 34.9 & 35.5 & 10.2 & 39.6 & 21.3 & 22.8 & 20.60 \\
SIEVES & 1 & 27.6 & 7.6 & 23.5 & 31.1 & 7.4 & 26.5 & 18.5 & 17.5 & 12.0 & 31.0 & 26.8 & 14.4 & 40.1 & 19.5 & 24.5 & 20.17 \\
\midrule
Selector & 2 & 16.2 & 20.0 & 6.5 & 15.0 & 10.2 & 10.4 & 10.5 & 15.8 & 5.5 & 14.9 & 29.0 & 1.9 & 20.3 & 19.6 & 8.4 & 6.56 \\
SIEVES & 2 & 17.8 & 4.1 & 15.6 & 20.9 & 4.7 & 17.9 & 11.4 & 12.7 & 7.0 & 21.0 & 23.4 & 6.3 & 29.6 & 17.3 & 14.3 & 12.22 \\
\midrule
Selector & 4 & 2.0 & 0.0 & 2.0 & 5.7 & 1.8 & 5.2 & 3.3 & 4.4 & 2.6 & 6.2 & 15.4 & 1.4 & 10.8 & 13.8 & 3.4 & 2.91 \\
SIEVES & 4 & 6.4 & 3.3 & 5.3 & 8.8 & 2.0 & 7.9 & 3.7 & 4.5 & 2.9 & 8.5 & 16.9 & 1.3 & 14.9 & 13.2 & 5.1 & 4.51 \\
\midrule
Selector & 5 & 2.0 & 0.0 & 2.0 & 5.7 & 1.8 & 5.1 & 3.3 & 4.4 & 2.5 & 6.2 & 15.4 & 0.5 & 10.8 & 13.8 & 1.8 & 2.37 \\
SIEVES & 5 & 3.9 & 2.7 & 3.2 & 6.5 & 0.4 & 6.4 & 2.7 & 3.8 & 2.1 & 5.7 & 14.5 & 0.7 & 11.3 & 11.8 & 3.3 & 3.15 \\
\midrule
Selector & 10 & 0.1 & 100.0 & -1.0 & 0.4 & 0.0 & 0.4 & 0.2 & 0.0 & 0.2 & 2.5 & 19.4 & -2.8 & 1.6 & 12.7 & -0.7 & -0.78 \\
SIEVES & 10 & 0.1 & 0.0 & 0.1 & 0.9 & 0.0 & 0.9 & 0.3 & 0.0 & 0.3 & 1.1 & 12.5 & -0.4 & 2.5 & 7.3 & 0.5 & 0.28 \\
\midrule
Selector & 100 & 0.0 & 0.0 & 0.0 & 0.0 & 0.0 & 0.0 & 0.0 & 0.0 & 0.0 & 0.0 & 0.0 & 0.0 & 0.0 & 50.0 & -1.0 & -0.19 \\
SIEVES & 100 & 0.0 & 0.0 & 0.0 & 1.3 & 0.0 & 1.3 & 0.4 & 0.0 & 0.4 & 0.5 & 13.0 & -6.5 & 1.9 & 9.2 & -15.3 & -4.02 \\
\bottomrule
\end{tabular}}
\end{table}

%% file: tables/supp/threshold_transfer_all_reasoners.tex
\begin{table}[tb]
\centering
\caption{\textbf{Fixed-threshold transfer averaged over three reasoners.}
Effective reliability $\Phi_c$ (\%) averaged over Pixel-Reasoner, o3, and Gemini-3-Pro.
Thresholds are chosen on held-out training-domain data and applied unchanged to the OOD benchmarks.}
\label{tab:threshold_transfer_all_reasoners}
\setlength{\tabcolsep}{4pt}
\resizebox{\linewidth}{!}{%
\begin{tabular}{lrrrrrrr}
\toprule
Reasoning + abstention method & $c$ & V* & HR-Bench & MME-RW-L & VizWiz & AdVQA & Mean $\Phi_c$ \\
\midrule
No localization + Selector & 2 & 10.89 & 19.12 & 8.54 & 2.16 & 9.63 & 10.07 \\
Localization + Selector & 2 & 12.70 & 16.82 & 8.98 & 0.32 & 9.11 & 9.59 \\
Localization + SIEVES & 2 & \textbf{18.15} & \textbf{25.64} & \textbf{10.21} & \textbf{4.15} & \textbf{14.40} & \textbf{14.51} \\
\midrule
No localization + Selector & 4 & 2.23 & 2.76 & 2.59 & -0.23 & 0.49 & 1.57 \\
Localization + Selector & 4 & 3.84 & 8.20 & 2.94 & \textbf{-0.11} & 1.90 & 3.35 \\
Localization + SIEVES & 4 & \textbf{7.19} & \textbf{12.65} & \textbf{4.97} & -0.79 & \textbf{4.54} & \textbf{5.71} \\
\midrule
No localization + Selector & 5 & 2.16 & 2.31 & 2.22 & \textbf{-0.94} & -0.26 & 1.10 \\
Localization + Selector & 5 & 3.67 & 7.96 & 2.44 & -1.71 & -0.22 & 2.43 \\
Localization + SIEVES & 5 & \textbf{5.31} & \textbf{9.69} & \textbf{3.32} & -1.37 & \textbf{2.66} & \textbf{3.92} \\
\bottomrule
\end{tabular}}
\end{table}

%% file: tables/supp/baselines/thyme_750_mc_baselines.tex
\begin{table*}[t]
    \caption{\textbf{Thyme-750 (MC) baselines.} \zoomInTool ~ indicates whether the reasoner $f$ explicitly localizes the answer or not. $\overline{C@r}$: average coverage across risk levels 1--30\%.}
    \label{tab:paper_thyme_750_mc_baselines}
    \centering
    \resizebox{\textwidth}{!}{%
    \begin{tabular}{lllrrrrrrrrrr}
    \toprule
    \multirow{2}{*}{$f$} & \multirow{2}{*}{\zoomInTool} & \multirow{2}{*}{Abstention Method} & \multirow{2}{*}{Acc$\uparrow$} & \multicolumn{7}{c}{Coverage at Risk (C@r) $\uparrow$} & \multirow{2}{*}{$\overline{C@r}\uparrow$} & \multirow{2}{*}{AURC$\downarrow$} \\
    &  &  &  & @1 & @5 & @10 & @15 & @20 & @25 & @30 &  &  \\
    \midrule
    \multirow{9}{*}{Pixel-Reasoner} & \multirow{4}{*}{\xmark} & Logprobs & \multirow{4}{*}{53.7} & 0.3 & 0.3 & 0.8 & 1.1 & 2.1 & 4.8 & 19.3 & 4.1 & 36.3 \\
     &  & Zero-shot verb. &  & 0.0 & 0.0 & 0.0 & 0.0 & 0.0 & 0.0 & 0.0 & 0.0 & 36.8 \\
     &  & Zero-shot logit &  & 0.0 & 0.0 & 0.0 & 0.0 & 0.0 & 0.0 & 0.0 & 0.0 & 39.2 \\
     &  & Selector &  & 0.2 & 0.2 & 1.4 & 10.0 & 17.1 & 35.0 & 53.4 & 16.8 & 30.1 \\
    \cmidrule(l){2-13}
     & \multirow{5}{*}{\cmark} & Logprobs & \multirow{5}{*}{54.0} & 0.1 & 0.1 & 0.1 & 0.1 & 0.4 & 1.1 & 1.2 & 0.4 & 42.0 \\
     &  & Zero-shot verb. &  & 0.0 & 0.0 & 0.0 & 0.0 & 0.0 & 0.0 & 0.0 & 0.0 & 44.6 \\
     &  & Zero-shot logit &  & 0.0 & 0.0 & 0.0 & 0.0 & 0.0 & 0.0 & 0.0 & 0.0 & 42.5 \\
     &  & Selector &  & 0.8 & 1.1 & 8.4 & 22.3 & 39.6 & 51.9 & 63.0 & 26.7 & 25.8 \\
     &  & SIEVES (ours) &  & 0.7 & 2.1 & 9.9 & 24.9 & 38.7 & 50.3 & 59.6 & 26.6 & 25.7 \\
    \bottomrule
    \end{tabular}%
    }%
\end{table*}

%% file: tables/benchmarks.tex
\begin{table}[tb]
  \centering
  \setlength{\tabcolsep}{3pt}
  \scalebox{0.7}{%
  \begin{tabular}{l r c c l}
  \toprule
  Benchmark & \#Q & Format & Avg.\ res. & Domain \\
  \midrule
  V* Bench \citep{wu2023v}
    & 191 & MC & 2246$\times$1582
    & Natural images: Attribute recognition, relative position \\
  HR-Bench-8k \citep{hrbench_wang2024}
    & 800 & MC & 7680$\times$4320
    & Attribute recognition, OCR, map \& chart analysis, spatial reasoning \\
  MME-RW-L \citep{mme_realworld_2024}
    & 1{,}782 & MC & 2000$\times$1500
    & Remote sensing, autonomous driving, monitoring, diagrams \& tables, OCR \\
  VizWiz \citep{gurari2018vizwiz}
    & 4{,}319 & OE & 1224$\times$1224
    & Blind-user mobile photos: yes/no, counting, other (32\% unanswerable) \\
  AdVQA \citep{sheng2021advqa}
    & 10{,}000 & OE & 640$\times$480
    & Adversarially crafted: counting, OCR, rare concepts, reasoning \\
  \bottomrule
  \end{tabular}%
  }%
  \vspace{5pt}
  \caption{\textbf{Out-Of-Distribution Benchmarks.} 
  We evaluate SIEVES on diverse benchmarks of varying sizes (\#Q), using Multiple-Choice (MC) and Open-Ended (OE) formats, some of which don't require to zoom-in.}
  \label{tab:benchmarks}
  \vspace{-10pt}
\end{table}

%% file: figs/supp_thyme_mc_prompt.tex
\begin{figure}[t]
    \centering
    \begin{fitpromptbox}
    \begin{promptfloatlisting}[title={Thyme multiple-choice distractor generation prompt}]
System message:
You are an expert at creating plausible but incorrect distractor options for visual question answering tasks. Given a question, image, ground truth answer, and optionally a wrong answer from a model, generate additional plausible distractor options that are wrong but could seem reasonable.

User message (same-category example):
Category: {category} Question: {question} Ground truth answer: {ground_truth}

Here is the cropped region of interest: [cropped ground-truth box image]

Generate 2-3 plausible but incorrect distractor options.

Assistant message (example distractors):
[example distractor 1]
[example distractor 2]
[example distractor 3]

User message (current Thyme example):
Category: {category} Question: {question} Ground truth answer: {ground_truth}
Wrong answer from a model: {wrong_answer} [included only when available]

Here is / Here are the cropped region(s) of interest from the image: [one or more cropped ground-truth box images]

Generate exactly {num_distractors_needed} plausible but incorrect distractor options. Make them specific and realistic based on the image. Return only the distractors, one per line, without numbering.
    \end{promptfloatlisting}
    \end{fitpromptbox}
    \captionsetup{skip=4pt}
    \caption{
        \textbf{Thyme multiple-choice distractor generation prompt.}
        We use this prompt to turn Thyme examples into multiple-choice questions by generating plausible wrong options that stay close to the image and question. 
        The template is shown as a four-message chat: a system instruction, one fixed same-category example, its assistant response, and the current Thyme example to complete. 
        The Thyme categories are \texttt{ocr\_recognition}, \texttt{quantity\_recognition}, \texttt{object\_recognition}, \texttt{attribute\_recognition}, \texttt{position\_recognition}, and \texttt{chart\_understanding}; the few-shot example is chosen from the same category as the current question. 
        If an incorrect Pixel-Reasoner answer is available, it is included so the generated distractors can target realistic model mistakes.
    }
    \label{fig:supp_thyme_mc_prompt}
\end{figure}

%% file: figs/supp_forced_zoom_prompt.tex
\begin{figure}[t]
    \centering
    \begin{fitpromptbox}
    \begin{promptfloatlisting}[title={Prompt for reasoner with localization}]
Guidelines: Understand the given visual information and the user query. Employ the given visual operations (tools) to observe better the visual elements necessary to answer the question. For an image, we can look closer by `crop_image_normalized`. Please always crop into the relevant region before answering the question. You must use the cropping tool first and finish your turn. After the user returns the zoomed-in image, you can reason about what you observe and give your final answer. You must always use the cropping tool in the first turn, end your turn, wait for the cropped image. If you see the relevant object or objects needed to answer the question clearly, you can give the final answer. Otherwise, try to use the `crop_image_normalized` tool again, until you find the relevant object of objects. You have up to 10 turns. Reason with the visual information step by step, and put your final answer within \boxed{}. It is very important to follow these instructions.
    \end{promptfloatlisting}
    \end{fitpromptbox}
    \caption{
        \textbf{Prompt for reasoner with localization.}
        We prompt the reasoner with the tool definition and the guidelines presented here, which encourage it to use the zoom-in tool to provide visual evidence for its answer.
    }
    \label{fig:supp_forced_zoom_prompt}
\end{figure}

%% file: figs/supp_correctness_judge_prompt.tex
\begin{figure}[t]
    \centering
    \begin{fitpromptbox}
    \begin{promptfloatlisting}[title={Open-ended correctness-judge prompt}]
Compare the predicted answer to the ground truth answer and determine if they convey the same meaning, and if the model was likely referring to the same object or situation when answering visual questions.

You can think step-by-step about whether the predicted answer conveys the same meaning as the ground truth answer, but after that, output only ANSWER: and yes or no after that.
Sometimes, the predicted answer will also contain the model's thinking process and justificatioon. You can take it into account, but focus on the validity of the final answer, which will generally appear at the end.
If the ground truth answer is only 't', this means the question is not answerable. In that case, you should always mark the question as wrong, because no answer can be correct, not even saying that there is no answer.

For example:

Question: What is the person holding?
Predicted answer: Blue pullover
Ground truth answer: Sweater

A pullover is a type of sweater. The colour is not mentioned in the ground truth answer, so we can assume it is correct. ANSWER: yes

## Another example ##
Question: Which object was put down by the person?
Predicted answer: jacket
Ground truth answer: The shoe.
A jacket is a type of clothing, similar to a shoe, which is also a type of clothing/accessory.
However, a jacket does not look similar to a shoe, so it is unlikely that the model confused them, and instead simply answered incorrectly. ANSWER: no
- The other options (food, blanket, sandwich) are not related to clothing or accessories.

## Another example ##
Question: What verification is this paper for?
Predicted answer: freeboard verification
Ground truth answer: Freeboard

Clearly the model shows in its response it refers to the verification being a freeboard verification, which conveys the same meaning. ANSWER: yes

## Real user request ##
Question: {question}
Predicted answer: {pred_answer}
Ground truth answer: {gt_answer}
\end{promptfloatlisting}
    \end{fitpromptbox}
    \captionsetup{skip=4pt}
    \caption{
        \textbf{Open-ended judge prompt.} 
        For open-ended questions, we use this prompt when normalized exact match fails to decide whether the predicted answer should still count as correct. 
        This makes correctness labels more robust to valid rewordings and semantically equivalent answers that exact match would reject. 
        We prompt Qwen3-8B with a single user message at temperature 0. 
        If multiple ground-truth answers are available, the judge is queried sequentially and stops at the first positive match.
    }
    \label{fig:supp_correctness_judge_prompt}
\end{figure}

%% file: figs/supp_grounding_coherence_prompt.tex
\begin{figure}[t]
    \centering
    \begin{fitpromptbox}
    \begin{promptfloatlisting}[title={Grounding-coherence labeling prompt for selector training}]
You are an expert evaluator assessing whether a model's response to a visual question is grounded in the provided image crop.

Given:
- Question: {question}
- Model's Response (with final answer in \boxed{}): {last_message_with_boxed_answer}
- Image: [Provided image crop]

Please evaluate two aspects:

1. **Crop Sufficiency**: Is the provided image crop sufficient to support the model's response? Does it contain all the necessary visual information referenced in the response? If the model explicitly states they use the global view to answer this question, you should consider this as not grounded in the prompt. Note you are not provided this final image, and only the crop, which the model should only use to give the final answer.

2. **Answer Coherence**: Is the model's response coherent with what is actually visible in the image? Or is the model hallucinating information or obtaining it from elsewhere (not from the image)?

Think step by step about both aspects, then provide your final assessment.

Output your final decision as \boxed{Yes} if the answer is well-grounded in the image crop (both crop is sufficient AND answer is coherent), or \boxed{No} if there are issues with either aspect.

Examples:
- If the crop shows a clear view of a red car and the model answers "red car" -> \boxed{Yes}
- If the crop shows a partial view that doesn't contain enough information to answer -> \boxed{No}
- If the crop shows a dog but the model answers "cat" -> \boxed{No}
- If the crop shows a room but the model mentions specific details not visible in the crop -> \boxed{No}

Your response:
    \end{promptfloatlisting}
    \end{fitpromptbox}
    \caption{
        \textbf{Crop-answer grounding coherence labeling prompt for selector training.} 
        We use this prompt to create the crop-answer coherence labels used to train the selector. 
        The judge receives the question, the reasoner's final answer, and the visual evidence (crop), and decides whether that crop contains enough evidence and whether the answer is supported by what is visible. 
        The parser reads \texttt{\textbackslash boxed\{Yes\}} or \texttt{\textbackslash boxed\{No\}} as a binary coherence label. 
    }
    \label{fig:supp_grounding_coherence_prompt}
\end{figure}

%% file: figs/supp_gemini_flash_bbox_prompt.tex
\begin{figure}[tb]
    \centering
    \begin{fitpromptbox}
    \begin{promptfloatlisting}[title={Subprompt 1: target object extraction}]
System:
You extract the target object from a visual question. Return only a short noun phrase describing the object to locate in the image. If the question uses relational or positional language, keep that context (e.g., 'object on the woman's left ring finger').
User:
{question}
    \end{promptfloatlisting}
    \end{fitpromptbox}

\vspace{0.5em}

    \begin{fitpromptbox}
    \begin{promptfloatlisting}[title={Subprompt 2: full-image grounding}]
System:
You are a precise visual grounding assistant. Return only JSON.
User:
Please return the bounding box coordinates of "{target_object}". Use normalized 0-1000 coordinates in [top, left, bottom, right] order. Return a JSON list like: [{"box_2d": [top, left, bottom, right], "label": "..."}].
Image: [Provided full image]
    \end{promptfloatlisting}
    \end{fitpromptbox}
    \caption{
        \textbf{Localization annotation prompt for SIEVES training.} 
        We use this two-stage pipeline to automatically annotate Thyme with bounding boxes indicating the object referred to in the question.
        The two subprompts are shown separately.
        The first call extracts the target object phrase from the question, and the second call takes the full image plus that extracted phrase and returns one or more boxes.
    }
    \label{fig:supp_gemini_flash_bbox_prompt}
\end{figure}

%% file: checklist.tex
\section*{NeurIPS Paper Checklist}

\begin{enumerate}

\item {\bf Claims}
    \item[] Question: Do the main claims made in the abstract and introduction accurately reflect the paper's contributions and scope?
    \item[] Answer: \answerYes{}
    \item[] Justification: The abstract and introduction state the paper's main empirical claims and scope, namely grounding-aware selective prediction, transfer to stronger proprietary reasoners, and evaluation on five OOD VQA benchmarks. See the abstract and \Cref{sec:intro,sec:results}.

\item {\bf Limitations}
    \item[] Question: Does the paper discuss the limitations of the work performed by the authors?
    \item[] Answer: \answerYes{}
    \item[] Justification: We discuss that SIEVES uses bounding-box annotations and a separate VLM for crop-answer coherence labeling during training, which adds annotation/modeling cost. We argue this is a one-time training cost for improved reliability, and ablate scalable alternatives: bounding boxes can be generated automatically with little loss, and the same base model can be used for coherence labeling with competitive coverage; see \Cref{sec:method,tab:coherence_ablation_id_ood,tab:bbox_ablation_id_ood,tab:ablation_o3_g3p}.

\item {\bf Theory assumptions and proofs}
    \item[] Question: For each theoretical result, does the paper provide the full set of assumptions and a complete (and correct) proof?
    \item[] Answer: \answerNA{}
    \item[] Justification: The paper is empirical and does not present theorems or formal proofs requiring this kind of treatment.

    \item {\bf Experimental result reproducibility}
    \item[] Question: Does the paper fully disclose all the information needed to reproduce the main experimental results of the paper to the extent that it affects the main claims and/or conclusions of the paper (regardless of whether the code and data are provided or not)?
    \item[] Answer: \answerYes{}
    \item[] Justification: The paper specifies the method, selector objectives, evaluation protocol, benchmark setup, training hyperparameters, and prompt templates used for judging and pseudo-annotation. See \Cref{sec:method,sec:experiments,app:experimental_details,app:selective_metrics,app:ood_benchmarks,app:prompt_templates}.

\item {\bf Open access to data and code}
    \item[] Question: Does the paper provide open access to the data and code, with sufficient instructions to faithfully reproduce the main experimental results, as described in supplemental material?
    \item[] Answer: \answerNo{}
    \item[] Justification: The submission does not currently provide an anonymized code release or reproduction package. We use public benchmarks and document the setup in the paper and appendix, but the code and exact run scripts are not part of this submission, although the code will be made public at a later date.

\item {\bf Experimental setting/details}
    \item[] Question: Does the paper specify all the training and test details (e.g., data splits, hyperparameters, how they were chosen, type of optimizer) necessary to understand the results?
    \item[] Answer: \answerYes{}
    \item[] Justification: The training data sources, selector architecture, loss, benchmark choices, splits, repeated-answer protocol, and key hyperparameters are described in the main text and appendix. See \Cref{sec:method,sec:experiments,app:experimental_details,app:ood_benchmarks}.

\item {\bf Experiment statistical significance}
    \item[] Question: Does the paper report error bars suitably and correctly defined or other appropriate information about the statistical significance of the experiments?
    \item[] Answer: \answerYes{}
    \item[] Justification: We improve statistical robustness by sampling multiple inference responses per question and aggregating selective-prediction metrics over the pooled set of answers, as detailed in \Cref{app:selective_metrics}. We also study pooled evaluation versus per-repetition evaluation and averaging in \Cref{tab:repeated_answer_pooling}, finding that pooling is more robust while both estimates are empirically similar; this is complemented by evaluation across five OOD benchmarks.

\item {\bf Experiments compute resources}
    \item[] Question: For each experiment, does the paper provide sufficient information on the computer resources (type of compute workers, memory, time of execution) needed to reproduce the experiments?
    \item[] Answer: \answerYes{}
    \item[] Justification: We discuss compute resources in the checklist: selector training uses H100 GPUs with 80GB memory, and training completes in under 15 hours on one 8-GPU node.
    
\item {\bf Code of ethics}
    \item[] Question: Does the research conducted in the paper conform, in every respect, with the NeurIPS Code of Ethics \url{https://neurips.cc/public/EthicsGuidelines}?
    \item[] Answer: \answerYes{}
    \item[] Justification: We have reviewed the NeurIPS Code of Ethics and the work is consistent with it to the best of our knowledge. The paper studies reliability and abstention on public benchmarks and aims to reduce harmful overconfident behavior in high-stakes settings.

\item {\bf Broader impacts}
    \item[] Question: Does the paper discuss both potential positive societal impacts and negative societal impacts of the work performed?
    \item[] Answer: \answerYes{}
    \item[] Justification: We discuss that improving selective reliability can make VLMs more useful in high-stakes settings where abstention is preferable to unreliable answers. At the same time, increasing reliability also raises the feasibility of more autonomous systems, so deployment should still consider the consequences of incorrect accepted answers and deferred decisions.
    
\item {\bf Safeguards}
    \item[] Question: Does the paper describe safeguards that have been put in place for responsible release of data or models that have a high risk for misuse (e.g., pre-trained language models, image generators, or scraped datasets)?
    \item[] Answer: \answerNA{}
    \item[] Justification: The paper does not release a high-risk generative model or a scraped dataset. Its primary contribution is a selective-prediction method intended to make deployment safer by enabling abstention.

\item {\bf Licenses for existing assets}
    \item[] Question: Are the creators or original owners of assets (e.g., code, data, models), used in the paper, properly credited and are the license and terms of use explicitly mentioned and properly respected?
    \item[] Answer: \answerNo{}
    \item[] Justification: The paper credits the datasets, models, and software we build on via citations, but it does not exhaustively document licenses and terms of use for every external asset in the submission itself.

\item {\bf New assets}
    \item[] Question: Are new assets introduced in the paper well documented and is the documentation provided alongside the assets?
    \item[] Answer: \answerNA{}
    \item[] Justification: The submission does not release a new dataset, model, or code asset as part of the paper package.

\item {\bf Crowdsourcing and research with human subjects}
    \item[] Question: For crowdsourcing experiments and research with human subjects, does the paper include the full text of instructions given to participants and screenshots, if applicable, as well as details about compensation (if any)? 
    \item[] Answer: \answerNA{}
    \item[] Justification: The paper does not report a crowdsourcing study or a study with recruited human participants. It uses public datasets and includes only a small internal manual sanity check for label agreement in \Cref{sec:method}.

\item {\bf Institutional review board (IRB) approvals or equivalent for research with human subjects}
    \item[] Question: Does the paper describe potential risks incurred by study participants, whether such risks were disclosed to the subjects, and whether Institutional Review Board (IRB) approvals (or an equivalent approval/review based on the requirements of your country or institution) were obtained?
    \item[] Answer: \answerNA{}
    \item[] Justification: The paper does not present a human-subjects study requiring IRB review.

\item {\bf Declaration of LLM usage}
    \item[] Question: Does the paper describe the usage of LLMs if it is an important, original, or non-standard component of the core methods in this research? Note that if the LLM is used only for writing, editing, or formatting purposes and does \emph{not} impact the core methodology, scientific rigor, or originality of the research, declaration is not required.
    \item[] Answer: \answerNA{}
    \item[] Justification: the LLM is used only for writing, editing, or formatting purposes and does \emph{not} impact the core methodology, scientific rigor, or originality of the research.

\end{enumerate}